\documentclass[runningheads]{llncs}

\usepackage{eccv}

\usepackage{eccvabbrv}
\usepackage{amsmath}
\usepackage{amssymb}
\usepackage{epigraph}
\usepackage{url}            %
\usepackage{amsfonts}       %
\usepackage{nicefrac}       %
\usepackage{xcolor,colortbl}         %
\usepackage{xspace}
\usepackage{multirow}
\usepackage{mathtools}
\usepackage{overpic}
\usepackage{caption}
\usepackage{chngcntr}
\usepackage{tikz}
\usepackage{algorithm}
\usepackage{algpseudocode}
\usepackage{array}

\newcolumntype{H}{>{\setbox0=\hbox\bgroup}c<{\egroup}@{}}
\newcolumntype{Z}{>{\setbox0=\hbox\bgroup}c<{\egroup}@{\hspace*{-\tabcolsep}}}

\usepackage{graphicx}
\usepackage{booktabs}

\usepackage[accsupp]{axessibility}  %

\usepackage{orcidlink}

\newif\ifcomments

\commentstrue

\newif\ifarxiv
\arxivtrue

\newcommand{\arxiv}[1]{\ifarxiv{#1}\fi}
\newcommand{\conf}[1]{\ifarxiv\else{#1}\fi}

\usepackage{wrapfig}

\usepackage{pifont}

\newcommand{\PAR}[1]{\vskip4pt \noindent {\bf #1~}}
\newcommand{\PARit}[1]{\vskip4pt \noindent {\it #1~}}

\newcommand{\DONE}[1]{{#1}}

\newcommand{\sal}{\texttt{\textbf{SAL}}\@\xspace}
\newcommand{\thing}{\texttt{things}\@\xspace}
\newcommand{\stuff}{\texttt{stuff}\@\xspace}
\newcommand{\lps}{\textit{Lidar Panoptic Segmentation}\@\xspace}

\newcommand{\crossmark}{\ding{55}}

\definecolor{citecolor}{RGB}{34,139,34}
\hypersetup{
    pagebackref=true,
    breaklinks=true,
    colorlinks,
    citecolor=citecolor,
    bookmarks=false
}

\begin{document}

\title{Better Call SAL: Towards Learning to \\Segment Anything in Lidar} 

\titlerunning{Segment Anything in Lidar}

\author{
    Aljoša Ošep\textsuperscript{1*} \quad
    Tim Meinhardt\textsuperscript{1*} \quad
    Francesco Ferroni\textsuperscript{1} \quad
    Neehar Peri\textsuperscript{2} \\
    Deva Ramanan\textsuperscript{2} \quad
    Laura Leal-Taixé\textsuperscript{1} \vspace{.5em}\\
    \small{\textsuperscript{*}Equal contribution}\\
    \arxiv{\small{\texttt{\url{https://github.com/nv-dvl/segment-anything-lidar}}}}
}

\authorrunning{A.~Ošep et al.}

\institute{\textsuperscript{1}NVIDIA \qquad \textsuperscript{2}CMU}

\maketitle

\begin{figure}[h!]
\centering
\captionsetup[subfigure]{labelformat=empty,justification=centering}
\begin{subfigure}{.23\textwidth}
  \centering
  \includegraphics[width=\linewidth]{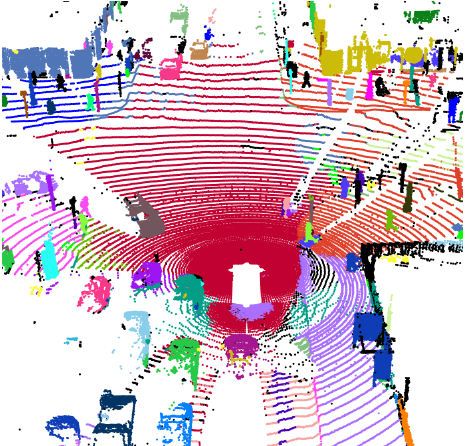}
  \caption{Class-agnostic \\instance segmentation}
  \label{fig:teaser_1}
\end{subfigure}%
\hspace{1mm} %
\begin{subfigure}{.23\textwidth}
  \centering
  \includegraphics[width=\linewidth]{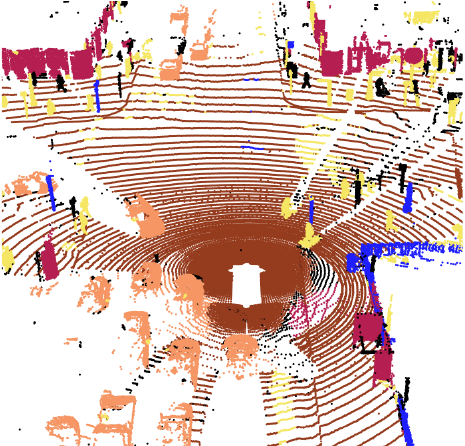}
  \caption{Text prompts: \\ \{\texttt{car}, \texttt{building}, $\ldots$\}}
  \label{fig:teaser_2}
\end{subfigure}
\hspace{0.3mm} %
\begin{subfigure}{.23\textwidth}
  \centering
  \includegraphics[width=\linewidth]{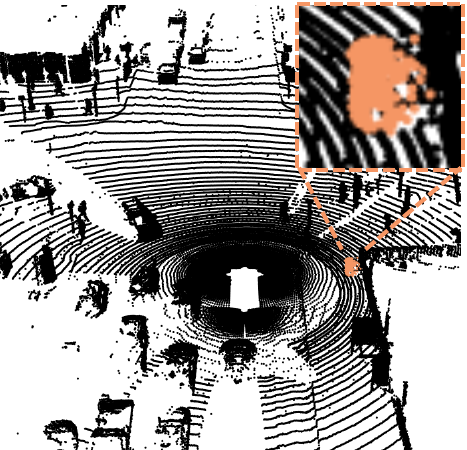}
  \caption{Text prompts: \\ \{\texttt{trash bin}\}}
  \label{fig:teaser_3}
\end{subfigure}
\hspace{0.3mm} %
\begin{subfigure}{.23\textwidth}
  \centering
\includegraphics[width=\linewidth]{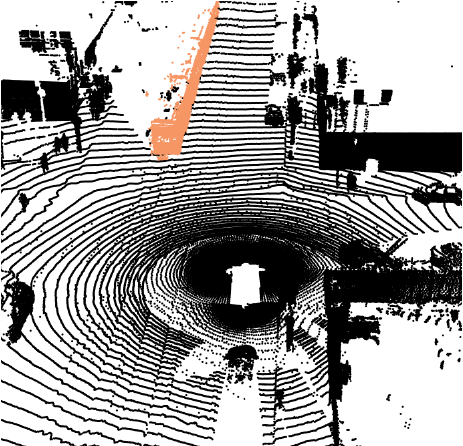}
  \caption{Text prompts: \\ \{\texttt{streetcar}\}}
  \label{fig:teaser_4}
\end{subfigure}
\caption{ 
The \sal (\texttt{\textbf{S}}egment \texttt{\textbf{A}}nything in \texttt{\textbf{L}}idar) model performs class-agnostic instance segmentation (i) and zero-shot classification via text prompting. This allows us to not only predict panoptic segmentation (ii) for fixed class vocabularies but also segment any object (iii and iv) in a given Lidar scan. 
}
\label{fig:teaser}
\end{figure}

\begin{abstract}
    We propose the \sal (\texttt{\textbf{S}}egment \texttt{\textbf{A}}nything in \texttt{\textbf{L}}idar) method consisting of a text-promptable zero-shot model for segmenting and classifying any object in Lidar, and a pseudo-labeling engine that facilitates model training without manual supervision. While the established paradigm for \lps (LPS) relies on manual supervision for a handful of object classes defined a priori, we utilize 2D vision foundation models to generate 3D supervision ``for free''. Our pseudo-labels consist of instance masks and corresponding CLIP tokens, which we lift to Lidar using calibrated multi-modal data. By training our model on these labels, we distill the 2D foundation models into our Lidar \sal model. Even without manual labels, our model achieves $91\%$ in terms of class-agnostic segmentation and $54\%$ in terms of zero-shot LPS of the fully supervised state-of-the-art. Furthermore, we outperform several baselines that do not distill but only lift image features to 3D. More importantly, we demonstrate that \sal supports arbitrary class prompts, can be easily extended to new datasets, and shows significant potential to improve with increasing amounts of self-labeled data. We release all models and the code.
\end{abstract}

\section{Introduction}

We tackle segmentation and recognition of objects in Lidar point clouds, a task commonly tackled via \lps (LPS).

\PAR{Status quo.} 
LPS has been gaining significant attention in the community due to its role in scene understanding, which is vital for safe autonomous navigation. 
Thanks to the availability of labeled datasets~\cite{behley2019iccv, fong21ral} and advances in learning representations from unordered point sets~\cite{Qi17CVPR_pointnet,Qi17NIPS, Thomas19ICCV,choy20194d}, the Lidar community made significant progress in learning to segment and classify instances of pre-defined and manually-labeled classes.  
While this progress has been impressive, existing models~\cite{marcuzzi2023ral,Agarwalla23iros,hong2021lidar,zhou2021panoptic,sirohi2021efficientlps,razani2021gp,li2022panoptic} cannot adapt to continually evolving class ontologies~\cite{lin2022continual} that may even vary in different geographic regions~\cite{wang2020train}. 

\PAR{Stirring the pot.} We challenge this well-established approach and investigate how we can train general LPS models that can be prompted to segment point clouds according to \textit{any} object class vocabulary. 
Towards such a promptable Lidar segmentation approach, we first need to be able to segment any object, a capability that remains elusive in the Lidar domain.

\begin{figure}[t]
\centering
\includegraphics[trim={0 50px 0 0},width=\linewidth]{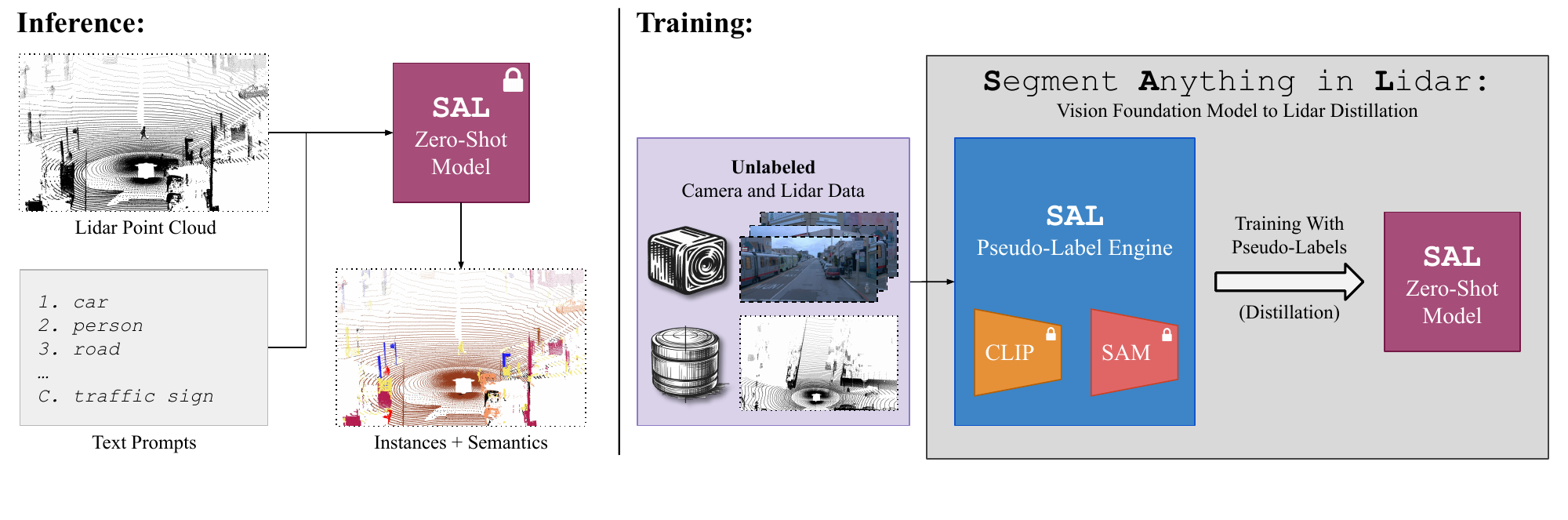}
\caption{
\textbf{\sal overview:} Given a Lidar scan and a class vocabulary prompt, specified as a list of per-class free-form text descriptions (\textit{left}), \sal segments and classifies objects (\thing and \stuff classes). 
As labeled data for training such a model does not exist, we supervise \sal by distilling off-the-shelf vision foundation models to Lidar (\textit{right}). 
}
\label{fig:method-main}
\end{figure}

\PAR{Towards segmenting anything in Lidar.} 
To address these challenges, we propose \sal, which consists of a text-promptable zero-shot model (\cref{fig:method-main}, \textit{left}) for panoptic segmentation of arbitrary objects (\cref{fig:teaser}), and pseudo-labeling engine (\cref{fig:method-main}, \textit{right}) that facilitates model training directly from raw sensory data. 
The \sal pseudo-label engine automatically labels Lidar sequences using image segmentation~\cite{kirillov2023segment} and vision-language models~\cite{radford2021learning}. We utilize SAM~\cite{kirillov2023segment} to generate class-agnostic masks in images, and CLIP~\cite{radford2021learning} to generate per-mask tokens that connect visual features to language, and finally, transfer both to Lidar using a calibrated sensory setup. 
Even though generated pseudo-labels only partially cover Lidar scans and are inherently noisy due to errors in the image-level generation process and imperfect sensory calibration, we demonstrate their effectiveness as self-supervised training data for the \sal zero-shot model. In contrast to prior work in LPS, \sal can be prompted with any semantic vocabulary during test-time without model re-training or tuning. Different from recent advances in zero-shot semantic segmentation~\cite{Peng2023OpenScene} \sal \textit{does not} require image features during inference and can segment full $360^{\circ}$ point clouds. 

\PAR{Talk to your \sal.} 
Our model predicts Lidar segmentation masks and their corresponding CLIP tokens, which allow us to perform zero-shot classification of segmented objects using arbitrary text prompts (\cref{fig:teaser}). 
We evaluate \sal by prompting our zero-shot model on standard benchmarks for LPS. 
For class-agnostic segmentation, we reach $91\%$ of the performance of manually supervised baselines. By classifying segmented objects in a zero-shot manner, we report the \textit{first} (and very encouraging) results for zero-shot LPS and reach $42$\% and $54\%$ of the supervised baselines trained on SemanticKITTI and nuScenes, respectively. 
Beyond that, as shown in \cref{fig:teaser}, we can prompt \sal model to segment objects outside of existing Lidar dataset class vocabularies.

\PAR{Contributions.} 
As our main contribution, we re-think the established approach to \lps and (i) present \sal for segmentation and classification of any object in a Lidar scan.
As we do not utilize labeled Lidar data, we propose (ii) a pseudo-label engine that distills vision foundation models to the Lidar domain.
The resulting pseudo-labels allow us to train (iii) a zero-shot LPS model without any human supervision.
We demonstrate (iv) encouraging results on standard LPS benchmarks and outline a clear path towards universal, promptable segmentation foundation models for Lidar data.

 \section{Related Work}
\label{sec:related}

Lidar has played a pivotal role since the dawn of embodied navigation~\cite{thorpe1991toward,thrun2006stanley,Petrovskaya09AR}.
Recent data-driven efforts in Lidar perception have been pushing boundaries in semantic segmentation~\cite{xiong11icra,Wu18ICRA,Wu19ICRA,Milioto19IROS,aksoy2020salsanet,razani2021lite,li2021multi,choy20194d,tang2020spvnas,zhu2020cylindrical},  object detection~\cite{yin2021center, Lang19CVPR, zhou2018voxelnet,yan2018second,liu2021iccv, pmlr-v205-peri23a, ma2023long, Peri_2023_ICCV, Peri_2022_CVPR}, and tracking~\cite{Teichman11ICRA,Moosmann13ICRA,Held14RSS}.

\PAR{Lidar panoptic segmentation.} Recently, \lps~\cite{behley2019iccv,Behley21icra,aygun21cvpr,fong21ral} has emerged as a holistic approach to Lidar-based dynamic scene understanding. Prior works~\cite{aygun21cvpr,hurtado2020mopt,zhou2021panoptic,hong2021lidar,razani2021gp,li2022panoptic,gasperini2020panoster,Agarwalla23iros,marcuzzi2023ral,marcuzzi2022contrastive, kreuzberg20224d, zhu20234d, marcuzzi2023mask4d,yilmaz2023mask4d} learn to group and classify points specified in the training data according to the target class vocabularies, while methods for open-set instance~\cite{Teichman11ICRA,Held16RSS,Moosmann13ICRA} and panoptic segmentation~\cite{wong2020identifying} rely on bottom-up grouping based on Euclidean distance between points. While these developments have been impressive, end-to-end methods remain limited to specific target classes that appear in the training data, while bottom-up, clustering-based approaches \cite{Held16RSS,wong2020identifying,Teichman11ICRA,Moosmann13ICRA,hu2020learning} are sensitive to a particular choice of clustering parameters and are unable to improve their performance in a data-driven fashion. 
In contrast, the \sal zero-shot model not only \textit{learns} to segment objects but is also capable of zero-shot classification.

\PAR{Self-supervised learning for Lidar.}
As labeled data in the Lidar domain is scarce, several works investigate how contrastive learning, proven effective in the image domain~\cite{he2020momentum,caron2020unsupervised}, can reduce the need for labeled data. 
SegContrast~\cite{nunes2022segcontrast} learns to align representations of point clouds and their augmented views, utilizing density-based clustering (DBSCAN~\cite{Ester96KDD}) to pool and contrast features. 
Rather than using DBSCAN,~\cite{sautier2022image,liu2023segment} leverage image-based methods~\cite{achanta2012slic} and foundation models~\cite{kirillov2023segment} to perform contrastive alignment between image and corresponding Lidar features.
The aforementioned methods only provide pretraining recipes for potential segmentation downstream tasks. Our \sal method goes a step further by training a full zero-shot \lps model that could benefit from such pre-training recipes.

\PAR{Self-supervised (3D) object detection.} Several methods (self-)supervise object detectors with (RGB-D) videos~\cite{Prest12CVPR,Osep18ICRA,Osep19ICRA,osep18ECCVW,pot2018self,harley2021track}, or Lidar sequences~\cite{najibi2022motion,zhang2023towards,seidenschwarz2024semoli} to discover objects in sensory streams. 
However, such motion-based supervision provides no semantic information. Therefore, \cite{najibi2023unsupervised} distills per-point (PCA quantized) CLIP~\cite{radford2021learning} features to Lidar for zero-shot classification of detected objects. 
Significantly different from~\cite{najibi2023unsupervised}, we tackle the more general and challenging task of LPS, which segments and classifies both moving and stationary \thing and \stuff classes. To this end, we distill CLIP features not per point but per object instance.
This provides a more holistic object-centric semantic distillation and allows us to supervise our model with non-quantized CLIP features. 

\PAR{Zero-shot recognition.} Zero-shot recognition (ZSR)~\cite{Xian18TPAMI} methods (recently also referred to as ``open-vocabulary recognition'') tackle the recognition of distinct, yet related, semantic concepts that are either not labeled in the training data (transductive) or not observed at all (inductive). 
Such methods assume a dataset with labels for \textit{some} classes, whereas unseen instances are supervised via attributes or class names. The latter is often used in conjunction with word embeddings~\cite{Mikolov13arxiv}. By learning to align image-language features, ZSR methods can infer class labels for unseen objects~\cite{radford2021learning}. 
ZSR has also been explored in the context of object detection, semantic segmentation, and panoptic segmentation~\cite{ding2023open, xu2023open} using word embedding models~\cite{Bansal18ECCV,miller18ICRA,Rahman18ACCV,bucher2019zero}, or CLIP vision-language embeddings~\cite{gu2021open, zareian2021open, zhong2022regionclip,li2022languagedriven,ghiasi2022scaling,rao2022denseclip, zhou2022maskclip, liang2023open, xu2023side}. 
In the context of semantic segmentation in RGB-D and Lidar data,~\cite{lu2023see, Peng2023OpenScene} augment image and language embeddings with features extracted from point clouds. 
Both rely on lifting dense image features from images to 3D during inference, which limits their applicability to setups with dense camera coverage. 
Similarly, \cite{takmaz2023openmask3d} segments instances in accumulated RGB-D point clouds (ScanNet~\cite{dai17cvpr}) and relies on manual supervision in conjunction with image-based dense CLIP features for zero-shot classification. 
Moreover, datasets such as ScanNet~\cite{dai17cvpr} were recorded with RGB-D sensors in \textit{static} environments and fused into 3D reconstructions based on scans, taken from multiple viewpoints. 
In contrast, \sal directly applies to \textit{full} Lidar point clouds while requiring no image features at the inference time. 

\begin{figure*}[t!]
\centering
\begin{subfigure}{.38\textwidth}
  \centering
  \includegraphics[trim={0 70px 550px 0},clip,width=\linewidth]{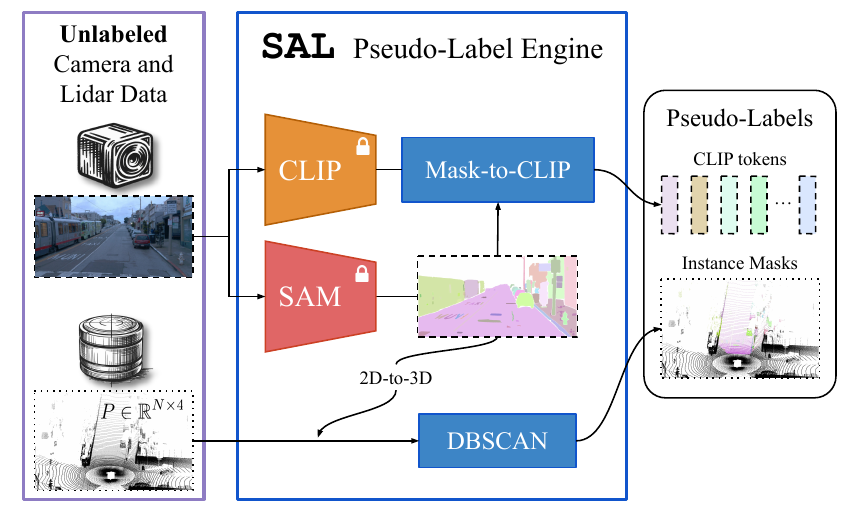}
  \caption{Pseudo-label engine}
  \label{fig:pseudo-engine}
\end{subfigure}%
\hspace{1mm}
\begin{subfigure}{.60\textwidth}
  \centering
  \includegraphics[trim={0 80px 330px 0},clip,width=\linewidth]{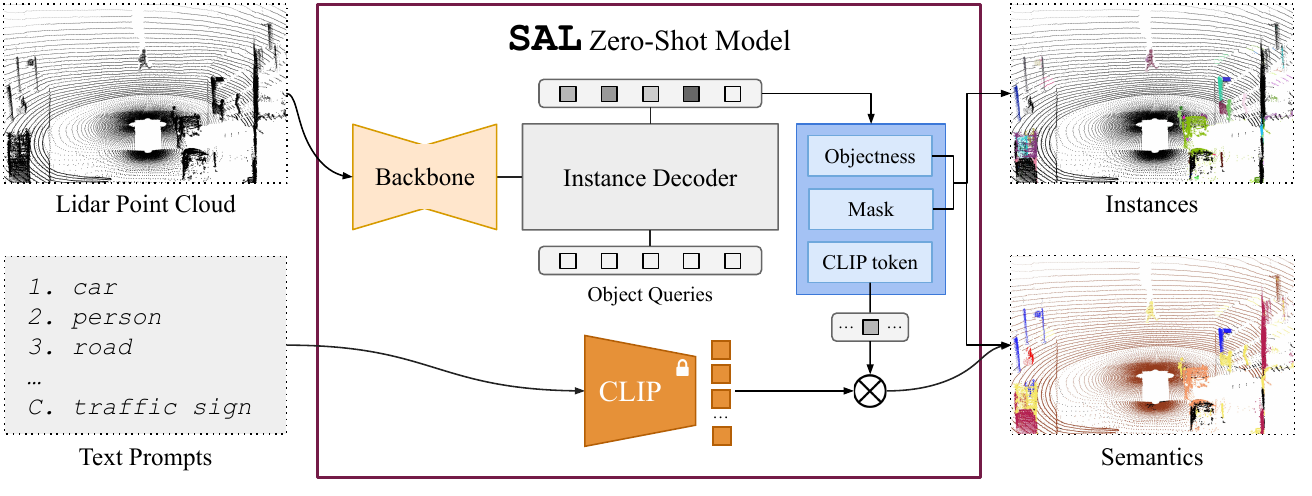}
  \caption{Zero-shot model}
  \label{fig:zero-shot-model}
\end{subfigure}
\caption{
Our \textbf{pseudo-label engine} (\cref{fig:pseudo-engine}) utilizes SAM~\cite{kirillov2023segment} to estimate segmentation masks in images, MaskCLIP~\cite{ding2023open} to estimate corresponding per-mask CLIP features, and a calibrated sensory setup to transfer them to the Lidar domain. 
We distill these pseudo-labels to our \textbf{zero-shot model} (\cref{fig:zero-shot-model}), which segments and classifies Lidar point clouds. The \sal zero-shot model employs a sparse-convolutional backbone~\cite{choy20194d}, followed by a Transformer decoder that predicts objectness scores, segmentation masks, and CLIP tokens for each query. To (optionally) perform zero-shot classification, we forward the dataset class vocabulary through the CLIP text encoder and match the encoded vocabulary with predicted CLIP tokens. Our model requires no retraining for different vocabularies and no image features at inference time. 
}
\label{fig:test}
\end{figure*}

\section{SAL: Segment Anything in Lidar}
\label{sec:method}

This section outlines the key challenges and components towards a Lidar model that segments and classifies any object. 
We first define the underlying problem formation in~\cref{sec:zero-shot_lps} and then present our (\texttt{\textbf{S}}egmenting \texttt{\textbf{A}}nything in \texttt{\textbf{L}}idar) \sal model in~\cref{sec:sal_framework}. 

\subsection{Zero-Shot Lidar Panoptic Segmentation}
\label{sec:zero-shot_lps}

\PAR{The task.} We discuss and evaluate \sal in the context of \lps (LPS), which tackles both instance segmentation and object recognition. LPS methods take as input an unordered point cloud $P \in \mathbb{R}^{N\times 4}$ encoding spatial coordinates and their sensor intensity. 
The currently established problem setting~\cite{Behley21icra, fong21ral} assumes supervision and task outputs as per-point semantic class and instance identity labels.
The class labels are confined to a pre-defined and fixed class vocabulary $\mathcal{V}$.
\textbf{In contrast}, we tackle LPS in a generalized, zero-shot setting, where the class vocabulary is only provided at inference, not training time. 
Such a vocabulary $\mathcal{V}$ specifies target classes as a list of $C$ class prompts.
Each prompt $c_i \in \mathcal{V}$ is specified via free-form text, \eg, with a class name and optional class description.
To perform zero-shot LPS a model must be designed to segment and classify any object in a Lidar scan.

\PAR{The challenge.} \textit{Where do we get the data to train such a model?}
Image-based models~\cite{kirillov2023segment,gu2021open, zareian2021open, zhong2022regionclip,ding2023open} rely on large and diverse datasets for pre-training~\cite{Deng09CVPR}, dense prediction~\cite{Lin14ECCV,Cordts16CVPR}, and foundation models that align images with textual descriptions~\cite{radford2021learning} -- a commodity not available in the Lidar domain. 

\subsection{\sal Overview}
\label{sec:sal_framework}
Our 2D-to-Lidar distillation method consists of two core components: 
(i) The \textbf{pseudo-label engine} transfers 2D vision foundation models into Lidar pseudo-labels using multi-modal inputs from a calibrated sensory setup, shown in \cref{fig:pseudo-engine}.
(ii) Our \textbf{zero-shot model} is trained on the generated pseudo-labels and is able to perform class-agnostic segmentation and zero-shot classification via text prompts. 
Given a semantic dataset vocabulary, this allows us to tackle zero-shot LPS as illustrated in~\cref{fig:zero-shot-model}. 

\subsection{\sal Pseudo-Label Engine}
\label{sec:pseudo_label_generation}

The \sal pseudo-label engine relies on a calibrated multi-modal sensory setup with $k\ge 1$ RGB cameras.
Furthermore, a sufficient overlap between the Lidar sensing area and each camera view is paramount.

\PAR{Mask generation:} To pseudo-label object instances, we utilize the \textit{Segment Anything} (SAM) model~\cite{kirillov2023segment}, which generates an overlapping set of segmentation masks. We flatten SAM's output mask hierarchy by non-maxima suppression (NMS) with a minimal overlap threshold to ensure mutually exclusive masks, suppressing object parts and subparts in favor of objects. This way, we obtain a set of non-overlapping binary masks $m_i^k \in \{ 0,1 \}^{W \times H}$ for each camera view $k$ with image plane of size $W \times H$.

\PAR{CLIP image token generation:} As shown by CLIP~\cite{radford2021learning}, aligning image and text features allows vision-language foundation models to perform zero-shot image classification based on arbitrary text prompts.
To transfer this capability to the Lidar domain, we generate a localized CLIP image feature token $f_i^k \in \mathbb{R}^{C_t}$ for each binary SAM mask $m_i^k$. To obtain image tokens for a masked region of the input image, we utilize~\cite{ding2023open} and their relative mask attention in the CLIP image encoder feature space. 
Note that we never classify segments in the image domain but merely distill CLIP image feature tokens to Lidar to facilitate zero-shot classification during inference where we do not use any image features.

\PAR{Image-to-Lidar unprojection:}
From the Lidar perspective, we unproject each image mask $m_i^k \in \{ 0,1 \}^{W \times H}$ to a binary Lidar segmentation mask $\tilde{m}_i \in \{0, 1 \}^N$ by transforming the respective camera coordinate frame to the Lidar space. For datasets with multiple cameras, such as nuScenes~\cite{fong21ral}, we process each image independently, followed by a cross-camera fusion of masks with a sufficient IoU overlap and averaging of their CLIP image tokens.
The unprojection yields pairs $\{\tilde{m}_i^k, f_i^k \}$ of Lidar masks and their corresponding CLIP features.

\begin{figure}[t]
    \centering

    \begin{subfigure}[b]{0.32\linewidth}
        \centering
        \includegraphics[width=\textwidth]{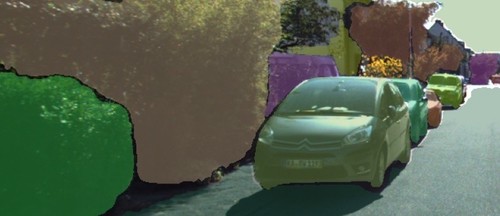}
        \caption{SAM masks}
        \label{fig:sam}
    \end{subfigure}
    \hfill %
    \begin{subfigure}[b]{0.32\linewidth}
        \centering
        \includegraphics[width=\textwidth]{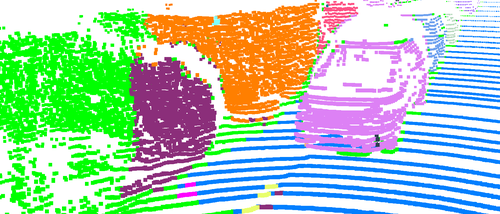}
        \caption{SAM unprojected}
        \label{fig:sam_unproj}
    \end{subfigure}
    \hfill %
    \begin{subfigure}[b]{0.32\linewidth}
        \centering
        \includegraphics[width=\textwidth]{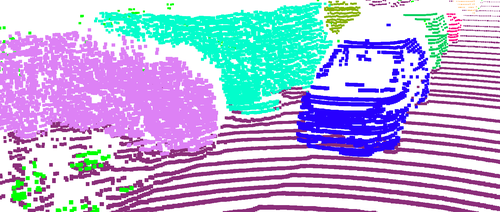}
        \caption{SAM + DBSCAN}
        \label{fig:sam_unproj_refined}
    \end{subfigure}
    \caption{\textbf{Refinement via clustering.} After transferring image masks (\cref{fig:sam}) to Lidar (\cref{fig:sam_unproj}), we obtain pseudo-labels that suffer from sensory misalignment-related issues. Our geometric refinement (\cref{fig:sam_unproj_refined}) improves localization.}
    \label{fig:geometric_refinement}
\end{figure}

\PAR{Refinement via density-based clustering:}
Image-to-Lidar transformation (\cref{fig:sam}$\to$\cref{fig:sam_unproj}) is inherently noisy due to imperfect calibration and issues with synchronization and Lidar rolling shutter. 
We improve our pseudo-label quality by creating an ensemble of DBSCAN~\cite{Ester96KDD} clusters $\tilde{m}_l \in \{ 0,1 \}^{N}$, obtained by varying the density threshold to compensate for varying density in Lidar point clouds \arxiv{(details in \cref{sec:impl_design_details_label_engine_app}).}\conf{(details in the appendix).} %
We replace each $\tilde{m}_i$ with its best-matching $\tilde{m}_l$ in case their IoU exceeds a minimal overlap threshold and retain the original mask otherwise to obtain a refined set of pseudo-labels (\cref{fig:sam_unproj_refined}) that retain their original cardinality and associated CLIP features.

\subsection{\sal Zero-Shot Model}
\label{sec:network}
The universal \sal zero-shot model deconstructs LPS into class-agnostic segmentation and zero-shot classification via text prompts.

\PAR{Universal architecture.} 
Instead of relying on highly specialized and engineered LPS models, we base the \sal model on a universal Transformer decoder architecture (\cref{fig:zero-shot-model}), similar to~\cite{marcuzzi2023ral}. 
Its 2D counterparts~\cite{carion2020end, cheng2022masked} provide top-performing results across semantic, instance, and panoptic segmentation for images.
With increasing amounts of (pseudo-)labeled data, Transformer decoders have the potential to achieve a similar impact in the Lidar domain.
Following~\cite{marcuzzi2023ral}, we deploy a Minkowski \mbox{U-Net}~\cite{choy20194d} backbone for feature extraction followed by a Transformer decoder architecture with object query to point/voxel feature cross-attention.
Training our model to segment and classify \emph{any} object and perform the evaluation in a zero-shot setting requires a unique design of the final task heads. 

\PAR{Class-agnostic segmentation.}
To localize segments in a point cloud, we rely on two model heads that predict segmentation masks and objectness scores for each query.
The architecture of the former follows~\cite{marcuzzi2023ral} and predicts binary output masks $\hat{m}_i \in \{0, 1\}^N$ by computing the dot product between queries and point features.
The objectness head reduces the multi-class problem to a binary object or no-object decision. 

\PAR{Zero-shot classification.} 
To equip our model with zero-shot classification capabilities without relying on any image input, we learn to predict CLIP~\cite{radford2021learning} Lidar tokens, \ie, features in the CLIP space obtained from Lidar inputs, for each query.
The token head consists of a three-layer MLP that directly regresses tokens $\hat{f}_i^j$.
At inference, the predicted tokens are matched to text prompts via the CLIP text encoder.
The matching yields a probability distribution over the prompts.
To perform zero-shot LPS on a pre-defined dataset, we use its class vocabulary as input text prompts (\cref{fig:zero-shot-model}).

\label{sec:training-sal}
We train our model jointly for both class-agnostic segmentation and zero-shot classification with the following loss:
\begin{equation}
    \label{eq:model_loss}
    \mathcal{L}_{\text{\sal}} = \mathcal{L}_{obj} + \mathcal{L}_{seg} + \mathcal{L}_{token},
\end{equation}
with binary cross-entropy loss $\mathcal{L}_{obj}$ and a cosine distance loss $\mathcal{L}_{token}$.
The segmentation loss $\mathcal{L}_{seg}$ follows~\cite{marcuzzi2023ral} and consists of a binary mask cross-entropy and dice loss.
The $\mathcal{L}_{\text{\sal}}$ loss is supervised by pseudo-label pairs $\{\tilde{m}_i^k, f_i^k \}$ obtained from our label engine and thereby distills both 2D foundation models (SAM and CLIP) into our LPS model. 

\PAR{How to train with partial labels?}
Partially (pseudo-)labeled point clouds present a challenge for \sal model training. For example, in SemanticKITTI, only $14\%$ of points are pseudo-labeled due to low camera coverage (see \arxiv{\cref{tab:pseudo_label_stats}}\conf{appendix}). 
If we naively train \sal on partially (pseudo-)labeled point clouds, the objectness loss $\mathcal{L}_{obj}$ would penalize any segmentation in these regions as a false positive, thereby teaching the model to ignore these points entirely. Empirically, we determined that the most effective training strategy is to (i) remove unlabeled points from point clouds, (ii) utilize standard data augmentations (rotations, flipping, scaling, and translations), in conjunction with proposed \textit{FrankenFrustum} augmentation (detailed in \arxiv{\cref{sec:training-sal-partial-labels}}\conf{the appendix}) that mimics fully-labeled point clouds during training by randomly removing unlabeled points and replicating labeled frustum regions around the vertical axis. This augmentation does not increase the overall label coverage, but our ablations in~\cref{sec:ablations} show its effectiveness in reducing the domain gap between training and inference input clouds. 

\section{Experiments}
\label{sec:experiments}

In the following section, we outline our experimental setup (\cref{sec:experimental_setup}) used to ablate the \sal model and its training strategies (\cref{sec:ablations}). 
Since \sal presents the first \textit{Zero-Shot} \lps (ZS-LPS) model, we design and compare with multiple hand-crafted baselines that implement concepts from related works. 
We conclude our analysis by comparing our zero-shot performance (\cref{sec:benchmarks}) to fully-supervised LPS methods.

\subsection{Experimental Setup}
\label{sec:experimental_setup}

\subsubsection{Datasets.}
We evaluate \sal on two public Lidar Panoptic Segmentation (LPS) datasets, SemanticKITTI~\cite{behley2019iccv, Behley21icra} and nuScenes~\cite{fong21ral}. 
We utilize provided, human-annotated ground-truth (GT) panoptic segmentation labels solely for evaluation purposes, linear probing ablations, and baselines. We train \sal model \textit{only} our pseudo-labels (\cref{sec:pseudo_label_generation}). 
To demonstrate the general applicability of \sal to other domains and datasets, we further show qualitative results of \sal trained on Waymo Open~\cite{sun20CVPR} dataset, which contains \textit{no} GT LPS labels.

\PAR{Metrics.} 
For the evaluation, we utilize the standard Panoptic Segmentation~\cite{Kirillov2018PanopticS} metrics. The Panoptic Quality $PQ = RQ \times SQ$ combines Segmentation Quality ($SQ$) and Recognition Quality ($RQ$), \ie, F-1 score. True positives (TPs) are mask predictions with sufficient intersection-over-union (IoU) overlap with GT masks of the same class.   

\PAR{Semantic Oracle (SO).} The aforementioned metrics assume GT labels and predictions with instance IDs and semantic classes for each point. Therefore, they are not suitable for evaluating class-agnostic segmentation. 
To assess the class-agnostic segmentation independently of semantics, we apply SO during the evaluation by assigning ground truth semantic classes $c_i$ to predicted masks $m_i$ via majority voting.
Evaluation with SO is only needed for evaluation of class-agnostic segmentation -- our zero-shot models provide both instance \textit{and} semantic class predictions.

\PAR{Stuff Merging (SM).} 
SO allows us to asses class-agnostic \textit{instance} segmentation for \thing classes. However, this approach is unsuitable for assessing the performance of \stuff classes, as existing Lidar datasets merge instances of \stuff classes into a single instance.
The \thing-
\stuff separation can be unintuitive and inconsistent across datasets (\eg, SemanticKITTI merges instances of traffic signs into a single \stuff class). In contrast, we fully embrace the philosophy that \textit{all} classes can be segmented into individual instances (\eg, segmentation models should localize individual trees/bushes in the \texttt{vegetation} class or distinguish individual \texttt{buildings}).
To evaluate our models on the target datasets in the SO regime on Lidar datasets that do not provide instance-level annotations for all classes, we additionally report results using a \textit{merge} strategy (SM), which merges all instances of a particular \stuff class into a single mask to ensure outputs of our models are consistent with the format of target datasets. 

\subsection{\sal Ablations}
\label{sec:ablations}

In this section, we discuss \textit{how} to train \sal zero-shot segmentation model using self-generated pseudo-labels (\cref{tab:ablation_franken_frustum}) and discuss design decisions behind our instance (\cref{tab:ablation_num_queries}) and semantic distillation (\cref{tab:ablation_lin_probe_clip_semantics}). 

\PAR{Learning with partial labels.} 
For the single-camera setup~\cite{behley2019iccv, Behley21icra}, our pseudo-labels only cover $14\%$ of the input point clouds (see \arxiv{\cref{tab:pseudo_label_stats}}\conf{appendix}). Training on partial label coverage presents a challenge for our model. 

In~\cref{tab:ablation_franken_frustum}, we only evaluate class-agnostic segmentation by applying SO and SM (see~\cref{sec:experimental_setup}).
We compare \sal models trained using \textit{pseudo} and \textit{GT} labels.
The \textit{Frustum Filter} removes all unlabeled points not visible in the camera frustum during training and/or evaluation.
As shown in \cref{tab:ablation_franken_frustum}, removing unlabeled points from pseudo-labeled point clouds during training shows a \textit{significant} benefit on the performance ($3^{rd}$ row, $59.3$ PQ) as compared to the variant, where we merely ignore unlabeled region during the training ($2^{nd}$ row, $22.2$ PQ). 
This is likely due to data imbalance in a single-camera setup, leading to a class imbalance between labeled/unlabeled points, incentivizing \sal to suppress predictions.

While utilizing standard augmentations (rotations, flipping, scaling, and translations) is crucial, we observe that performing our \mbox{\textit{FrankenFrustum}} augmentation, which concatenates the visible portion of the point cloud around the z-axis (see~\cref{sec:training-sal}), significantly improves results ($4^{th}$ row, $62.5$ PQ).
Mixing point clouds from different scans ($5^{th}$ row, $62.8$ PQ) further boosts results.
Overall, we obtain $69$ PQ from GT supervision ($1st$ row) and $62.8$ PQ when using our pseudo-labels ($5th$ row). Remarkably, our labels yield $91\%$ of the GT performance while covering only $14\%$ of the input point cloud (see\arxiv{~\cref{tab:pseudo_label_stats}}\conf{\;appendix}). 
To further contextualize these results, we note that this evaluation is performed \textit{only} on classes for which GT instance labels are available. As shown in~\cref{fig:teaser}, our model learns to segment a much larger variety of classes. 
Evaluating only the subset of points visible in the camera ($14\%$ of all points, rows $6\&7$), the gap between the GT-supervised model ($71.8$ PQ) and pseudo-supervised model ($70.7$ PQ) shrinks further. 
\textit{We conclude that by training \sal model using pseudo-labels, we distill the notion of objectness from SAM~\cite{kirillov2023segment} into our model and obtain segmentation capabilities, similar to the model trained on GT data.}

\begin{table}[ht]
    \centering
    \scriptsize
    \begin{minipage}[t]{0.53\linewidth} %
        \centering
        \caption{
        \textbf{
        Class-agnostic segmentation.
        } 
        By cropping unlabeled points and performing data augmentations (in combination with our FrankenFrustum), \sal successfully learns to segment full point clouds even when only $14\%$ of points are (pseudo)-labeled.
        }
        \resizebox{\textwidth}{!}{%
        \begin{tabular}{cZ|cc|cHHZ|cHHc| c H H  c H H Z}
            \toprule
            \multirow{2}{*}{Labels} & & \multicolumn{2}{c|}{Frust.-Filter} &  Franken & & & & \multirow{2}{*}{PQ} & \multirow{2}{*}{PQ\textsuperscript{\textdagger}} & \multirow{2}{*}{RQ} & \multirow{2}{*}{SQ} & \multirow{2}{*}{PQ\textsuperscript{Th}} & \multirow{2}{*}{RQ\textsuperscript{Th}} & \multirow{2}{*}{SQ\textsuperscript{Th}} & \multirow{2}{*}{PQ\textsuperscript{St}} & \multirow{2}{*}{RQ\textsuperscript{St}} & \multirow{2}{*}{SQ\textsuperscript{St}} & \multirow{2}{*}{mIoU} \\
             & Eval & Train & Eval &  Frust. & \# Pts  & \# Queries / \# Inst. & FrankenFrustum &  &  &  &  &  & &  & &  & \\
            \midrule
            GT & GT & &  & & 80000 &  300 / All & & $69.0$ & $74.2$ & $80.1$ & $83.5$ & $81.6$ & $89.6$ & $90.9$ & $59.7$ & $73.3$ & $78.1$ & $73.9$  \\
            \midrule
            Pseu. & GT &  &         &  & ?  & 300 / All &             & $22.2$ & $26.6$ & $27.0$ & $67.5$ & $45.1$ & $53.2$ & $84.4$ & $5.5$ & $7.9$ & $55.2$ & $16.7$ \\
            Pseu. & GT & \checkmark &  & & ?  & 300 / All &             & $59.3$ & $66.5$ & $74.4$ & $78.2$ & $65.8$ & $77.2$ & $85.0$ & $54.5$ & $72.4$ & $73.2$ & $71.2$ \\
            \midrule
            Pseu. & GT  & \checkmark  && \checkmark  & ? & 300 / 100 & \checkmark & $62.5$ & $69.2$ & $77.8$ & $79.1$ & $67.6$ & $79.1$ & $85.1$ & $58.8$ & $76.8$ & $74.8$ & $76.0$ \\
            Pseu. & GT  & \checkmark\hphantom & & \checkmark (mix)\hphantom & ? & 300 / 100 & \checkmark (mix)\hphantom & $62.8$ & $69.3$ & $78.1$ & $79.0$ & $69.0$ & $80.4$ & $85.3$ & $58.3$ & $76.5$ & $74.3$ & $75.7$ \\
            \midrule
            GT & GT & & \checkmark &  & 80000 &  300 / All & & $71.8$ & $74.7$ & $83.0$ & $84.8$ & $84.6$ & $93.5$ & $90.2$ & $62.5$ & $75.4$ & $80.8$ & $73.0$ \\
            Pseu. & GT  & \checkmark & \checkmark & \checkmark (mix) & ? & 300 / 100 & \checkmark (mix)\hphantom  & $70.7$ & $74.5$ & $85.6$ & $81.9$ & $75.4$ & $87.1$ & $86.4$ & $67.3$ & $84.4$ & $78.7$ & $79.7$ \\
            \bottomrule
        \end{tabular}
        }
        \label{tab:ablation_franken_frustum}
    \end{minipage}
    \hfill
    \begin{minipage}[t]{0.45\linewidth} %
        \centering
        \caption{
        \textbf{Scaling.}
        By contrast to GT data/labels, by increasing the number of queries, \sal improves performance on \stuff classes. By increasing the amount of labeled data, we further improve segmentation performance.
        }
        \resizebox{\textwidth}{!}{%
        \begin{tabular}{ccZ|HHcZ| cHHc | c H H  c H H Z}
            \toprule
            \multicolumn{2}{c}{Train} & Eval & Train Frustum Filter & \# Pts  & \multirow{2}{*}{\shortstack[x]{Num.\\Quer.}} & \multirow{2}{*}{FrankenLidar} & \multirow{2}{*}{PQ} & \multirow{2}{*}{PQ\textsuperscript{\textdagger}} & \multirow{2}{*}{RQ} & \multirow{2}{*}{SQ} & \multirow{2}{*}{PQ\textsuperscript{Th}} & \multirow{2}{*}{RQ\textsuperscript{Th}} & \multirow{2}{*}{SQ\textsuperscript{Th}} & \multirow{2}{*}{PQ\textsuperscript{St}} & \multirow{2}{*}{RQ\textsuperscript{St}} & \multirow{2}{*}{SQ\textsuperscript{St}} & \multirow{2}{*}{mIoU} \\
            Labels & Set &  &  &  & & & & &  &  && &  &&\\
            \midrule
            GT & train & GT & & 80000 & 100 &  & $67.8$ & $72.8$ & $79.6$ & $82.4$ & $79.7$ & $88.8$ & $89.4$ & $59.2$ & $73.0$ & $77.3$ & $74.2$ \\
            GT & train & GT & & 80000 & 200 &  & $67.9$ & $73.3$ & $78.9$ & $83.6$ & $79.8$ & $87.7$ & $90.5$ & $59.3$ & $72.5$ & $78.6$ & $75.0$ \\
            GT & train & GT &  & 80000 &  300 & & $69.0$ & $74.2$ & $80.1$ & $83.5$ & $81.6$ & $89.6$ & $90.9$ & $59.7$ & $73.3$ & $78.1$ & $73.9$  \\

            \midrule
            Pseu. & train & GT & \checkmark & 80000 & 100 & \checkmark  (mix) & $53.9$ & $60.8$ & $69.6$ & $75.3$ & $62.2$ & $75.4$ & $81.2$ & $47.8$ & $65.4$ & $71.1$ & $66.2$ \\
            Pseu. & train & GT & \checkmark & 80000 & 200 & \checkmark  (mix) & $60.8$ & $67.5$ & $77.3$ & $77.0$ & $66.7$ & $79.9$ & $82.7$ & $56.5$ & $75.5$ & $72.9$ & $73.9$ \\
            Pseu. & train & GT & \checkmark & X & 300 & \checkmark (mix) & $62.8$ & $69.3$ & $78.1$ & $79.0$ & $69.0$ & $80.4$ & $85.3$ & $58.3$ & $76.5$ & $74.3$ & $75.7$ \\
            \midrule
            Pseu.& bigtrain & GT & \checkmark & X & 300 & \checkmark (mix) & 65.3 & X & X & 79.5 & 71.9 & X & X & 60.5 & X & X & X \\
            \bottomrule
        \end{tabular}
        }
        \label{tab:ablation_num_queries}
    \end{minipage}
    \label{tab:ablations_combined}
\end{table}

\PAR{Scaling queries and data.}
Analogous to the grid inference of SAM, which operates with hundreds of spatial queries (prompts), we analyze the effect of different numbers of decoder queries. We subsample instances during training to train models with fewer queries than the maximum number of segments per scan. 
As visualized in~\cref{tab:ablation_num_queries}, increasing the number of decoder queries improves the recognition of \thing for both training with pseudo and GT labels. 
However, we observe a significant performance boost with \textit{pseudo} labels for \stuff classes ($58.3$ PQ$^{St}$, $+10.8$), while the improvement for GT labels is marginal ($59.7$ PQ$^{St}$, $+0.5$). Our model benefits from a larger number of queries for \stuff classes as it learns a fine-grained segmentation model that is not limited to a prefixed set of \thing classes.
In the last row, we concatenate pseudo-labeled \textit{train} and \textit{test} sets (neither used for the validation) into a \textit{bigtrain} set and observe an improvement of $+2.5$ PQ, suggesting that scaling the amount of training data has the potential to improve the performance of our models further.

\setlength{\intextsep}{1pt}
\begin{wraptable}{l}{0.42\textwidth}
  \centering
  \scriptsize
  \caption{
  \textbf{Semantic distillation.}
  We linearly probe the \sal model, trained with and without semantic distillation loss $\mathcal{L}_{token}$. We train a linear classifier with GT labels while keeping backbone and decoder features frozen. 
  As can be seen, $\mathcal{L}_{token}$ successfully distills a notion of semantics into our model. 
  }
  \resizebox{0.42\textwidth}{!}{%
      \begin{tabular}{cc|cccc}
      \toprule
      Linear prob. & $\mathcal{L}_{token}$ & PQ & RQ & SQ & mIoU \\
      \midrule

      $\times$    &          & $20.0$ & $26.3$ & $55.2$ & $23.4$ \\

      $\times$    & $\times$ & $33.1$ & $41.9$ & $68.3$ & $40.0$ \\
     
      \midrule
      
      & $\times$   & $24.8$ & $32.3$ & $66.8$ & $29.7$ \\
      \bottomrule
      \end{tabular}
  }
  \label{tab:ablation_lin_probe_clip_semantics}
\end{wraptable}

\PAR{Semantic distillation.} At inference, \sal predicts a binary segmentation mask and objectness score for each query. To perform zero-shot classification, our model additionally predicts CLIP image tokens. This allows us to not rely on image features during inference but prompt our predicted tokens with arbitrary texts, \ie, object classes. 
\textit{Does the token head loss $\mathcal{L}_{token}$ successfully distill a notion of object semantics from CLIP into our model?} To verify this, we perform \textit{linear probing} experiments: we replace the token distillation head of a trained model with a linear classifier (backbone and instance decoder remain frozen) and train it using GT labels. 
As can be seen in \cref{tab:ablation_lin_probe_clip_semantics}, distilling CLIP features into our model \textit{significantly} improves linear probing performance
($33.1$ PQ, $40$ mIoU) compared to the baseline, not trained with the semantic distillation loss $\mathcal{L}_{token}$ ($20.0$ PQ, $23.4$ mIoU). This confirms that by distilling CLIP features, we inject a notion of semantics into our model, even though our model is not explicitly supervised with any labels containing semantic information -- the notion of semantics is learned implicitly via CLIP feature distillation. Without this loss, our zero-shot model 
($24.8$ PQ, $29.7$ mIoU) outperforms the linearly probed model tuned using GT semantic labels.

\subsection{Class-Agnostic and Zero-Shot Lidar Panoptic Segmentation}

As the first study tackling \textit{Zero-Shot Lidar Panoptic Segmentation} (ZS-LPS), we devise several strong baselines inspired by prior works in RGB-D semantic segmentation. During inference, these methods utilize image-based models and lift image features to 3D~\cite{Peng2023OpenScene}. 
To ensure these are evaluated fairly, we report results in \cref{tab:zero_shot_baselines} on the subset of points visible in at least one camera (\textit{Frust. Eval.}). Our \sal model does not have this limitation. Therefore, we additionally report results evaluated on \textit{full} point clouds in the bottom of \cref{tab:zero_shot_baselines}.

\PAR{Class-agnostic segmentation.} Our first baseline (SAM~\cite{kirillov2023segment}) generates masks in images and unprojects them to Lidar (see \cref{sec:pseudo_label_generation}), and leads to $46$ PQ.
We observe bleeding edges (see~\cref{fig:sam_unproj}) after unprojecting from the cameras to Lidar caused by imperfect calibrations. To mitigate these artifacts, we experiment with slightly eroding SAM predictions in the image domain. However, this leads to a performance drop ($42.2$ PQ), likely because smaller segments after erosion fail to pass $>0.5$ overlap with labeled instances.

\begin{table*}[t]
    \centering
    \scriptsize
    \caption{\textbf{Zero-shot panoptic segmentation.}
    We utilize prior efforts in the image domain~\cite{kirillov2023segment} and Lidar~\cite{Peng2023OpenScene} domain to craft multiple baselines that only unproject segmentation masks and lift image features to Lidar. 
    By contrast, \sal distills outputs of such baselines (pseudo-labels) into a stronger Lidar segmentation model. With \textit{Image Feat.} we denote methods that require image features at inference time, and \textit{Frust. Eval.} denotes the evaluation of a subset of points visible in the camera.
    }
    \resizebox{\textwidth}{!}{%
    \begin{tabular}{lcc|cc|cc|c||cc|cc|c}
    \toprule
        
        \multirow{2}{*}{Method} & Frust. & Image & \multicolumn{5}{c||}{Default classes} & \multicolumn{5}{c}{Super classes} \\ %
        & Eval & Feat. &
        PQ & SQ & 
        PQ\textsuperscript{Th} & PQ\textsuperscript{St} & 
        mIoU & 
        PQ & SQ & 
        PQ\textsuperscript{Th} &  PQ\textsuperscript{St} & 
        mIoU \\
        \midrule
        \multicolumn{13}{c}{Class-agnostic Segmentation (Semantic Oracle)}\\
        \midrule
        SAM  & \checkmark & \checkmark & $46.0$ & $72.1$ & $49.7$ & $43.4$ & -- & --  &  -- & -- & -- & -- \\
        SAM+Erosion & \checkmark  & \checkmark & $42.2$ & $69.4$ & $45.6$ & $39.6$ & -- & --  & --  & -- & -- & -- \\
        \midrule

        SAM+DBS (filter) & \checkmark & \checkmark & $46.7$ & $70.3$ & $76.8$ & $24.8$ & -- &  -- & --  & -- & -- & -- \\
        SAM+DBS (replace) & \checkmark & \checkmark & $48.7$ & $73.7$ & $53.1$ & $45.4$ & -- & --  & --  & -- & -- & -- \\
        \midrule
        \sal  & \checkmark & \crossmark & $70.7$ & $81.9$ & $75.4$ & $67.3$ & -- & --  & --  & -- & -- & -- \\
        \midrule
        \midrule
        \multicolumn{13}{c}{Zero-Shot Lidar Panoptic Segmentation} \\
        \midrule

        SAM+DBS+CLIP & \checkmark & \checkmark & $27.5$ & $71.5$ & $31.7$ & $24.5$ & $30.6$ & $51.1$ & $77.5$ & $71.2$ & $41.0$ & $54.3$ \\

        \sal & \checkmark & \crossmark & $33.1$ & $71.4$ & $22.8$ & $40.5$ & $33.5$ & $63.9$ & $84.2$ & $88.3$ & $51.7$ & $66.4$ \\
        
        \midrule

        SAM+DBS+CLIP & \crossmark & \crossmark & $8.2$ & $56.4$ & $18.6$ & $0.6$ & $7.5$ & $11.5$ & $47.6$ & $0.0$ & $17.3$ & $11.2$ \\

        \sal & \crossmark & \crossmark & $24.8$ & $66.8$ & $17.4$ & $30.2$ & $28.7$ & $48.5$ & $78.8$ & $80.4$ & $32.6$ & $52.8$ \\
    \bottomrule
    \end{tabular}
    }
    \label{tab:zero_shot_baselines}
\end{table*}

\PARit{Density-based filtering:} Alternatively, to mitigate unprojection errors, we experiment with density-based clustering methods. We generate a large pool of DBSCAN (DBS) clusters and test two strategies to improve our SAM-based segmentation labels: (i) \emph{filtering} segments without a sufficient overlap with any DBSCAN cluster or (ii) \emph{replacing} segments with large overlaps. 

\PARit{SAM\!+\!DBS (filter):} Intuitively, the filter approach removed Lidar segments only segmentable in the image domain, \eg, a SAM mask that segments a shadow on a flat wall. While this strategy improves precision for \thing classes ($76.8$ PQ$^{Th}$, $+27.1$), it significantly degrades performance on \stuff classes ($24.8$ PQ$^{St}$ $-18.6$) and overall degrades PQ to $46.7$. 

\PARit{SAM\!+\!DBS (replace):} The replace strategy (that replaces SAM masks with DBSCAN clusters with sufficient mutual overlap), on the other hand, improves both \thing and \stuff classes to an overall PQ of $48.7$ ($+3.9$) and and SQ $73.7$ ($+1.1$), respectively. This improvement can be visually verified in~\cref{fig:sam_unproj_refined}. 

\PARit{\sal:} We train our \sal model using top-performing \textit{SAM\!+\!DBS (replace)} and observe a significant improvement ($70.7$ PQ) in terms of both \thing and \stuff classes. Even though \sal obtains significantly higher PQ than hand-crafted baselines, it also improves in terms of SQ ($81.9$ \vs $73.7$ closest competitor), suggesting that the distilled model is robust to projection artifacts. 

\PAR{Do we need a vision segmentation foundation model?} To generate a set with a sufficiently high recall, we need to cluster points with varying DBSCAN radius parameters (details in \arxiv{\cref{sec:impl_design_details_label_engine_app}}\conf{the appendix}). This leads to a large set of predicted masks per scan ($5,413$ avg.). While most of the \textit{correct} segments are in this set, separating the signal from the noise is difficult. Prior works, such as~\cite{najibi2022motion}, are limited to \texttt{moving thing} classes as they rely on motion cues to filter static point cloud regions and DBSCAN to group remaining points. Our core insight is that we can utilize vision foundation models that already learned a notion of objectness: SAM generates \textit{only} $171$ masks per image, reduced to $45$ via flattening, and to $39$ masks after the unprojection, achieving a high recall with \textit{only a few instances}, or in other words, a high signal-to-noise ratio.

\PAR{Zero-shot segmentation.} So far, we discussed class-agnostic segmentation. To obtain semantic class predictions per mask in a zero-shot fashion, we devise a simple baseline inspired by \cite{Peng2023OpenScene}. Once masks are unprojected from the image to Lidar space, we utilize associated CLIP features (directly extracted from images using~\cite{ding2023open}) to perform zero-shot classification via dot product between encoded class prompts and (lifted) CLIP features~\cite{radford2021learning}. \textit{This baseline (\textit{SAM\!+\!DBS\!+\!CLIP}) can be understood as an approach that directly lifts image features from state-of-the-art image-based models~\cite{ding2023open} to Lidar.} We detail prompts in \arxiv{\cref{sec:txt_prompt_eng_app}}\conf{the appendix}. 

There is a significant performance drop between \textit{SAM\!+\!DBS (replace)} baseline, evaluated with Semantic Oracle ($5^{th}$ row, $48.7$ PQ), and zero-shot baseline (\textit{SAM\!+\!DBS (replace)\!+\!CLIP}), $7^{th}$ row, $27.5$ PQ.
This is not surprising, as zero-shot segmentation is a challenging problem in both image~\cite{ding2023open} and Lidar domains. To this end, even when transferring features of state-of-the-art image-based models to Lidar we observe a significant performance drop. Remarkably, \sal distills such noisy image-based features to a stronger model ($8^{th}$ row, $33.1$ PQ, $+5.6$ PQ).
Finally, to emphasize the inherent limitation of baselines, relying on image features, we evaluate on full $360^{\circ}$ point clouds. \sal performs similarly in both settings. The baseline, on the other hand, relies on image features and is, therefore, unable to predict labels out of the camera frustum. This problem can be mitigated by utilizing a setup with denser camera coverage. 
\PARit{Evaluation on super-classes.} We observe that the CLIP classifier often confuses related classes, \eg, \texttt{car} and \texttt{other vehicle}. We additionally evaluate our semantics by prompting with super-classes (as defined in~\cite{behley2019iccv}). This alone increases PQ to 
$63.9$ ($48.5$ when evaluating on full point clouds) and suggests significant potential for improving prompting and text-to-image alignment. 

\begin{table*}[t]
    \centering
    \scriptsize
    \caption{
        \textbf{Lidar Panoptic Segmentation (LPS) on SemanticKITTI and nuScenes validation sets.}
        We prompt our zero-shot \sal model with the respective class vocabularies and compare its performance to fully-supervised baselines.
        On SemanticKITTI and nuScenes, we reach $42 \%$ and $54 \%$ of the fully-supervised model, respectively.
        This gap reduces significantly when we evaluate super classes. 
    } 
    \begin{tabular}{cll|cHcc|cHHcHZc||cHccHHHHHHc HHHHHHHHHHZ}
    \toprule
        
        & \multirow{2}{*}{Method} & \multirow{2}{*}{Supervision} & \multicolumn{11}{c||}{Default classes} & \multicolumn{11}{c}{Super classes} & \\

        \cmidrule{4-14} 
        \cmidrule{14-36}

        &  &  & PQ & PQ\textsuperscript{\textdagger} & RQ & SQ & PQ\textsuperscript{Th} & RQ\textsuperscript{Th} & SQ\textsuperscript{Th} & PQ\textsuperscript{St} & RQ\textsuperscript{St} & SQ\textsuperscript{St} & mIoU & PQ & PQ\textsuperscript{\textdagger} & RQ & SQ & PQ\textsuperscript{Th} & RQ\textsuperscript{Th} & SQ\textsuperscript{Th} & PQ\textsuperscript{St} & RQ\textsuperscript{St} & SQ\textsuperscript{St} & mIoU & PQ & PQ\textsuperscript{\textdagger} & RQ & SQ & PQ\textsuperscript{Th} & RQ\textsuperscript{Th} & SQ\textsuperscript{Th} & PQ\textsuperscript{St} & RQ\textsuperscript{St} & SQ\textsuperscript{St} & mIoU\\
        \midrule
        \multirow{7}{*}{\rotatebox[origin=c]{90}{SemanticKITTI}}
        & DS-Net~\cite{hong2021lidar} & Full & $57.7$ & $63.4$ & $68.0$ & $77.6$ & $61.8$ & $68.8$ & $78.2$ & $54.8$ & $67.3$ & $77.1$ & $63.5$ & -- & -- & -- & -- & -- & -- & -- & -- & -- & -- & -- & -- & -- & -- & -- & -- & -- & -- & -- & -- & -- & --\\ 
        & PolarSeg~\cite{zhou2021panoptic} & Full & $59.1$ & $64.1$ & $70.2$ & $78.3$ & $65.7$ & $74.7$ & $87.4$ & $54.3$ & $66.9$ & $71.6$ & $64.5$ & -- & -- & -- & -- & -- & -- & -- & -- & -- & -- & -- & -- & -- & -- & -- & -- & -- & -- & -- & -- & -- & --\\ 
        & EfficientLPS~\cite{sirohi2021efficientlps}  & Full & $59.2$ & $65.1$ & $69.8$ & $75.0$ & $58.0$ & $68.2$ & $78.0$ & $60.9$& $71.0$ & $72.8$ & $64.9$ & -- & -- & -- & -- & -- & -- & -- & -- & -- & -- & -- & -- & -- & -- & -- & -- & -- & -- & -- & -- & -- & --\\ 
        & GP-S3Net~\cite{razani2021gp} & Full & $63.3$ & $71.5$ & $75.9$ & $81.4$ & $70.2$ & $80.1$ & $86.2$ & $58.3$ & $72.9$ & $77.9$ & $73.0$ & -- & -- & -- & -- & -- & -- & -- & -- & -- & -- & -- & -- & -- & -- & -- & -- & -- & -- & -- & -- & -- & --\\ 
        & MaskPLS~\cite{marcuzzi2023ral} & Full & $59.8$ & -- & $69.0$ & $76.3$ & -- & -- & -- & -- & -- & -- & -- & $78.4$ & $84.3$ & $87.1$ & $88.2$ & $93.6$ & $99.8$ & $93.7$ & $70.8$ & $80.6$ & $85.5$ & $84.5$ & $68.2$ & $73.3$ & $79.3$ & $83.4$ & $79.6$ & $87.3$ & $90.9$ & $59.9$ & $73.4$ & $78.0$ & $73.3$\\
        \cmidrule{2-36}
        & \sal  & Full & $59.5$ & $63.9$ & $69.2$ & $75.7$ & $62.3$ & $68.1$ & $79.6$ & $57.4$ & $70.1$ & $72.7$ & $63.8$ & $81.7$ & $85.8$ & $90.0$ & $89.2$ & $94.0$ & $100.0$ & $94.0$ & $75.6$ & $85.1$ & $86.8$ & $85.9$ & $69.0$ & $74.2$ & $80.1$ & $83.5$ & $81.6$ & $89.6$ & $90.9$ & $59.7$ & $73.3$ & $78.1$ & $73.9$ \\

        & \sal & Zero-shot & \DONE{$24.8$} & \DONE{$30.7$} & \DONE{$32.3$} & \DONE{$66.8$} & \DONE{$17.4$} & \DONE{$19.4$} & \DONE{$67.1$} & \DONE{$30.2$} & \DONE{$41.6$} & \DONE{$66.7$} & \DONE{$28.7$} & \DONE{$48.5$} & \DONE{$52.5$} & \DONE{$59.4$} & \DONE{$78.8$} & \DONE{$80.4$} & \DONE{$99.6$} & \DONE{$80.7$} & \DONE{$32.6$} & \DONE{$39.3$} & \DONE{$77.8$} & \DONE{$52.8$} & $X$ & $X$ & $X$ & $X$ & $X$ & $X$ & $X$ & $X$ & $X$ & $X$ \\

        \midrule
        \multirow{8}{*}{\rotatebox[origin=c]{90}{nuScenes}}
        & PHNet~\cite{li2022panoptic} & Full & $74.7$ & - & $84.2$ & $88.2$ & $74.0$ & - & - & $75.9$ & - & - & $79.7$ & -- & -- & -- & -- & -- & -- & -- & -- & -- & -- & -- & -- & -- & -- & -- & -- & -- & -- & -- & -- & -- & --\\ 
        & DS-Net~\cite{hong2021lidar} & Full & $51.2$ & - & $59.0$ & $86.1$ & $38.4$ & $43.8$ & $86.7$ & $72.3$ & $84.2$ & $85.0$ & $73.5$  & -- & -- & -- & -- & -- & -- & -- & -- & -- & -- & -- & -- & -- & -- & -- & -- & -- & -- & -- & -- & -- & --\\ 
        & GP-S3Net~\cite{razani2021gp} & Full & $61.0$ & $67.5$ & $72.0$ & $84.1$ & $56.0$ & $65.2$ & $85.3$ & $66.0$ & $78.7$ & $82.9$ & $75.8$  & -- & -- & -- & -- & -- & -- & -- & -- & -- & -- & -- & -- & -- & -- & -- & -- & -- & -- & -- & -- & -- & --\\ 
        & EfficientLPS~\cite{sirohi2021efficientlps} & Full & $62.0$ & $65.6$ & $73.9$ & $83.4$ & $56.8$ & $68.0$ & $83.2$ & $70.6$ & $83.6$ & $83.8$ & $65.6$  & -- & -- & -- & -- & -- & -- & -- & -- & -- & -- & -- & -- & -- & -- & -- & -- & -- & -- & -- & -- & -- & --\\ 
        & PolarSeg~\cite{zhou2021panoptic} & Full & $63.4$ & $67.2$ & $75.3$ & $83.9$ & $59.2$ & $70.3$ & $84.1$ & $70.4$ & $83.5$ & $83.6$ & $66.9$  & -- & -- & -- & -- & -- & -- & -- & -- & -- & -- & -- & -- & -- & -- & -- & -- & -- & -- & -- & -- & -- & --\\ 
        & MaskPLS~\cite{marcuzzi2023ral} & Full & $57.7$ & $60.2$ & $66.0$ & $71.8$ & $64.4$ & $73.3$ & $84.8$ & $52.2$ & $60.7$ & $62.4$ & $62.5$ & $71.5$ & $79.2$ & $81.0$ & $86.2$ & $84.4$ & $96.6$ & $88.0$ & $65.1$ & $76.7$ & $85.3$ & $80.6$ & $68.2$ & $77.6$ & $80.7$ & $83.2$ & $75.9$ & $88.2$ & $85.7$ & $60.5$ & $73.2$ & $80.8$ & $76.5$ \\
        \cmidrule{2-36}
        & \sal  & Full  & $70.5$ & $78.9$ & $80.8$ & $85.9$ & $79.4$ & $88.3$ & $89.6$ & $61.7$ & $73.3$ & $82.3$ & $72.8$ & $74.2$ & $82.9$ & $82.7$ & $87.1$ & $86.6$ & $96.7$ & $89.4$ & $68.0$ & $75.7$ & $86.0$ & $84.0$ & $73.5$ & $82.1$ & $84.6$ & $85.7$ & $82.9$ & $93.0$ & $89.0$ & $64.2$ & $76.3$ & $82.5$ & $79.1$ \\

        & \sal & Zero-shot & \DONE{$38.4$} & \DONE{$44.0$} & \DONE{$47.8$} & \DONE{$77.2$} & \DONE{$47.5$} & \DONE{$56.0$} & \DONE{$84.0$} & \DONE{$29.2$} & \DONE{$39.6$} & \DONE{$70.4$} & \DONE{$33.9$} & \DONE{$52.6$} & \DONE{$50.9$} & \DONE{$63.5$} & \DONE{$77.3$} & \DONE{$73.3$} & \DONE{$89.3$} & \DONE{$81.6$} & \DONE{$42.2$} & \DONE{$50.5$} & \DONE{$75.1$} & \DONE{$52.6$} & $X$ & $X$ & $X$ & $X$ & $X$ & $X$ & $X$ & $X$ & $X$ & $X$ \\

    \bottomrule
    \end{tabular}
    \label{tab:semkitti_panoseg}
    
\end{table*}

\subsection{Lidar Panoptic Segmentation Evaluation}
\label{sec:benchmarks}

Finally, we compare \sal with state-of-the-art LPS methods on standard benchmarks~\cite{behley2019iccv, fong21ral}. 
We focus this discussion on comparing \sal, trained using labeled data, to our \sal zero-shot model. For the zero-shot evaluation, we specify classes, defined in respective dataset vocabularies~\cite{behley2019iccv, fong21ral}, as text prompts (we detail these prompts in \arxiv{\cref{sec:zero_shot_classification_app}}\conf{appendix}). Importantly, zero-shot results are obtained without training on \emph{any} labeled data. \textit{In contrast, all baselines are trained using human-labeled data and are constrained to their respective pre-defined class vocabularies.}

\PAR{SemanticKITTI.}
When evaluating with the \textit{default class} vocabulary (\cref{tab:semkitti_panoseg}), our zero-shot model reaches $24.8$ PQ or $42\%$ of the fully-supervised model ($59.5$ PQ). %
As discussed in~\cref{sec:ablations}, our method is mainly limited by the quality of the generated CLIP features.
In practice, our zero-shot semantic classifier mainly confuses objects within super classes, \eg, \texttt{car} \vs \texttt{other-vehicle}. 
Therefore, \textbf{without any retraining}, we prompt our zero-shot model with \textit{super classes} and obtain $48.5$ PQ, $59\%$ of the supervised model ($81.7$ PQ). %

\PAR{nuScenes.} We report our result in~\cref{tab:semkitti_panoseg} and reach similar conclusions. 
\sal reaches $38.4$ PQ on \textit{default classes}, $54 \%$ of the supervised model ($70.5$ PQ). 
As with SemanticKITTI, results consistently improve when evaluated with \textit{super-classes}. 
In this setting, zero-shot model yields $52.6$ PQ. 
Note that \sal model, supervised with labeled data, significantly outperforms MaskPLS.

\PAR{Towards closing the gap.} 
Compared to SemanticKITTI ($24.8$ PQ, $42 \%$), we observe a significantly smaller gap between the model trained on pseudo- and ground truth labels for nuScenes ($38.4$ PQ, $54\%$).
Despite its sparse Lidar signal and more diverse input scenarios, \sal performs better on this challenging Lidar dataset. 
We attribute this to a larger pseudo-label coverage of nuScenes ($48\%$, compared to $14\%$ coverage on SemanticKITTI), and, therefore, effectively more labeled data, which is also consistent with our observations on SemanticKITTI \wrt increasing the amount of labeled data (\cref{tab:ablation_num_queries}). 
This trend suggests a clear path forward: \emph{applying \sal to pseudo-label more data with zero annotation cost.}

\PAR{Beyond labeled datasets.} Datasets, such as SemanticKITTI and nuScenes, provide dense, point-level panoptic segmentation labels, which we use for analysis and the evaluation of our method on a small set of classes that were labeled in the respective dataset ($14/11$ \thing/\stuff in SemanticKITTI, respectively, and $23/6$ in nuScenes). To demonstrate the versatility of \sal, we additionally pseudo-label and train a model on Waymo Open~\cite{sun20CVPR} dataset and show qualitative results in \cref{fig:teaser} and \arxiv{\cref{sec:qualitative-appendix}}\conf{more in the appendix}. This effort takes less than a week of pseudo-labeling and model training efforts (we provide details in \arxiv{\cref{sec:pseudo_label_timing}}\conf{the appendix}). 

\PAR{The good, the (breaking) bad, and the ugly.} 
While we are thrilled about class-agnostic segmentation performance, our experimental analysis reveals a substantial gap between full supervision and our zero-shot model. Finally, at the moment, \sal is limited to training on specific datasets. As in fully-supervised Lidar perception, cross-sensor generalization remains a challenge for \sal. This could be addressed by utilizing temporal context, data scaling efforts and investigating sensor-agnostic backbone networks. 

\section{Conclusions}

We proposed \sal, a method for \textit{Zero-Shot} \lps. The key \sal components are a pseudo-label engine that utilizes vision foundation models and a zero-shot model trained via self-supervision. 
While \sal is the first step in the direction of learning general Lidar segmentation models by distilling image-based foundation to Lidar, we believe we just scratched the surface and opened the door for training Lidar segmentation models without manual supervision.

\footnotesize{\PAR{Acknowledgments.} This project was funded, in parts, by ERC
Starting Grant DynAI (ERC-101043189).} 
We are grateful to Žan Gojčič, Guillem Braso, Cristiano Saltori, Sérgio Agostinho, and Jonas Schult for their feedback on the paper and their insightful comments. Special thanks to Maxim Maximov for his help on figures.

\clearpage  %

\bibliographystyle{splncs04}
\bibliography{refs}

\arxiv{\clearpage}

\appendix
\counterwithin{figure}{section}
\counterwithin{table}{section}

\begin{center}
    {\Large \bfseries Appendix}
\end{center}

\begin{abstract}
    The appendix provides additional implementation and design details of the \sal pseudo-label engine (\ref{sec:impl_design_details_label_engine_app}) and zero-shot model (\ref{sec:impl_design_details_zero_shot_model_app}), along with a discussion on different training strategies with partial pseudo-labels (\ref{sec:training-sal-partial-labels}). 
    In~\ref{sec:txt_prompt_eng_app}, we discuss the zero-shot text prompt engineering.
    Further ablations on our pseudo-labels and model are provided in~\ref{sec:pseudo_label_ablations_app} and~\ref{sec:model_ablations_app}, respectively, including fine-grained, per-class results (\ref{sec:qual_pre_class_app}) and detailed statistics on the generated pseudo-labels (\ref{sec:pseudo_label_stats}) and time and compute analysis for pseudo-labeling (\ref{sec:pseudo_label_timing}). 
    We conclude with a discussion on detailed qualitative results in~\ref{sec:qualitative-appendix}.
\end{abstract}

\section{Implementation Details}
\label{sec:impl_design_details_app}

\subsection{Pseudo-label Engine}
\label{sec:impl_design_details_label_engine_app}

This section details the overview of our pseudo-labeling procedure \arxiv{(\cref{sec:pseudo_label_generation}} of the main paper). We detail our pseudo-label generation steps in \cref{alg:pseudo_label_generation}. As input, we assume a sequence of multi-modal data that consists of $k$ images $\mathcal{I}_{t, k} \in \mathbb{R}^{W\times H \times 3}$ (one per camera view, per frame $t, t \in 1, \ldots, T$), Lidar point clouds $P_t \in \mathbb{R}^{N \times 4}$, and image and camera-to-Lidar calibration. 
The output is pairs of Lidar segmentation masks $\tilde{m}_t \in \{0, 1\}^{N}$ and corresponding CLIP feature tokens $f_t \in \mathbb{R}^{768}$, $t \in 1, \ldots, T$. 

\PAR{Mask generation.} We start by generating an overlapping set of segmentation masks for each image (camera view) (\cref{alg:pseudo_label_generation}, L6) using \textit{Segment Anything} (SAM) foundation model~\cite{kirillov2023segment}. To generate masks, we utilize parameters, as described in \cref{tab:hyperparameters}. This step yields $171$ masks per image on average. We then flatten SAM's output mask hierarchy by non-maxima suppression (NMS) with a small overlap threshold (\cref{alg:pseudo_label_generation}, L7, govern by \textit{NMS IoU threshold} in \cref{tab:hyperparameters}) to obtain a small, non-overlapping set of masks per-scan ($45$ on average after flatening). This way, we obtain a set of non-overlapping binary masks $m_t^k \in \{ 0,1 \}^{W \times H}$ for each camera view $k$ with an image plane of size $W \times H$. 

\begin{figure*}[t]
    \centering
    \begin{tikzpicture}
        \draw (0, 0) node[inner sep=0] {\includegraphics[width=0.93\linewidth]{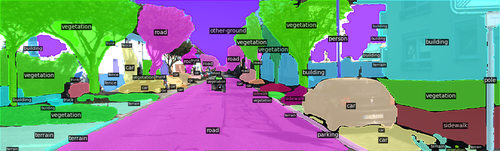}};
        \draw (0, 0) node {\huge \color{white} front};
    \end{tikzpicture}
    \begin{tikzpicture}
        \draw (0, 0) node[inner sep=0] {\includegraphics[width=0.93\linewidth]{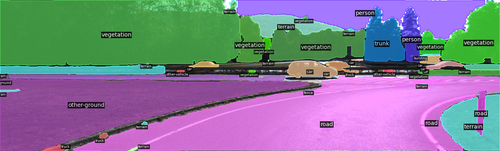}};
        \draw (0, 0) node {\huge \color{white} front};
    \end{tikzpicture}
    \begin{tikzpicture}
        \draw (0, 0) node[inner sep=0] {\includegraphics[width=0.93\linewidth]{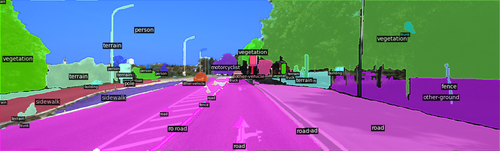}};
        \draw (0, 0) node {\huge \color{white} front};
    \end{tikzpicture}
    \begin{tikzpicture}
        \draw (0, 0) node[inner sep=0] {\includegraphics[width=0.93\linewidth]{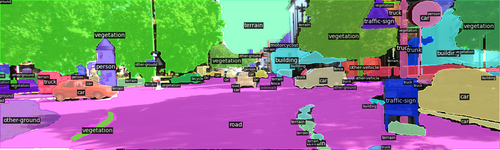}};
        \draw (0, 0) node {\huge \color{white} front};
    \end{tikzpicture}

    \caption{
    \textbf{SAM mask and CLIP vocabulary predictions - SemanticKITTI.}
    We show the front camera view of the first scan of the 00, 01, 02, and 08 sequences.
    We obtain masks with SAM~\cite{kirillov2023segment} and compute per-mask CLIP~\cite{radford2021learning} image tokens with MaskCLIP~\cite{ding2023open}.
    To visualize classes, we prompt generated tokens with SemanticKITTI~\cite{behley2019iccv} class vocabulary. 
    These classes are not transferred to Lidar.
    Our pseudo-labels contain masks and image tokens, no explicit class labels.
    Instances of the same class are indicated in different tones of the same color.
    }
    \label{fig:sam_and_clip_image_results-kitti}
\end{figure*}

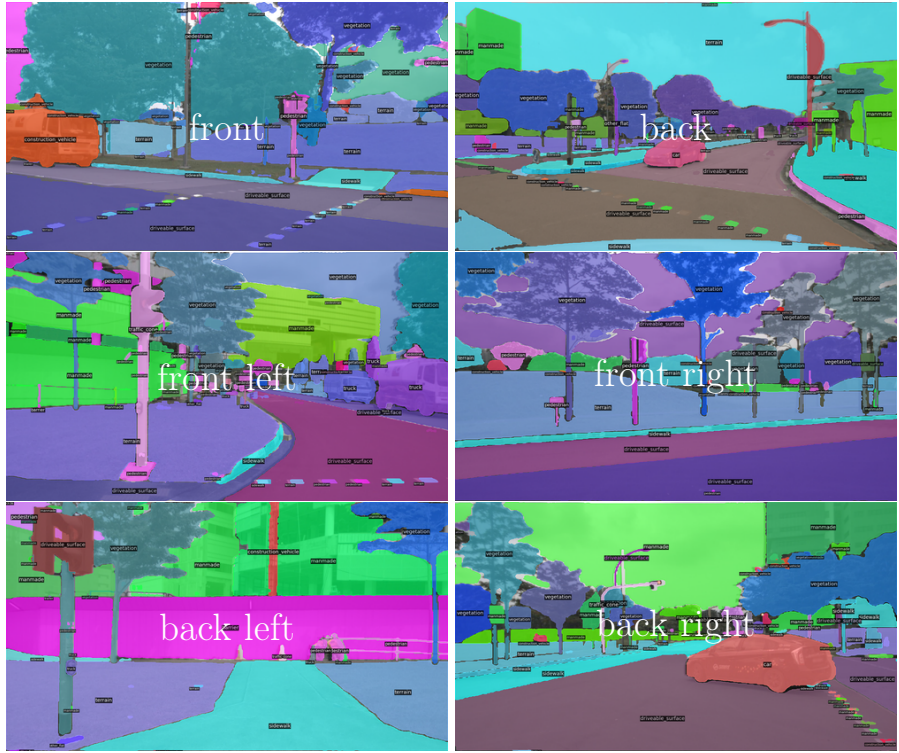
\begin{figure*}[t]
    \centering
    \begin{tikzpicture}
        \draw (0, 0) node[inner sep=0] {\includegraphics[width=0.48\linewidth]{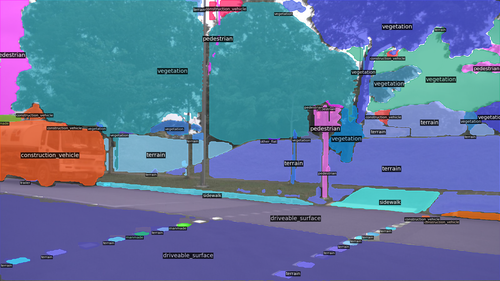}};
        \draw (0, 0) node {\Large \color{white} front};
    \end{tikzpicture}
    \begin{tikzpicture}
        \draw (0, 0) node[inner sep=0] {\includegraphics[width=0.48\linewidth]{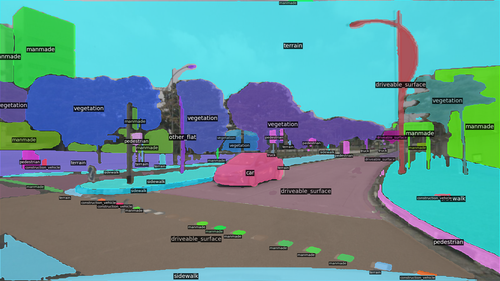}};
        \draw (0, 0) node {\Large \color{white} back};
    \end{tikzpicture}
    
    \begin{tikzpicture}
        \draw (0, 0) node[inner sep=0] {\includegraphics[width=0.48\linewidth]{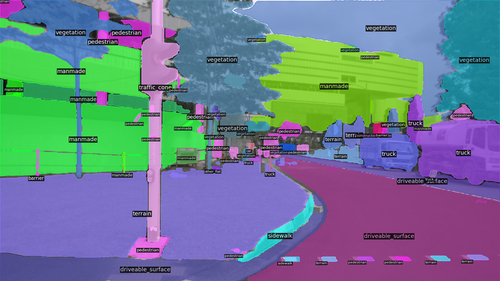}};
        \draw (0, 0) node {\Large \color{white} front left};
    \end{tikzpicture}
    \begin{tikzpicture}
        \draw (0, 0) node[inner sep=0] {\includegraphics[width=0.48\linewidth]{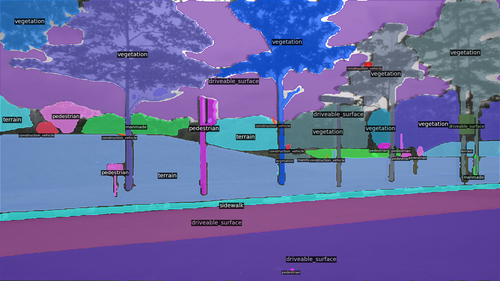}};
        \draw (0, 0) node {\Large \color{white} front right};
    \end{tikzpicture}
    
    \begin{tikzpicture}
        \draw (0, 0) node[inner sep=0] {\includegraphics[width=0.48\linewidth]{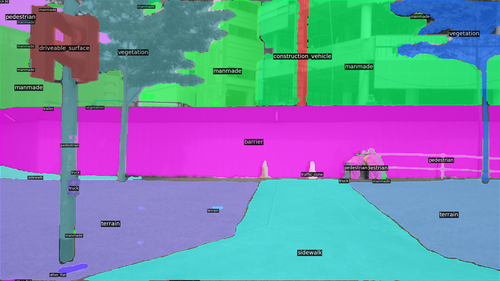}};
        \draw (0, 0) node {\Large \color{white} back left};
    \end{tikzpicture}
    \begin{tikzpicture}
        \draw (0, 0) node[inner sep=0] {\includegraphics[width=0.48\linewidth]{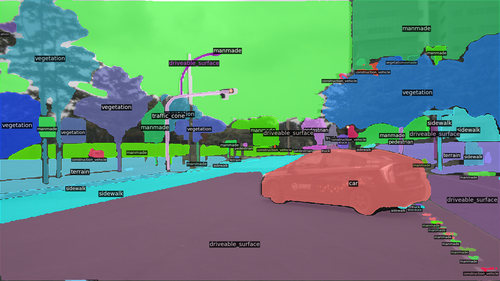}};
        \draw (0, 0) node {\Large \color{white} back right};
    \end{tikzpicture}
    \caption{
    \textbf{SAM mask and CLIP vocabulary predictions - nuScenes.}
    We show all six camera views of the first scan of the \emph{0001} sequence.
    We obtain masks with SAM~\cite{kirillov2023segment} and compute per-mask CLIP image tokens with MaskCLIP~\cite{ding2023open}.
    To visualize classes, we prompt generated tokens with nuScenes~\cite{fong21ral} class vocabulary.
    These classes are not transferred to Lidar.
    Our pseudo-labels contain masks and image tokens, no explicit class labels.
    Instances of the same class are indicated in different tones of the same color.
    }
    \label{fig:sam_and_clip_image_results-nuscenes}
\end{figure*}

In practice, during mask suppression, we sort masks based on the mask area (rather than score, usually done in NMS) -- this criterion favors objects over their parts and subparts. As can be seen in \cref{tab:label_validation__detailed}, this approach (\textit{NMS area}) performs significantly better as compared to objectness score-based suppression (\textit{NMS score}), which may favor object parts over objects.
We visualize the flattened segmentation masks in images in \cref{fig:sam_and_clip_image_results-kitti} (SemanticKITTI) and \cref{fig:sam_and_clip_image_results-nuscenes} (nuScenes).

\begin{figure*}[t]
    \centering
    \begin{tikzpicture}
        \draw (0, 0) node[inner sep=0] {\includegraphics[width=0.48\linewidth]{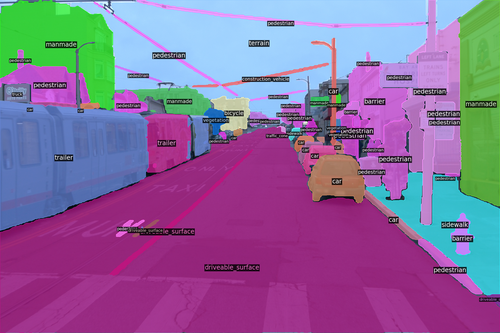}};
        \draw (0, 0) node {\Large \color{white} front};
    \end{tikzpicture}
    
    \begin{tikzpicture}
        \draw (0, 0) node[inner sep=0] {\includegraphics[width=0.48\linewidth]{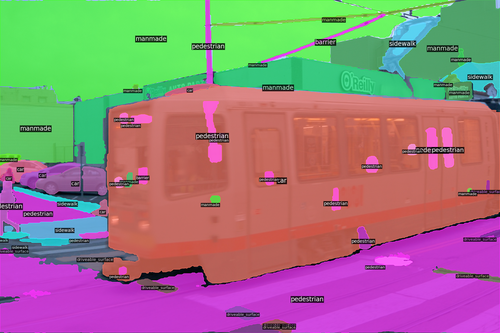}};
        \draw (0, 0) node {\Large \color{white} front left};
    \end{tikzpicture}
    \begin{tikzpicture}
        \draw (0, 0) node[inner sep=0] {\includegraphics[width=0.48\linewidth]{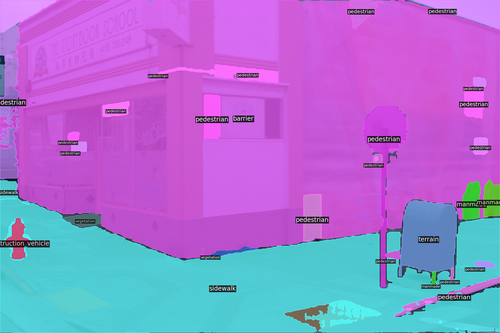}};
        \draw (0, 0) node {\Large \color{white} front right};
    \end{tikzpicture}

    \begin{tikzpicture}
        \draw (0, 0) node[inner sep=0] {\includegraphics[width=0.68\linewidth]{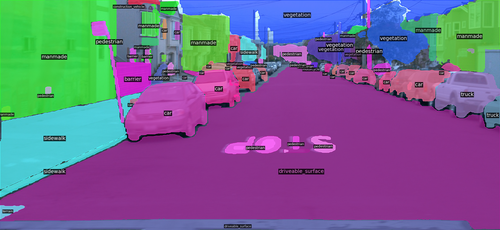}};
        \draw (0, 0) node {\Large \color{white} left};
    \end{tikzpicture}
    
    \begin{tikzpicture}
        \draw (0, 0) node[inner sep=0] {\includegraphics[width=0.68\linewidth]{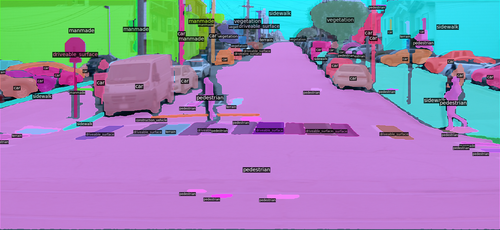}};
        \draw (0, 0) node {\Large \color{white} right};
    \end{tikzpicture}
    \caption{
    \textbf{SAM mask and CLIP vocabulary predictions - Waymo.}
    We show all five camera views of the first frame of sequence emph{008} of the test set.
    We obtain masks with SAM~\cite{kirillov2023segment} and compute per-mask CLIP image tokens with MaskCLIP~\cite{ding2023open}.
    Sine WAYMO has no panoptic ground truth, we visualize classes by prompting generated tokens with the nuScenes~\cite{fong21ral} class vocabulary.
    These classes are not transferred to Lidar.
    Our pseudo-labels contain masks and image tokens, no explicit class labels.
    Instances of the same class are indicated in different tones of the same color.
    }
    \label{fig:sam_and_clip_image_results-waymo}
\end{figure*}
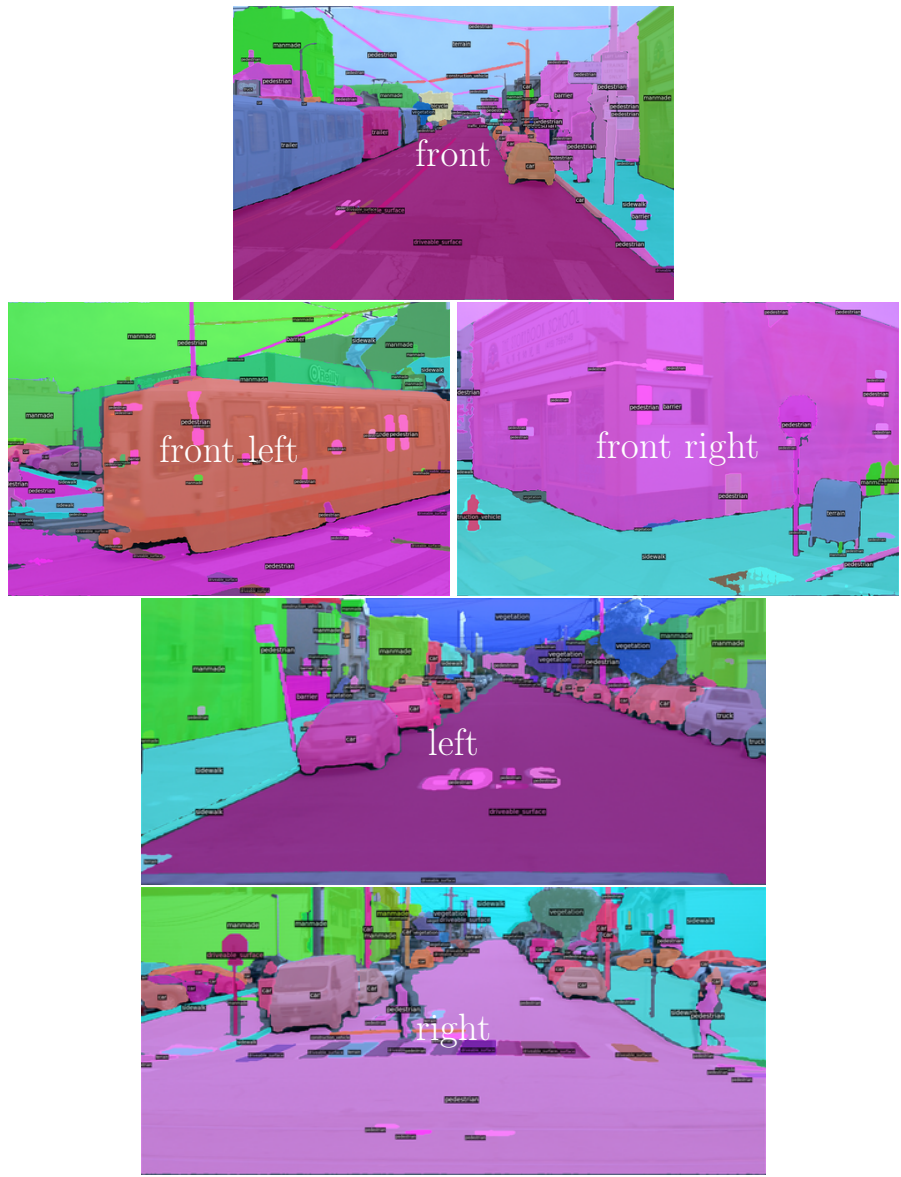

\PAR{CLIP image token generation.} We proceed by generating localized CLIP~\cite{radford2021learning} image feature tokens $f_t^k \in \mathbb{R}^{768}$ for each binary mask $m_t^k$ (\cref{alg:pseudo_label_generation}, L8).
To obtain image tokens for a masked region of the input image, we utilize MaskCLIP~\cite{ding2023open} and their relative mask attention in the CLIP image encoder feature space. 
\DONE{The original MaskCLIP pipeline forwards the entire image and all masks at once. For our use case, we observed better per-mask classification results by generating image tokens for each mask separately. To this end, we forward individual image crops.}
In \cref{fig:sam_and_clip_image_results-kitti} and \cref{fig:sam_and_clip_image_results-nuscenes}, we visualize for each generated mask in the image the most-likely class, assigned using the generated CLIP token (according to the class vocabularies, as detailed in \cref{tab:text_prompts}). Importantly, we report class names only for visualization purposes in~\cref{fig:sam_and_clip_image_results-kitti}, \cref{fig:sam_and_clip_image_results-nuscenes}, and \cref{fig:sam_and_clip_image_results-waymo}. We \textbf{never} classify segments in the image domain but merely distill CLIP image feature tokens to Lidar to facilitate zero-shot classification during inference where we do not use any image features.

\PAR{Image-to-Lidar unprojection.} From the Lidar perspective, we unproject each image mask $m_{t,i}^k \in \{ 0,1 \}^{W \times H}$ to a binary Lidar segmentation mask $\tilde{m}^k_{t,i} \in \{0, 1 \}^N$ by transforming the respective camera coordinate frame to the Lidar space (\cref{alg:pseudo_label_generation}, L9--L10). 
The unprojection yields pairs $\{\tilde{m}_t^k, f_t^k \}$ of Lidar masks and their corresponding CLIP features \wrt camera $k$. On average, from $45$ masks generated in images, $39$ are transferred to the Lidar domain -- the rest are either too small or not supported by Lidar measurements (due to signal sparsity or lack of Lidar coverage). We visualize masks, unprojected from a single camera to Lidar in the middle row of \cref{fig:qual_gt_pseudo_outputs-kitti} (SemanticKITTI). 
    
We then insert these masks to the output sets $\tilde{m}_t, f_t$ as follows. To all masks in $\tilde{m}_t^k$ that do not overlap significantly overlap with masks in $\tilde{m}_t$ (\cref{tab:hyperparameters}, \textit{multi-view IoU threshold}), we assign a new ID, and insert masks and their corresponding CLIP features to the output sets $\tilde{m}_t, f_t$. For masks whose overlap threshold exceeds this limit, we update existing masks with a union of the two, and the feature with the average. We visualize masks, unprojected to Lidar from several views, followed by fusion, in the middle row of \cref{fig:qual_gt_pseudo_outputs-nuscenes} (nuScenes).

\PAR{Refinement via clustering.} Finally, we improve our pseudo-label quality by creating an ensemble of DBSCAN~\cite{Ester96KDD} clusters $\tilde{m}_t^{DBSCAN} \in \{ 0,1 \}^{N}$, obtained by varying the density thresholds (\cref{tab:hyperparameters}, \textit{DBSCAN density thresholds}) to compensate for varying density in Lidar point clouds. %
We replace each $\tilde{m}_i \in \tilde{m}_t$ with its best-matching $\tilde{m}_l \in \tilde{m}_t^{DBSCAN}$ in case their IoU exceeds a minimal overlap threshold (\cref{tab:hyperparameters}, \textit{DBSCAN IoU overlap threshold}) and retain the original mask otherwise to obtain a refined set of pseudo-labels that retain their original cardinality and associated CLIP features. 

\PAR{DBSCAN clustering implementation.} To create the ensemble of DBSCAN clusters $\tilde{m}_t^{DBSCAN}$, we first perform geometric plane fitting and remove ground points (to estimate the ground points, we use~\cite{lee2022patchwork++} and its publicly-available implementation). We then perform DBSCAN on the ground-filtered point clouds using six density thresholds (reported in \cref{tab:hyperparameters}, \textit{DBSCAN IoU overlap threshold}). This leads to a large set of (overlapping) segmentation masks, induced by the estimated point clusters (in SemanticKITTI, $5,413$ on average per scan)

In \cref{tab:label_validation__detailed}, we report two alternatives. Firstly, rather than using DBSCAN, we perform erosion in the image domain to minimize the ``edge bleeding'' artifacts. As can be seen, this variant decreases the PQ score ($46.0$ to $42.2$). Second, rather than replacing segments with DBSCAN (\textit{DBSCAN replace}), we filter the pool of SAM-generated segments, only retaining those that have a sufficient overlap (in terms of intersection-over-union), (\textit{DBSCAN filter}). While this variant improves PQ for \thing classes, it significantly reduces PQ for \stuff classes, leading to overall lower performance compared to the replace variant ($46.7$ PQ for \textit{filter} \vs $48.7$ PQ for \textit{replace}).

\PAR{Parameter tuning.} We tune all hyperparameters on a subset of $40$ Lidar scans of SemanticKITTI validation set (we sample every $100^{th}$ scan). We perform no parameter tuning for nuScenes. 

\begin{algorithm*}[t]
\caption{Pseudo-label Engine}
\begin{algorithmic}[1]
\Require Lidar point clouds $P_t$, $k$ camera views $\mathcal{I}_{t, k}$, $k$ camera calibrations $c_k$, $t \in 1, \ldots, T$
\Ensure $\{\tilde{m}_t, f_t \}, t \in 1, \ldots, T$ // Lidar segmentation masks and corresponding CLIP image feature tokens.
\For{each timestamp $t$}
    \State $P_t$ $\leftarrow$ load\_lidar($t$)
    \State $\tilde{m}_t = \emptyset$, $f_t = \emptyset$
    \State $\tilde{m}_t^{DBSCAN}$ $\leftarrow$ DBSCAN\_ensamble($P_t$) // Generate (overlapping) Lidar mask ensemble
    \For{each camera $k$}
        \State $\mathcal{I}_{t, k}$ $\leftarrow$ load\_image($t, k$)
        \State $m_t^k$ $\leftarrow$ SAM($\mathcal{I}_{t, k}$) // Generate masks in image $k$
        \State $m_t^k$ $\leftarrow$ NMS($m_t^k$) // Apply Non-Maximum Suppression (NMS) to masks
        \State $f_t^k \leftarrow$ MaskCLIP($\mathcal{I}_{t, k}$, $m_t^k$) // Obtain localized CLIP features for each mask
        \State $\tilde{m}_t^k \leftarrow$ label\_point\_cloud ($P_{t, k}, m_t^k$, $c_k$) // Generate segmentation pseudo-labels \wrt camera $k$
        \State $\tilde{m}_t^k \leftarrow$ replace($\tilde{m}_t^k, \tilde{m}_t^{DBSCAN}$) // Replace with sufficiently overlapping DBSCAN masks
        \State \{$\tilde{m}_t, f_t \} \leftarrow$ insert\_or\_merge($\tilde{m}_t, f_t, \tilde{m}_t^k$, $f_t^k$) // Cross-camera fusion: Insert new or merge with existing 
    \EndFor
\EndFor
\end{algorithmic}
\label{alg:pseudo_label_generation}
\end{algorithm*}

\subsection{Zero-shot Model}
\label{sec:impl_design_details_zero_shot_model_app}

\begin{table}[t]
    \centering
    \scriptsize
    \caption{
        \textbf{\sal hyperparameters.}
        We show parameters for both components of our framework: (i) SAM model~\cite{kirillov2023segment}, which we use to generate segmentation masks in images; (ii) the pseudo-label generation engine and (iii) our zero-shot model.
        For the latter, we only highlight parameters that deviate from~\cite{marcuzzi2023ral}.
    }
    \begin{tabular}{l|l}
    \toprule
    Parameter & Value \\
    \midrule
    \multicolumn{2}{c}{SAM~\cite{kirillov2023segment}} \\
    \midrule
    Model & sam\_vit\_h\_4b8939 \\
    Inference POINTS\_PER\_SIDE & $32$ \\
    Inference PRED\_IOU\_THRESH & $0.84$ \\
    Inference STABILITY\_SCORE\_THRESH & $0.86$ \\
    Inference MIN\_MASK\_REGION\_AREA & $100$ \\    
    \midrule
    \multicolumn{2}{c}{Pseudo-label engine} \\
    \midrule

    NMS IoU threshold & $0.01$ \\
    Multi-view IoU threshold & $0.01$ \\
    \DONE{DBSCAN IoU overlap threshold} & \DONE{$0.5$}\\
    DBSCAN density thresholds & ($1.2488, 0.8136, 0.6952,$ \\
                               & $0.594, 0.4353, 0.3221$) \\

    \midrule
    \multicolumn{2}{c}{Zero-shot model} \\
    \midrule

    GPUs & 8 $\times$ 32GB (V100) \\ 
    Batch size & 24 (3 per GPU) \\
    Learning rate (LR) & 0.0003 \\
    Number of epochs & 30 \\
    LR drop & 15 \\
    Number of queries & 300 \\
    Overlap threshold & 0.0 \\
    Loss weights & 2.0, 5.0, 5.0, 5.0, 2.0 \\

    \bottomrule
    \end{tabular}
    \label{tab:hyperparameters}
\end{table}

\begin{figure}[t]
    \centering
    \includegraphics[trim={0 80px 330px 0},clip,width=\linewidth]{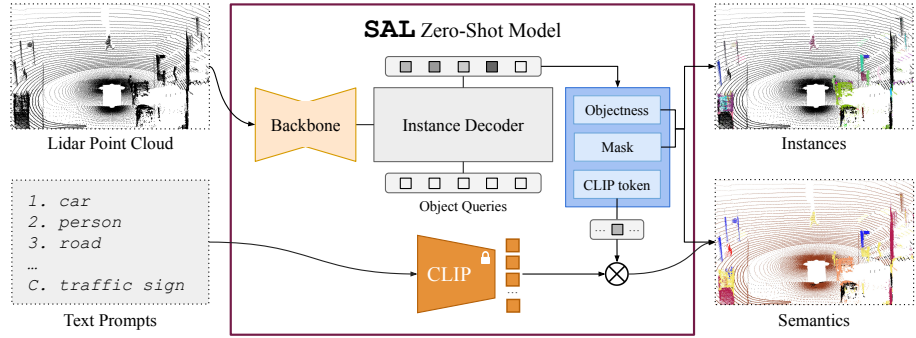}
    \caption{\textbf{\sal zero-shot model.}
    Our model takes Lidar point clouds and text prompts as inputs.
    Its architecture relies on a 3D sparse-convolutional Minkowski backbone~\cite{choy20194d} followed by a Transformer decoder for object instance segmentation.
    The decoder computes cross-attention between object queries and backbone features.
    Three task heads predict objectness scores, segmentation masks, and CLIP tokens for each query.  
    Once trained, we forward the dataset class vocabulary through the CLIP text encoder and perform zero-shot classification via matching with predicted CLIP tokens.
    The model requires no retraining for different vocabularies.
    }
    \label{fig:model}
\end{figure}

We use our \sal model, trained using labels generated by the pseudo-label engine (\cref{sec:impl_design_details_label_engine_app}), to segment and classify \emph{any} object. 
The model implements a Transformer decoder architecture and adds a CLIP token prediction head (see~\cref{fig:model}).
In contrast to MaskPLS~\cite{marcuzzi2023ral}, our decoder operates not in point but voxel space and only on a single backbone feature scale.
These adaptations obtained empirically better results from an overall more lightweight model.
As it is common for Transformer decoders, the computation of its loss $\mathcal{L}_{\text{\sal}}$ relies on a bipartite cost matching between predictions and (pseudo-)labels.
Empirically, we obtain better results if this matching only relies on the class-agnostic segmentation and does not incorporate the image token prediction as an additional cost.
Unlike~\cite{marcuzzi2023ral}, we train all models from scratch and do not initialize our backbone with SegContrast~\cite{nunes2022segcontrast}.
In~\cref{tab:hyperparameters}, we highlight hyperparameter choices that deviate from the default setup in~\cite{marcuzzi2023ral}.

The \emph{overlap threshold} filters segments during inference.
Since the segmentation head predicts binary masks per query, the resulting segments are potentially overlapping.
To flatten the output and obtain panoptic segmentation results, overlapping points are assigned based on probability.
If a percentage of points are assigned to another segment, the overlap threshold removes the segment entirely from the output.
Training on partial and noisy pseudo-labels increases the amount of overlap during inference, in particular, for areas that are out of the camera frustum(s) and hence not labeled in the training data.
Deactivating the overlap filter entirely yields the best results for our \sal model.

The \emph{loss weights} balance each component of our full model loss:
\begin{align}
    \label{model_loss_full}
    \begin{split}
        \mathcal{L}_{\text{\sal}} &= \mathcal{L}_{obj} + \mathcal{L}_{mask} + \mathcal{L}_{dice} \\                            &+ \mathcal{L}_{token} + \mathcal{L}_{token\_aux}.
    \end{split}
\end{align}
The $\mathcal{L}_{token\_aux}$ is an auxiliary segmentation loss as applied in~\cite{marcuzzi2023ral}.
This loss computes the semantic segmentation quality not for the decoder queries but based on the backbone features alone.
To this end, a per-point semantic segmentation head is added after the backbone.
In contrast to~\cite{marcuzzi2023ral}, our head regresses a token per-point and not class per point.
This auxiliary head is only added for training and discarded during inference.
All of our trainings apply common spatial augmentations, including random rotations, flipping, scaling, and translations.

\subsection{Text Prompt Engineering}
\label{sec:txt_prompt_eng_app}
To perform zero-shot classification within a pre-defined class vocabulary, we complement and enrich the otherwise ambiguous and uninformative class names.
As shown in~\cref{tab:text_prompts}, each class is predictable not only by its own name but a set of additional prompts.
In particular, all \texttt{other-X} classes are ambiguous prompts.
The otherness only works in a fully supervised setting where a model can learn to predict, for example, all vehicles except the types already covered by other classes.
In our case, we must directly prompt for other vehicle types, such as \texttt{trailer}, \texttt{bus}, \texttt{tram}, or \texttt{train}.
This problem could also be solved by adding negations/exclusions to prompts.
However, this approach did not yield the desired outcome in our experiments.
We apply the same set of super classes for SemanticKITTI~\cite{behley2019iccv} and nuScenes~\cite{fong21ral} and merge their respective default class prompts.
Furthermore, we follow~\cite{radford2021learning} and wrap every prompt into a list of full-sentence templates.
This results in text prompts like \texttt{a photo of a car}, which better align with the image caption training data of the CLIP text encoder.
In~\cref{sec:zero_shot_classification_app}, we show additional ablations on the aforementioned text prompt engineering.

For panoptic segmentation outputs, every point is classified to one dataset vocabulary class.
To avoid classifying all segments to the same class when given a single text prompt, we append a second background prompt to the prompt set.
More specifically, all predicted CLIP tokens are prompted with the target text and the word \texttt{other}.
Empirically, we observed that this broad term reliably matches to all objects unless the actual text prompt, \eg, \texttt{fire hydrant}, is being segmented.

\begin{table}[t]
    \centering
    \scriptsize
    \caption{
        \textbf{CLIP token distillation and text prompt engineering.}
        To evaluate our token prediction, we prompt the SemanticKITTI class vocabulary to generate labeled training data and train a non-zero-shot model (row 1).
        Furthermore, we demonstrate the insufficiency of vanilla class names (\texttt{car}) as text prompts and the boost from engineering a rich set of terms (\texttt{car}, \texttt{jeep}, \texttt{SUV}, \texttt{van}) as explained in~\cref{sec:txt_prompt_eng_app}.
    }
    \begin{tabular}{cc|cccc}
    \toprule
    \multirow{2}{*}{\shortstack[c]{Text prompt  \\ engineering}} & \multirow{2}{*}{$\mathcal{L}_{token}$} & \multirow{2}{*}{PQ} & \multirow{2}{*}{RQ} & \multirow{2}{*}{SQ} & \multirow{2}{*}{mIoU} \\
     & & & & & \\
    \midrule
    \multicolumn{6}{c}{Default classes} \\
    \midrule

    $\times$ &          & $25.1$ & $33.5$ & $68.4$ & $25.9$ \\
    
    & $\times$          & \DONE{$20.6$} & \DONE{$27.1$} & \DONE{$65.2$} & \DONE{$20.9$} \\
    
    \midrule

    $\times$ & $\times$   & \DONE{$24.8$} & \DONE{$32.3$} & \DONE{$66.8$} & \DONE{$29.7$} \\

    \midrule
    \multicolumn{6}{c}{Super classes} \\
    \midrule

    & $\times$    & \DONE{$27.4$} & \DONE{$33.8$} & \DONE{$71.8$} & \DONE{$24.9$} \\

    $\times$ & $\times$    & \DONE{$48.5$} & \DONE{$59.4$} & \DONE{$78.8$} & \DONE{$52.8$} \\

    \bottomrule
    \end{tabular}
    \label{tab:ablation_semantics}
\end{table}

\subsection{How to Train on Partial Labels?} 
\label{sec:training-sal-partial-labels}

\begin{figure}[t]
    \centering
    \begin{subfigure}[t]{0.49\linewidth}
      \includegraphics[width=\textwidth]{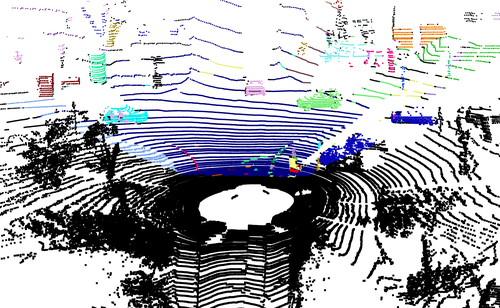}
      \caption{Partial pseudo-labels}
      \label{fig:segmentation_pseudo_label_vis_b}
    \end{subfigure}
    \begin{subfigure}[t]{0.49\linewidth}
      \includegraphics[width=\textwidth]{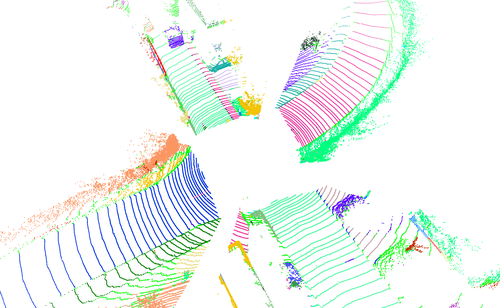}
      \caption{FrankenFrustum}
      \label{fig:segmentation_pseudo_label_vis_c}
    \end{subfigure}
    \caption{\textbf{Training on partial labels.}
    Unprojecting image-based pseudo labels results in a partially (pseudo) labeled point cloud (\cref{fig:segmentation_pseudo_label_vis_b}). We construct supervisory signal by concatenating multiple partially labeled point clouds (\cref{fig:segmentation_pseudo_label_vis_c}).
    }
    \label{fig:segmentation_pseudo_label_vis}
\end{figure}

Pseudo-labels provide only partial supervision within the camera frustum (see \cref{fig:segmentation_pseudo_label_vis_b}), leaving the majority of Lidar points unlabeled (see \cref{tab:pseudo_label_stats}). How can we train \sal with such partial supervision?
\PAR{Ignore unlabeled region + standard data augmentations.} 
During training, we remove all unlabeled points from the point cloud. Otherwise, the $\mathcal{L}_{obj}$ loss would penalize any segmentation in these regions as a false positive, thereby teaching the model to ignore them entirely. As visualized in~\cref{fig:segmentation_pseudo_label_vis_b} and quantified in~\cref{tab:pseudo_label_stats}, the 360$^{\circ}$ Lidar label coverage is particularly low for single camera setups, as in SemanticKITTI~\cite{behley2019iccv}.

\PAR{FrankenFrustum.} 
To generalize to full clouds during inference, we propose a simple but very effective \textit{FrankenFrustum} augmentation (\cref{fig:segmentation_pseudo_label_vis_c}).
It mimics full point clouds during training by randomly removing unlabeled points and replicating labeled frustum regions around the vertical axis. 
This augmentation does not increase the overall label coverage. However, our ablations in the main paper \arxiv{(\cref{sec:ablations}, \cref{tab:ablation_franken_frustum})} show its effectiveness in reducing the domain gap between training and inference input clouds. 

\section{Pseudo-label Analysis}
\label{sec:pseudo_label_ablations_app}

\subsection{Pseudo-label Statistics}
\label{sec:pseudo_label_stats}

\PAR{Point coverage.} 
As shown in~\cref{tab:pseudo_label_stats}, the single-camera setup of Semantic-KITTI~\cite{behley2019iccv} allows us to label a significantly smaller portion of Lidar point clouds ($14\%$) compared to GT labels ($98\%$). 
Interestingly, even for nuScenes dataset~\cite{fong21ral}, which provides a setup with five cameras and $360^{\circ}$ view coverage, pseudo-labels cover only $48\%$ of all points. 
Furthermore, we quantify pseudo-label coverage on SemanticKITTI when filtering all points outside of the camera frustum. This leads to coverage of $89\%$. The remaining missing labels can be explained by errors committed by our image-based segmentation foundation model, SAM~\cite{kirillov2023segment}. \textit{This analysis confirms that even when utilizing strong foundation models, image-to-Lidar distillation remains a challenging problem.}

\PAR{\texttt{Things} \vs \stuff.} Furthermore, the \emph{segment anything} philosophy transferred from SAM~\cite{kirillov2023segment} to our pseudo-labels yields a significantly larger number of instances per scan ($19$ GT \vs $40$ pseudo on average). This transfer is particularly notable in the shift of ratios between \thing and \stuff instances ($0.84$ \vs $0.14$ for GT and pseudo-labels, respectively), as existing datasets (\eg, SemanticKITTI and nuScenes) merge individual instances of classes such as \texttt{pole} or \texttt{trees} into single instances. \textit{By contrast, our pseudo-labels provide a finer-grained segmentation of object instances, as needed for learning to segment a variety of objects}.

\begin{table*}[t] %
    \centering
    \scriptsize
    \caption{\textbf{Pseudo-label statistics.}
    We outline the label coverage of point clouds, the total, max, and mean number of instances per scan, and the ratio of things/stuff instances on the full point cloud and point cloud areas that overlap with the camera view frustum (Filter Frustum). As can be seen, due to the single-camera setup, pseudo-label coverage in SemanticKITTI~\cite{behley2019iccv} is very low ($14\%$ of points). Even though nuScenes~\cite{fong21ral} dataset provides $360^{\circ}$ view coverage, only $48\%$ are labeled due to blind spots. Even when only retaining points, that overlap with the camera view frustum (SemanticKITTI, Filter Frustum), we observe coverage of $89\%$. This can be explained by mistakes (\eg, false negatives) committed by our segmentation foundation model (SAM~\cite{kirillov2023segment}).
    }
    \begin{tabular}{lc|c|cccccc}
    \toprule

    \multirow{2}{*}{Label} & \multirow{2}{*}{\shortstack[c]{Filter \\ Frustum}} & \multirow{2}{*}{ \shortstack[c]{Label \\ coverage}} & \multicolumn{6}{c}{Instances}  \\
    &  &  & Total & Max & Mean & \texttt{Things} & \texttt{Stuff} & $\frac{\#\text{\thing}}{\#\text{\stuff}}$ \\

    \midrule
    \multicolumn{9}{c}{\textit{SemanticKITTI~\cite{behley2019iccv}}} \\
    \midrule
    GT & & $98\%$ & $372478$ & $65$ & $19$ & $54\%$ & $46\%$ & $0.84$ \\
    Pseudo & & $14\%$ & $767450$ & $122$ & $40$ & $13\%$ & $87\%$ & $0.15$ \\
    \midrule
    GT & $\times$ & $98\%$ & $224104$ & $39$ & $11$ & $32\%$ & $68\%$ & $0.47$ \\
    Pseudo & $\times$ & $89\%$ & $756203$ & $120$ & $39$ & $13\%$ & $87\%$ & $0.15$ \\
    \midrule
    \multicolumn{9}{c}{\textit{nuScenes~\cite{fong21ral}}} \\
    \midrule
    GT     & & 70\% & 818971 & 119 & 29 & 63\% & 37\% & 1.70 \\
    Pseudo & & 48\% & 5594800 & 647 & 198 & 15\% & 84\% & 0.18 \\
    \bottomrule
    \noalign{\vskip 2.5mm}    
    \end{tabular}
    \label{tab:pseudo_label_stats}
\end{table*}

\PAR{Fine-grained analysis.} In \cref{tab:pseudo_label_stats-per_class}, we additionally report per-class label statistics for train and validation splits for SemanticKITTI dataset~\cite{behley2019iccv}. As can be seen, the label coverage is consistent in train and val splits ($97\%$ GT \vs $14\%$ pseudo-labels in val, and $98\%$ GT \vs $15\%$ pseudo-labels in train). Overall, the max. number of instances is larger in the train set ($65$ GT \& $122$ pseudo) as compared to the validation set ($53$ GT \& $82$ pseudo), whereas the average number of instances remains consistent for pseudo-labels, while for GT labels are lower in the train ($19$ GT \& $40$ pseudo) compared to val ($24$ GT \& $41$ pseudo). The larger number of instances reflect the most frequent classes: on the train set (Filter Frustum), the highest percentage of instances are due to \texttt{vegetation} ($27.1\%$), \texttt{building} ($16.9\%$), and \texttt{road} ($13.4\%$) classes. 
As can be confirmed in \cref{fig:sam_and_clip_image_results-kitti} and \cref{fig:sam_and_clip_image_results-nuscenes}, our pseudo-labels indeed often localize individual trees, bushes, and buildings, leading to a large number of overall instances for these classes. 

\begin{table*}[t]
\centering
\scriptsize
\caption{\textbf{Pseudo-label statistics per class on SemanticKITTI.}
We compare ground truth  label and pseudo-label statistics on SemanticKITTI~\cite{behley2019iccv} train and validation sets. 
Label coverage and instance class distributions are reported in percentage.
Due to camera visibility coverage, pseudo-labels cover a significantly smaller portion of the dataset than GT labels. This is especially prominent in the SemanticKITTI dataset ($14\%$ coverage). Pseudo-labels cover a significantly larger number of instances per scan than GT labels ($3\times$ more when only compared in the \textit{camera view frustum}), as needed to learn to segment a large variety of objects. While GT labels treat stuff classes as a ``single instance,'' our pseudo-labels hypothesize a variety of plausible segmentations of \stuff classes, leading to a higher percentage of \stuff class labels. We use a semantic oracle to report per-class statistics on pseudo-labels%
}
\setlength{\tabcolsep}{1.4pt} %
\begin{tabular}{ll|ccZ||ccccccccccccccccccc}
\toprule
& Label & \rotatebox{90}{Coverage (\%) } & \rotatebox{90}{Max / Avg. Inst.} & \rotatebox{90}{Num. Inst.} &\rotatebox{90}{\texttt{car }} & \rotatebox{90}{\texttt{bicycle }} & \rotatebox{90}{\texttt{motorcycle }} & \rotatebox{90}{\texttt{truck }} & \rotatebox{90}{\texttt{other-vehicle }} & \rotatebox{90}{\texttt{person }} & \rotatebox{90}{\texttt{bicyclist }} & \rotatebox{90}{\texttt{motorcyclist }} & \rotatebox{90}{\texttt{road }} & \rotatebox{90}{\texttt{parking }} & \rotatebox{90}{\texttt{sidewalk }} & \rotatebox{90}{\texttt{other-ground }} & \rotatebox{90}{\texttt{building }} & \rotatebox{90}{\texttt{fence }} & \rotatebox{90}{\texttt{vegetation }} & \rotatebox{90}{\texttt{trunk }} & \rotatebox{90}{\texttt{terrain }} & \rotatebox{90}{\texttt{pole }} & \rotatebox{90}{\texttt{traffic-sign }} \\
\midrule
\multicolumn{24}{c}{\textit{Full point cloud}} \\
\cmidrule{2-24}
\multirow{6}{*}{\rotatebox[origin=c]{90}{Val}}  
& GT & $97$ & $53/24$ & $97889$ & $46.9$ & $4.2$ & $1.3$ & $0.3$ & $3.2$ & $4.5$ & $1.5$ & $0.2$ & $4.2$ & $1.4$ & $4.2$ & $0.9$ & $4.2$ & $3.9$ & $4.2$ & $4.0$ & $4.2$ & $4.2$ & $2.9$ \\
& Pseudo & $14$ & $82/41$ & $170348$ & $15.6$ & $0.2$ & $0.2$ & $0.3$ & $1.0$ & $0.6$ & $0.5$ & $0.0$ & $12.5$ & $0.8$ & $9.0$ & $0.2$ & $15.1$ & $2.2$ & $27.6$ & $2.6$ & $9.1$ & $2.0$ & $0.6$ \\
\cmidrule{2-24}
\multicolumn{24}{c}{\textit{Camera view frustum}} \\
\cmidrule{2-24}
& GT & $97$ & $27/13$ & $55554$ & $31.4$ & $2.2$ & $0.8$ & $0.2$ & $1.9$ & $2.8$ & $1.3$ & $0.1$ & $7.3$ & $1.9$ & $7.3$ & $0.5$ & $6.7$ & $5.2$ & $7.3$ & $6.3$ & $6.9$ & $6.8$ & $3.1$ \\
& Pseudo & $88$ & $81/41$ & $168649$ & $15.6$ & $0.2$ & $0.2$ & $0.3$ & $1.0$ & $0.6$ & $0.5$ & $0.0$ & $12.5$ & $0.8$ & $9.0$ & $0.2$ & $15.0$ & $2.2$ & $27.6$ & $2.6$ & $9.1$ & $2.0$ & $0.6$ \\
\midrule
\midrule
\multicolumn{24}{c}{\textit{Full point cloud}} \\
\cmidrule{2-24}
\multirow{6}{*}{\rotatebox[origin=c]{90}{Train}} & GT & $98$ & $65/19$ & $372478$ & $44.8$ & $2.0$ & $1.0$ & $0.7$ & $2.1$ & $2.5$ & $0.5$ & $0.1$ & $5.1$ & $2.1$ & $4.8$ & $1.4$ & $4.6$ & $5.0$ & $5.1$ & $4.6$ & $5.0$ & $5.0$ & $3.5$ \\
& Pseudo & $15$ & $122/40$ & $767450$ & $11.4$ & $0.1$ & $0.1$ & $0.5$ & $0.6$ & $0.2$ & $0.1$ & $0.1$ & $13.4$ & $1.1$ & $8.8$ & $0.5$ & $16.8$ & $9.3$ & $27.1$ & $2.1$ & $5.3$ & $1.9$ & $0.6$ \\
\cmidrule{2-24}
\multicolumn{24}{c}{\textit{Camera view frustum}} \\
\cmidrule{2-24}
& GT & $98$ & $39/11$ & $224104$ & $27.4$ & $1.0$ & $0.6$ & $0.5$ & $1.1$ & $1.3$ & $0.3$ & $0.2$ & $8.5$ & $2.6$ & $8.0$ & $1.2$ & $6.8$ & $7.7$ & $8.5$ & $5.8$ & $7.4$ & $7.7$ & $3.2$ \\
& Pseudo & $89$ & $120/39$ & $756203$ & $11.4$ & $0.1$ & $0.1$ & $0.5$ & $0.6$ & $0.2$ & $0.1$ & $0.1$ & $13.4$ & $1.1$ & $8.8$ & $0.5$ & $16.9$ & $9.3$ & $27.1$ & $2.1$ & $5.3$ & $1.9$ & $0.6$ \\
\bottomrule
\end{tabular}
\setlength{\tabcolsep}{6pt} %
\label{tab:pseudo_label_stats-per_class}
\end{table*}

\begin{table*}[t]
    \centering
    \scriptsize
    \caption{
    \textbf{Segmentation pseudo-labels analysis.}
    This table extends \arxiv{\cref{tab:zero_shot_baselines} from} the main paper with additional metrics and baselines. 
    We report results on the SemanticKITTI validation set using the semantic oracle, as detailed in the main paper \arxiv{ (\cref{sec:experimental_setup})}. We evaluate the region of the point cloud visible in the camera (Filter Frustum). 
    }
    \scriptsize
    \begin{tabular}{l|lHlll|lHll|lHll}
    \toprule
    Model & PQ & UQ & RQ & SQ & IoU & PQ$^{Th}$ & UQ$^{Th}$  & RQ$^{Th}$ & SQ$^{Th}$  & PQ$^{St}$ & UQ$^{St}$  & RQ$^{St}$ & SQ$^{St}$ \\
    \midrule
    \midrule
    SAM (NMS area) & $46.0$ & $44.0$ & $62.3$ & $72.1$ & $58.9$ & $49.7$ & $49.6$ & $66.2$ & $73.1$ & $43.4$ & $39.9$ & $59.5$ & $71.4$ \\
    SAM (NMS score) & $28.5$ & $27.4$ & $39.3$ & $71.9$ & $45.8$ & $19.4$ & $20.8$ & $26.4$ & $73.8$ & $35.0$ & $32.1$ & $48.6$ & $70.5$ \\
    \midrule
    SAM\!+\!Erosion & $42.2$ & $39.5$ & $58.6$ & $69.4$ & $55.9$ & $45.6$ & $44.2$ & $62.3$ & $70.0$ & $39.6$ & $36.1$ & $55.9$ & $69.0$ \\
    \midrule
    SAM\!+\!DBS (filter) & $46.7$ & $41.3$ & $56.3$ & $70.3$ & $45.8$ & $76.8$ & $69.1$ & $85.5$ & $89.4$ & $24.8$ & $21.2$ & $35.1$ & $56.4$ \\
    SAM\!+\!DBS (replace) & $48.7$ & $46.6$ & $64.8$ & $73.7$ & $59.8$ & $53.1$ & $53.2$ & $69.2$ & $75.1$ & $45.4$ & $41.8$ & $61.6$ & $72.7$ \\
    \midrule
    \midrule
    \sal & $70.7$ & & $85.6$ & $81.9$ & $79.7$ & $75.4$ & & $87.1$ & $86.4$ & $67.3$ & & $84.4$ & $78.7$ \\

    \bottomrule
    \end{tabular}
    \label{tab:label_validation__detailed}
\end{table*}

\begin{table*}[t]
    \centering
    \scriptsize
    \caption{
    \textbf{Segmentation pseudo-labels ablation - per class.}
    This table extends \arxiv{\cref{tab:zero_shot_baselines} from} the main paper with per-class PQ scores. 
    We report results on the SemanticKITTI validation set using the semantic oracle, as detailed in the main paper \arxiv{ (\cref{sec:experimental_setup})}.
    We evaluate the region of the point cloud visible in the camera (Filter Frustum). 
    }
    \setlength{\tabcolsep}{2.0pt}
    \resizebox{\textwidth}{!}{
    \begin{tabular}{l|HHccccccccccccccccccc}
    \toprule
    Model & \rotatebox{90}{\texttt{all}} & \rotatebox{90}{unlabeled} & \rotatebox{90}{\texttt{car}} & \rotatebox{90}{\texttt{bicycle}} & \rotatebox{90}{\texttt{motorcycle}} & \rotatebox{90}{\texttt{truck}} & \rotatebox{90}{\texttt{other-vehicle}} & \rotatebox{90}{\texttt{person}} & \rotatebox{90}{\texttt{bicyclist}} & \rotatebox{90}{\texttt{motorcyclist}} & \rotatebox{90}{\texttt{road}} & \rotatebox{90}{\texttt{parking}} & \rotatebox{90}{\texttt{sidewalk}} & \rotatebox{90}{\texttt{other-ground}} & \rotatebox{90}{\texttt{building}} & \rotatebox{90}{\texttt{fence}} & \rotatebox{90}{\texttt{vegetation}} & \rotatebox{90}{\texttt{trunk}} & \rotatebox{90}{\texttt{terrain}} & \rotatebox{90}{\texttt{pole}} & \rotatebox{90}{\texttt{traffic-sign}} \\
    \midrule
    \midrule
    SAM (NMS area) & $46.0$ & $0.0$ & $74.2$ & $13.4$ & $44.0$ & $77.2$ & $66.3$ & $61.9$ & $39.1$ & $21.5$ & $83.1$ & $24.7$ & $38.4$ & $27.9$ & $58.3$ & $34.5$ & $71.8$ & $26.2$ & $40.9$ & $21.1$ & $49.9$ \\
    SAM (NMS score) & $28.5$ & $0.0$ & $26.2$ & $9.5$ & $17.4$ & $15.8$ & $29.1$ & $24.4$ & $13.2$ & $20.0$ & $56.2$ & $32.5$ & $34.8$ & $14.9$ & $13.7$ & $28.0$ & $57.6$ & $33.9$ & $35.6$ & $16.3$ & $62.0$ \\
    \midrule
    SAM\!+\!Erosion & $42.2$ & $0.0$ & $72.2$ & $5.6$ & $45.0$ & $78.4$ & $65.4$ & $55.1$ & $22.9$ & $20.3$ & $82.4$ & $23.1$ & $35.9$ & $24.3$ & $54.4$ & $31.1$ & $68.8$ & $22.0$ & $37.3$ & $13.4$ & $43.3$ \\
    \midrule
    SAM\!+\!DBS (filter) & $46.7$ & $0.0$ & $84.9$ & $56.7$ & $78.5$ & $90.4$ & $78.8$ & $75.0$ & $85.9$ & $64.1$ & $26.9$ & $0.0$ & $0.0$ & $0.8$ & $51.3$ & $23.3$ & $43.6$ & $37.3$ & $1.1$ & $32.2$ & $56.5$ \\
    SAM\!+\!DBS (replace) & $48.7$ & $0.0$ & $75.8$ & $17.5$ & $50.0$ & $78.2$ & $67.3$ & $66.9$ & $42.9$ & $26.4$ & $83.1$ & $24.7$ & $38.4$ & $28.0$ & $62.2$ & $36.2$ & $71.5$ & $33.5$ & $40.9$ & $27.0$ & $54.4$ \\
    \midrule
    \midrule
    \sal                & $70.7$  & $0.0$ & $90.7$ & $65.8$ & $79.9$ & $59.5$ & $78.1$ & $89.8$ & $65.4$ & $74.1$ & $92.8$ & $24.8$ & $74.2$ & $47.4$ & $82.4$ & $59.5$ & $85.0$ & $73.3$ & $60.7$ & $75.3$ & $65.5$ \\
    \bottomrule
    \end{tabular}
    }
    \label{tab:label_validation__per_class__pq}
\end{table*}

\subsection{Pseudo-label Timing \& Effort Analysis}
\label{sec:pseudo_label_timing}

We report information and statistics on three datasets (SemanticKITTI~\cite{behley2019iccv}, Panoptic nuScenes~\cite{fong21ral} and Waymo Open~\cite{sun20CVPR}) that we pseudo-label and use to train \sal, in \cref{tab:pseudo-labeling-effort}. As can be seen, the processing time needed to process a single scan depends on multiple dataset characteristics, such as the number of cameras. In the case of the Waymo Open dataset, we pseudo-label every $10^{th}$ scan, and we do not perform DBSCAN refinement. The reason is two-fold: (i) we observe image-lidar calibration on Waymo is more accurate compared to KITTI and nuScenes, and (ii) we save processing effort/time to pseudo-label Waymo. Importantly, our pseudo-labeling setup is general: it supports various sensory setups (single camera, multi-camera) and multiple Lidar types and can cope well with datasets with different degrees of accuracy of the image-lidar calibration/synchronization.

In \cref{tab:pseudo-labeling-effort-per-scan}, we provide fine-grained per-scan timing analysis for \sal pseudo-labeling engine. The most costly component is extracting image-level segmentation masks using the segmentation foundation model (SAM~\cite{kirillov2023segment}), which needs to be performed, in general, for each camera.

\begin{table*}[t]
    \centering
    \scriptsize
    \caption{\textbf{Dataset pseudo-labeling analysis.} We report dataset information and statistics, along with pseudo-labeling effort and time analysis. We note that in SemanticKITTI, we have four cameras in total. However, only one is used for pseudo-labeling, as all cameras have (roughly) the same field of view. We report the total number of scans, as well as per-scan processing time (NVIDIA A100D-80C GPU). Processing time depends on the number of cameras, as well as individual dataset characteristics (such as camera resolution, point cloud size, density, \etc.). Finally, we report the total time needed to pseudo-label a dataset if processed sequentially (in practice, we can pseudo-label datasets in 1-3 days using a compute cluster). We note that in the case of Waymo, due to sheer size, we (i) pseudo-label only $10\%$ of point clouds and (ii) skip the postprocessing with DBSCAN.}
    \begin{tabular}{l|cc|c|ccc}
    \toprule
         \multirow{2}{*}{Dataset} & \multirow{2}{*}{\# cam.} & \multirow{2}{*}{\shortstack[c]{Cam. \\ cov. ($^{\circ}$)}} & \multirow{2}{*}{Lidar} &  
         \multirow{2}{*}{\shortstack[c]{\# scans \\ Total / Pseudo-lab.}}
         &  
        \multirow{2}{*}{\shortstack[c]{Time (s) \\ Per scan}}
         & 
        \multirow{2}{*}{\shortstack[c]{Time (days) \\ Total}} \\
         & & & & & & \\
         \midrule
         SemanticKITTI& $1$ &  90 &  Velodyne HDL-64E&  $43592/43592$ & $32$ & $16$\\
         Panoptic nuScenes& $6$ & $360$ &  Velodyne HDL-32E&  $40157/40157$ &  $129$ & $59$\\
        Waymo Open& $5$ & $270$ & Waymo Proprietary& $227101/22710$ & $92$ & $24$ \\
    \bottomrule
    \end{tabular}
    \label{tab:pseudo-labeling-effort}
\end{table*}

\begin{table*}[t]
    \centering
    \scriptsize
    \caption{\textbf{Per-scan pseudo-labeling timing analysis.} We report fine-grained per-dataset timing analysis for our pseudo-labeling system. In particular, we report individual timings for core components of \sal pseudo-label engine: (i) running segmentation foundation model (SAM~\cite{kirillov2023segment}) in the image domain, (ii) extracting corresponding CLIP features~\cite{ding2023open, radford2021learning}, (iii) image-to-Lidar unprojection, and finally, (iv) DBSCAN refinement. As can be seen, the most costly step is due to SAM; this step could be reduced in the future using newer and more time-efficient variants of SAM. DBSCAN is the second bottleneck, however, it is only executed once per point cloud, and does not depend on the number of cameras. The analysis was conducted using NVIDIA A100D-80C GPU.}
    \begin{tabular}{l|c|ccccc}
    \toprule
        \multirow{2}{*}{Dataset}& \multirow{2}{*}{\# cam.} & \multicolumn{5}{c}{Time (s)} \\
        & & Scan & SAM & CLIP & Unprojection & DBSCAN \\
        \midrule
        SemanticKITTI & $1$ & $31.9$ &  $6.7$ &  $1.8$ &  $0.9$ & $22.2$ \\
        Panoptic nuScenes& $6$ & $129.2$ & $56.4$ & $14.8$ & $14.34$ & $41.1$\\
        Waymo Open & $5$ & $92$ & $56.1$ & $16.0$ & $17.5$& --\\
    \bottomrule
    \end{tabular}
    \label{tab:pseudo-labeling-effort-per-scan}
\end{table*}

\subsection{Qualitative Analysis}

Both SemanticKITTI~\cite{behley2019iccv} and nuScenes~\cite{fong21ral} do not provide ground truth annotations in 2D.
To illustrate the performance of the foundation models we utilize for pseudo-labeling in the image domain, we visualize predicted masks and dataset class vocabularies from SAM~\cite{kirillov2023segment} and CLIP~\cite{radford2021learning}. 
In \cref{fig:sam_and_clip_image_results-kitti}, we show the single (front) camera view corresponding to the first scan of four different SemanticKITTI sequences.
For nuScenes~\cref{fig:sam_and_clip_image_results-nuscenes} and Waymo Open, we focus on the first scan of a single sequence and show all camera views. 
As can be seen, in all datasets, the output contains many correct segmentations and classifications, \eg, for the \texttt{road}, \texttt{car}, or \texttt{vegetation} classes. Moreover, the class-agnostic SAM masks introduce many smaller instances by segmenting individual road markings. However, without any corresponding points in the cloud, such segments are filtered by the unprojection/transfer to Lidar. 

While the SAM masks are generally correct, noisy CLIP predictions confirm there is room for improvement. For example, the sky is always misclassified since it is not part of the class vocabulary.
As shown in the main paper \arxiv{(\cref{tab:ablation_lin_probe_clip_semantics})}, the distillation of SAM and CLIP to 3D yields an analogous behavior of our \sal model. 
It should be noted that the clip predictions shown in~\cref{fig:sam_and_clip_image_results-kitti} and~\cref{fig:sam_and_clip_image_results-nuscenes} are only for visualization purposes.
We do not directly transfer class labels to 3D, but only the CLIP embeddings. 

\subsection{Pseudo-label Quality}

We additionally report per-class pseudo-label results in \cref{tab:label_validation__per_class__pq}. As can be seen, \textit{DBSCAN replace} consistently improves over tuned SAM variant on \thing classes (\eg, for \texttt{car} class, $+1.6$ PQ, \texttt{person} class $+4.8$ PQ). While \textit{DBSCAN filter} leads to remarkable improvements on some classes (\eg, for \texttt{car} $+10.7$, \texttt{bicycle} $+43.2$), it severely degrades stuff classes (\eg, $-39.8$ for \texttt{terrain} class). On the other hand, \textit{DBSCAN replace} either does not affect \stuff classes or leads to improvements (\eg, \texttt{building} $+3.9$).

\section{Model Ablations}
\label{sec:model_ablations_app}

\subsection{Zero-shot Classification}
\label{sec:zero_shot_classification_app}
The \sal model performs zero-shot classification by matching text prompts with predicted CLIP tokens.
To predict tokens only with Lidar input, our label engine generates pairs of Lidar segments and corresponding CLIP image tokens.
Training on these pairs distills the CLIP image encoder into our model.
The first row of~\cref{tab:ablation_semantics} demonstrates a different approach where we train on pairs of Lidar segments and class labels obtained by directly prompting the CLIP image encoder.
Examples of these labels are visualized in~\cref{fig:sam_and_clip_image_results-kitti}. 
The difference ($- 1.5$) between the first and third row in~\cref{tab:ablation_semantics} indicates the performance drop of the distillation into our model.
This further demonstrates the effectiveness of our framework.
In particular, our insufficiencies with respect to zero-shot classification are not caused by our design choices in 3D but merely a reflection of the same limitations in the image domain.

\PAR{Text prompt engineering.}
The CLIP tokens predicted by our model can be prompted with any arbitrary text.
This allows our model to classify potentially \emph{any} object.
The text prompts must adhere to the respective class vocabulary to perform zero-shot classification on an annotated dataset. 
In~\cref{tab:ablation_semantics}, we demonstrate the performance boost from engineering the set of text prompts, \eg, by adding similar class terms to each class set (see~\cref{tab:text_prompts}).
In particular, for super-classes, where we add all default class names to the respective super set, our engineering yields a huge gain of $+23.1$.
This is due to the ambiguous nature of the super class names, \eg, \texttt{object} or \texttt{structure}.

\begin{table*}[t]
    \centering
    \caption{
        \textbf{\sal per class PQ results on the SemanticKITTI and nuScenes validation sets.}
        We report metrics for zero-shot (ZS) and linear probing (LP) supervision.
    }
    \scriptsize
    \setlength{\tabcolsep}{2.0pt}
    \resizebox{\textwidth}{!}{
    \begin{tabular}{HHl | HHHH HHH HHH H c | ccccccccccccccccccc}
        \toprule
        & & & \multicolumn{31}{c}{SemanticKITTI~\cite{behley2019iccv}} \\
        \cmidrule{4-34}
        & Method & \rotatebox{90}{Supervision} & PQ & PQ\textsuperscript{\textdagger} & RQ & SQ & PQ\textsuperscript{Th} & RQ\textsuperscript{Th} & SQ\textsuperscript{Th} & PQ\textsuperscript{St} & RQ\textsuperscript{St} & SQ\textsuperscript{St} & mIoU & \rotatebox{90}{\texttt{all}} & \rotatebox{90}{\texttt{car}} & \rotatebox{90}{\texttt{bicycle}} & \rotatebox{90}{\texttt{motorcycle}} & \rotatebox{90}{\texttt{truck}} & \rotatebox{90}{\texttt{other-vehicle}} & \rotatebox{90}{\texttt{person}} & \rotatebox{90}{\texttt{bicyclist}} & \rotatebox{90}{\texttt{motorcyclist}} & \rotatebox{90}{\texttt{road}} & \rotatebox{90}{\texttt{parking}} & \rotatebox{90}{\texttt{sidewalk}} & \rotatebox{90}{\texttt{other-ground}} & \rotatebox{90}{\texttt{building}} & \rotatebox{90}{\texttt{fence}} & \rotatebox{90}{\texttt{vegetation}} & \rotatebox{90}{\texttt{trunk}} & \rotatebox{90}{\texttt{terrain}} & \rotatebox{90}{\texttt{pole}} & \rotatebox{90}{\texttt{traffic-sign}} \\
        \midrule

        & \sal (Ours)  & ZS & $X$ & $X$ & $X$ & $X$ & $X$ & $X$ & $X$ & $X$ & $X$ & $X$ & $X$ &
            \DONE{$24.8$} & \DONE{$82.3$} & \DONE{$22.3$} & \DONE{$10.9$} & \DONE{$9.5$} & \DONE{$9.2$} & \DONE{$5.4$} & \DONE{$0.0$} & \DONE{$0.0$} & \DONE{$67.0$} & \DONE{$0.2$} & \DONE{$27.6$} & \DONE{$0.1$} & \DONE{$33.2$} & \DONE{$3.6$} & \DONE{$80.7$} & \DONE{$16.9$} & \DONE{$32.4$} & \DONE{$45.8$} & \DONE{$24.9$} \\

        & \sal (Ours)  & LP & $33.1$ & $39.3$ & $41.9$ & $68.3$ & $31.8$ & $36.1$ & $74.7$ & $34.1$ & $46.1$ & $63.7$ & $40.0$ & $33.1$ & $78.0$ & $1.5$ & $25.0$ & $25.8$ & $20.3$ & $41.5$ & $62.6$ & $0.0$ & $79.7$ & $17.6$ & $33.2$ & $0.0$ & $74.2$ & $13.3$ & $74.0$ & $21.5$ & $34.4$ & $32.9$ & $9.8$\\

        \midrule
        & & & \multicolumn{31}{c}{nuScenes~\cite{fong21ral}} \\
        \cmidrule{4-34}
        &  & & PQ & PQ\textsuperscript{\textdagger} & RQ & SQ & PQ\textsuperscript{Th} & RQ\textsuperscript{Th} & SQ\textsuperscript{Th} & PQ\textsuperscript{St} & RQ\textsuperscript{St} & SQ\textsuperscript{St} & mIoU & \rotatebox{90}{\texttt{all}} & \rotatebox{90}{\texttt{barrier}} & \rotatebox{90}{\texttt{bicycle}} & \rotatebox{90}{\texttt{bus}} & \rotatebox{90}{\texttt{car}} & \rotatebox{90}{\texttt{construction\_vehicle}} & \rotatebox{90}{\texttt{motorcycle}} & \rotatebox{90}{\texttt{pedestrian}} & \rotatebox{90}{\texttt{traffic\_cone}} & \rotatebox{90}{\texttt{trailer}} & \rotatebox{90}{\texttt{truck}} & \rotatebox{90}{\texttt{driveable\_surface}} & \rotatebox{90}{\texttt{other\_flat}} & \rotatebox{90}{\texttt{sidewalk}} & \rotatebox{90}{\texttt{terrain}} & \rotatebox{90}{\texttt{manmade}} & \rotatebox{90}{\texttt{vegetation}} \\
        \midrule

        & \sal (Ours)  & ZS & $X$ & $X$ & $X$ & $X$ & $X$ & $X$ & $X$ & $X$ & $X$ & $X$ & $X$ & 
            \DONE{$38.4$} & \DONE{$0.7$} & \DONE{$53.4$} & \DONE{$46.6$} & \DONE{$82.2$} & \DONE{$17.9$} & \DONE{$52.5$} & \DONE{$50.6$} & \DONE{$35.4$} & \DONE{$30.0$} & \DONE{$47.2$} & \DONE{$57.6$} & \DONE{$0.1$} & \DONE{$13.5$} & \DONE{$29.2$} & \DONE{$29.7$} & \DONE{$67.5$} \\

        & \sal (Ours)  & LP & $40.1$ & $46.3$ & $50.5$ & $76.4$ & $45.5$ & $53.8$ & $82.7$ & $34.6$ & $47.3$ & $70.1$ & $39.0$ & $40.1$ & $3.2$ & $17.6$ & $53.8$ & $77.3$ & $15.8$ & $40.8$ & $83.2$ & $27.2$ & $28.2$ & $47.7$ & $67.1$ & $23.1$ & $10.2$ & $15.8$ & $64.2$ & $65.8$\\

    \bottomrule
    \end{tabular} 
    }
    \label{tab:kitti_nuscenes-baselines-per_class}
    
\end{table*}

\section{Per-class Results}
\label{sec:qual_pre_class_app}
In~\cref{tab:kitti_nuscenes-baselines-per_class}, we report per-class PQ results of the \sal model on SemanticKITTI~\cite{behley2019iccv} and nuScenes~\cite{fong21ral}.
These results correspond to the default class evaluations shown in the main paper \arxiv{(\cref{tab:semkitti_panoseg})}.
For \texttt{other-X} classes which are defined in delimitation to other classes, \eg, \texttt{other-vehicle} than \texttt{car}, our zero-shot model suffers from the ambiguous class prompt (see~\cref{sec:txt_prompt_eng_app}).
Linear probing improves across most classes except, for example, \texttt{bicycle}.
The performance drop can be explained by the different object-notion between our pseudo and the ground truth labels.

\section{Qualitative Results}
\label{sec:qualitative-appendix}

\begin{figure}[ht]
    \centering
    \includegraphics[width=\textwidth]{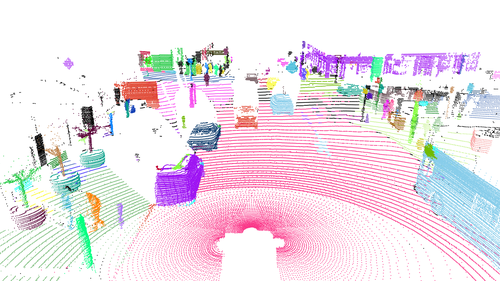}
    
    \begin{subfigure}{0.32\textwidth}
        \centering
        \includegraphics[width=\linewidth]{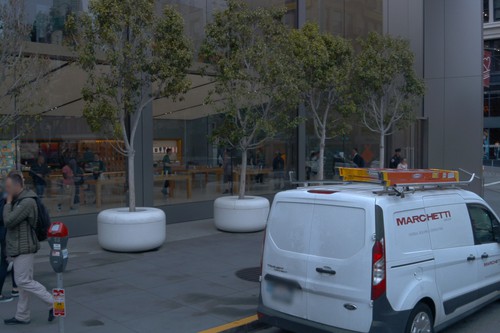}
        \caption{Front-left camera}
        \label{fig:sub2}
    \end{subfigure}
    \hfill
   \begin{subfigure}{0.32\textwidth}
        \centering
        \includegraphics[width=\linewidth]{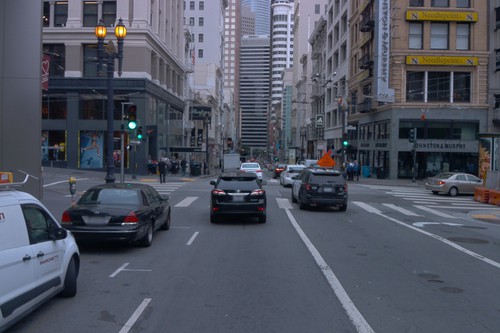}
        \caption{Front camera}
        \label{fig:sub1}
    \end{subfigure}
    \hfill
    \begin{subfigure}{0.32\textwidth}
        \centering
        \includegraphics[width=\linewidth]{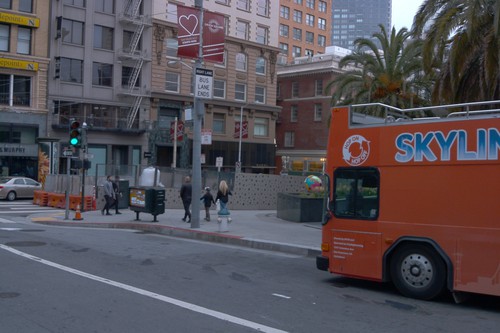}
        \caption{Front-right camera}
        \label{fig:sub3}
    \end{subfigure}
    \hfill
\caption{
\textbf{\textbf{Class-agnostic segmentation on Waymo Open~\cite{sun20CVPR}} from first-person perspective.} We visually outline the Lidar point cloud, where points are colored according to estimated instance IDs, estimated by \sal. We show corresponding camera views (not used for inference) for reference. As can be seen, \sal accurately segments a large variety of objects, including \texttt{parking meters}, \texttt{potted trees} (\texttt{pots} as well as \texttt{trees}), \texttt{rooftop ladder}, \texttt{water hydrant}, \texttt{post box}, \texttt{traffic cone}, \texttt{traffic barrier}, and more. Canonical objects, such as \texttt{car}, \texttt{van}, \texttt{bus}, and \texttt{pedestrian} are segmented as well. This class-agnostic segmentation is a basis for zero-shot classification.
}
\label{fig:qualitative-segmentation}
\end{figure}

\PAR{Class-agnostic segmentation.} 
In \cref{fig:qualitative-segmentation}, we provide qualitative results for class-agnostic segmentation on Waymo Open dataset. Colors encode identities (IDs) of individual instances. For reference, we provide images of corresponding camera views, even though these are not used during the inference. As can be seen, \sal learns to segment full point clouds, \thing and \stuff classes, even though supervision is only partial. Interestingly, \sal segments well large structures (often classified as \stuff classes), such as \texttt{building}, \texttt{road}, and \texttt{sidewalk}. In addition to canonical \thing classes, such as \texttt{car}, \texttt{van}, \texttt{bus}, and \texttt{pedestrian}, \sal also learns a variety of classes, that are not covered in class vocabularies of existing datasets of Lidar Panoptic Segmentation (SemanticKITTI~\cite{behley2019iccv} and nuScenes~\cite{fong21ral}). Examples of such classes, segmented in \cref{fig:qualitative-segmentation}, are \texttt{parking meters}, \texttt{potted trees}, \texttt{rooftop ladder}, \texttt{water hydrant}, \texttt{post box}, \texttt{traffic cone} and \texttt{traffic barrier}.

\begin{figure}[t]
\centering
\includegraphics[width=\linewidth]{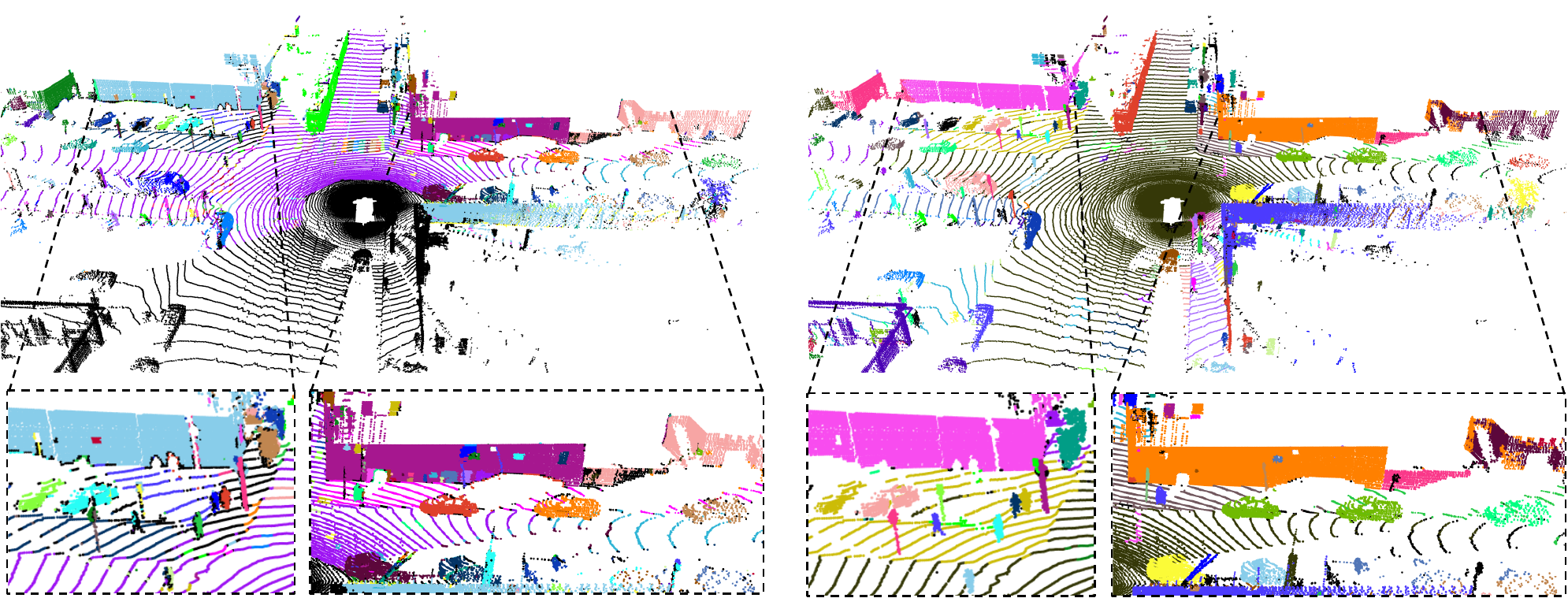}
\caption{
\textbf{\textbf{Class-Agnostic Segmentation on Waymo Open~\cite{sun20CVPR}.}} We visually compare class-agnostic segmentation results. Colors encode object instance IDs. \textit{Left}: baseline (SAM~\cite{kirillov2023segment}, unprojected to Lidar), and, \textit{right}, \sal. As can be seen, the baseline that directly lifts SAM masks to Lidar is limited to $270^{\circ}$ field of view, which overlaps with the camera ring. By contrast, \sal segments the full Lidar point cloud and is not limited by the camera coverage. Zoomed-in regions show that the baseline is sensitive to edge bleeding (\eg, see \texttt{pedestrian} and \texttt{traffic sign} masks, partially projected to the blue wall). \sal, by contrast, distills noisy SAM masks into crisp segmentation masks.
}
\label{fig:qualitative-segmentation-comparison}
\end{figure}

In \cref{fig:qualitative-segmentation-comparison}, we contrast a baseline that simply lifts SAM~\cite{Kirillov19CVPR} masks to Lidar (this baseline is evaluated in \cref{tab:label_validation__detailed}, as well as \arxiv{\cref{tab:zero_shot_baselines}} in the main paper). As can be seen in \cref{fig:qualitative-segmentation-comparison}, \textit{left}, Waymo Open provides $270^{\circ}$ coverage (four camera views), leaving a ``blind spot'' behind the vehicle. This is an inherent limitation of the baseline that requires camera views for the inference. By contrast, \sal (\cref{fig:qualitative-segmentation-comparison}, \textit{right}) learns to distill such image-generated pseudo-labels to a full \lps model. Therefore, it segments full $360^{\circ}$ point clouds and does not require camera views during the inference (only during the model training).

For further insights, we zoom in on certain regions of point clouds. As can be seen in \cref{fig:qualitative-segmentation-comparison}, \textit{left}, masks transferred from images to Lidar often lead to bleeding edges (\eg, red and blue pedestrians ``bleed'' to the pale blue wall), and thinner structures are often not segmented well due to non-perfect calibration and rolling shutter nature of the Lidar sensor. While such issues can be (partially) mitigated by postprocessing via density-based clustering (we refer to \arxiv{\cref{tab:label_validation__detailed} and \cref{tab:zero_shot_baselines} in} the main paper for quantitative analysis), we show in \cref{fig:qualitative-segmentation-comparison}, \textit{right}, results obtained with \sal model, trained directly on SAM-transferred masks (without DBSCAN as postprocessing). Remarkably, the distilled model does not suffer from these artifacts.

\begin{figure}[htp]
    \centering

    \begin{subfigure}{.48\textwidth}
        \centering
        \includegraphics[width=\linewidth]{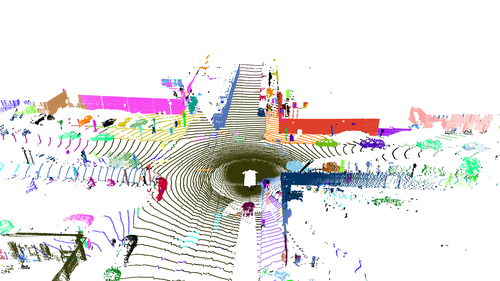}
        \caption{Class-agnostic segmentation}
    \end{subfigure}
    \hfill %
    \begin{subfigure}{.48\textwidth}
        \centering
        \includegraphics[width=\linewidth]{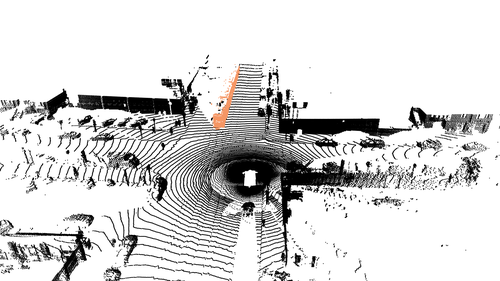}
        \caption{Prompt: \{\texttt{tram}\}}
    \end{subfigure}

    \begin{subfigure}{.48\textwidth}
        \centering
        \includegraphics[width=\linewidth]{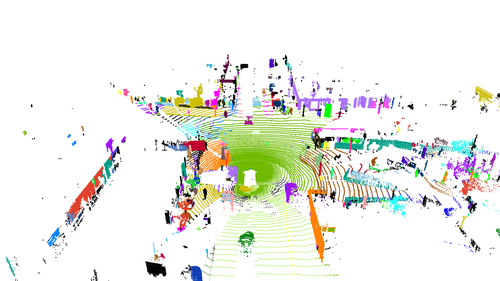}
        \caption{Class-agnostic segmentation}
    \end{subfigure}
    \begin{subfigure}{.48\textwidth}
        \centering
        \includegraphics[width=\linewidth]{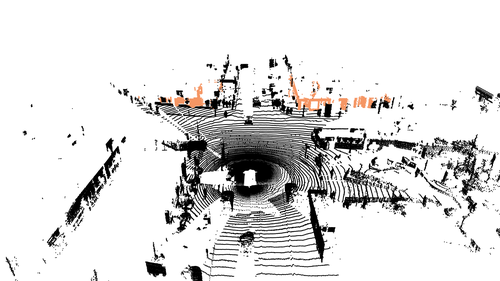}
        \caption{Prompt: \{\texttt{store front}\}}
    \end{subfigure}

    \begin{subfigure}{.48\textwidth}
        \centering
        \includegraphics[width=\linewidth]{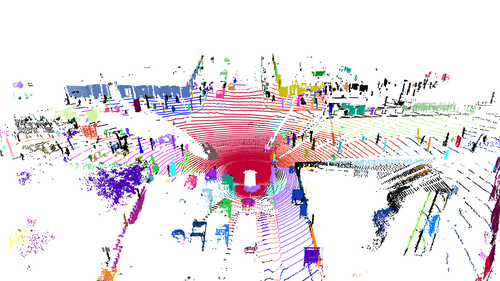}
        \caption{Class-agnostic segmentation}
    \end{subfigure}
    \hfill %
    \begin{subfigure}{.48\textwidth}
        \centering
        \includegraphics[width=\linewidth]{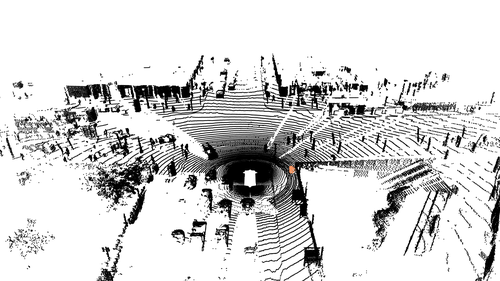}
        \caption{Prompt: \{\texttt{trash bin}\}}
    \end{subfigure}

    \begin{subfigure}{.48\textwidth}
        \centering
        \includegraphics[width=\linewidth]{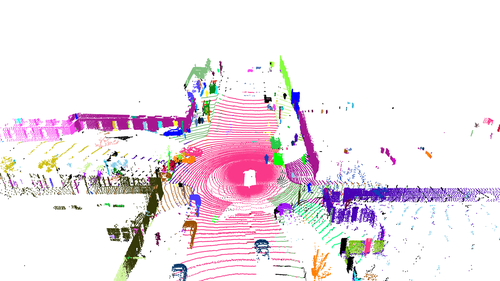}
        \caption{Class-agnostic segmentation}
    \end{subfigure}
    \hfill %
    \begin{subfigure}{.48\textwidth}
        \centering
        \includegraphics[width=\linewidth]{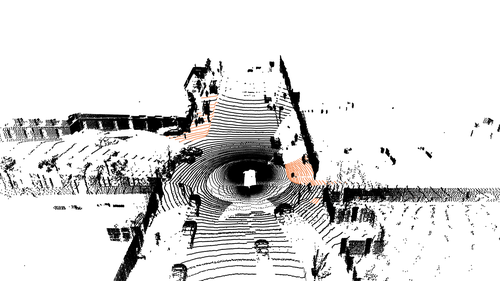}
        \caption{Prompt: \{\texttt{curb}\}}
    \end{subfigure}
    
    \caption{\textbf{Zero-shot per-class prompting on Waymo Open~\cite{sun20CVPR}.} \sal predicts a set of object instances (\textit{left}), along with their objectness scores and \textit{distilled} CLIP~\cite{radford2021learning} features. We can use text prompts and query these instances for specific classes specified as prompts. On the \textit{right}, we highlight several such examples that are outside of class-vocabularies of SemanticKITTI, nuScenes, and Waymo Open datasets. As can be seen on the left, a basis for such zero-shot prompting is accurate and, importantly, \textit{diverse} class-agnostic segmentation.}
    \label{fig:zero-shot-prompting}
\end{figure}

\PAR{Zero-shot prompting.}
In \cref{fig:zero-shot-prompting}, we highlight the capability of \sal for prompting specific semantic classes. On the \textit{left} side, we visualize class-agnostic segmentations of point clouds, and on the \textit{right}, we highlight prompts and highlight segmented regions. We show two \thing classes (\texttt{tram} and \texttt{trash bin}) and two \stuff lasses (\texttt{store front} and \texttt{curb}). A basis for such zero-shot prompting is our class-agnostic segmentation model, which segments input point clouds into a set of segmented objects. \sal model predicts for each segmented object CLIP feature token that we use for zero-shot prompting (for details, we refer to \arxiv{\cref{sec:method} of} the main paper).

\begin{figure*}[t]
    \centering
    \begin{tikzpicture}
        \draw (0, 0) node[inner sep=0] {\includegraphics[width=0.32\linewidth]{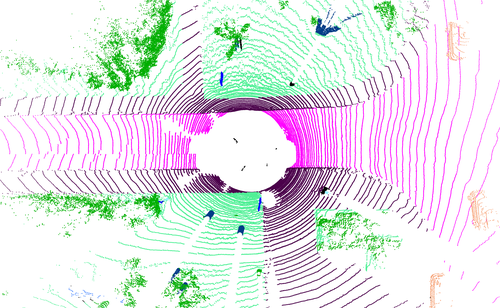}};
        \draw (0, 0) node {\huge \color{black} GT};
    \end{tikzpicture}
    \begin{tikzpicture}
        \draw (0, 0) node[inner sep=0] {\includegraphics[width=0.32\linewidth]{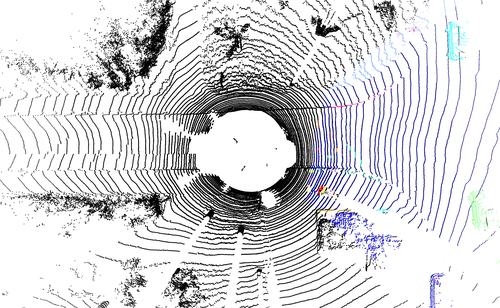}};
        \draw (0, 0) node {\huge \color{black} Pseudo};
    \end{tikzpicture}
    \begin{tikzpicture}
        \draw (0, 0) node[inner sep=0] {\includegraphics[width=0.32\linewidth]{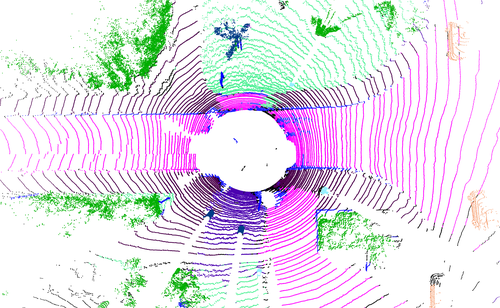}};
        \draw (0, 0) node {\huge \color{black} Output};
    \end{tikzpicture}

    \begin{tikzpicture}
        \draw (0, 0) node[inner sep=0] {\includegraphics[width=0.32\linewidth]{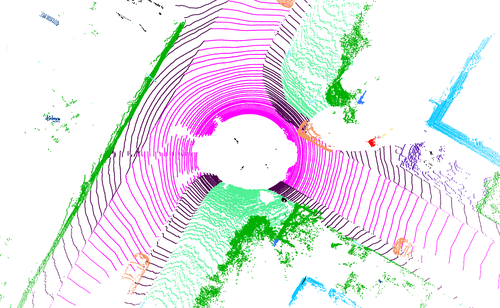}};
    \end{tikzpicture}
    \begin{tikzpicture}
        \draw (0, 0) node[inner sep=0] {\includegraphics[width=0.32\linewidth]{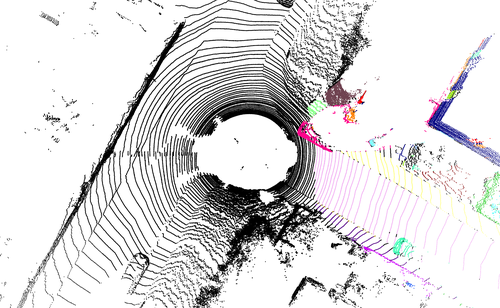}};
    \end{tikzpicture}
    \begin{tikzpicture}
        \draw (0, 0) node[inner sep=0] {\includegraphics[width=0.32\linewidth]{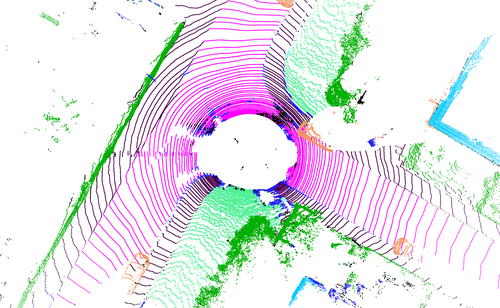}};
    \end{tikzpicture}

    \begin{tikzpicture}
        \draw (0, 0) node[inner sep=0] {\includegraphics[width=0.32\linewidth]{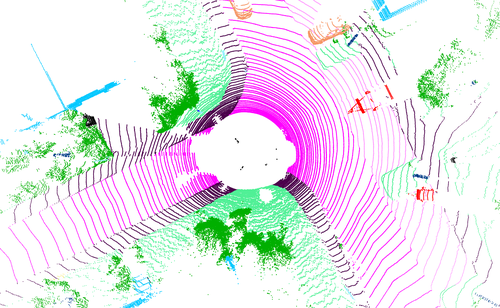}};
    \end{tikzpicture}
    \begin{tikzpicture}
        \draw (0, 0) node[inner sep=0] {\includegraphics[width=0.32\linewidth]{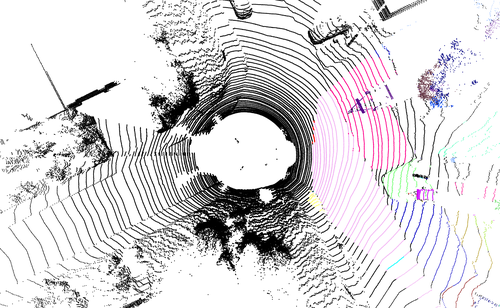}};
    \end{tikzpicture}
    \begin{tikzpicture}
        \draw (0, 0) node[inner sep=0] {\includegraphics[width=0.32\linewidth]{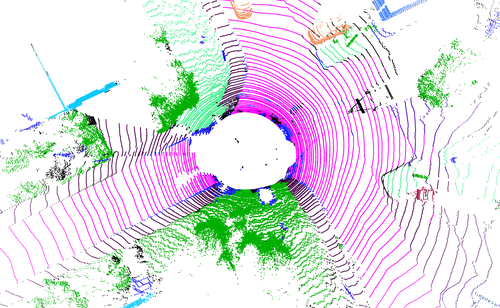}};
    \end{tikzpicture}

    \begin{tikzpicture}
        \draw (0, 0) node[inner sep=0] {\includegraphics[width=0.32\linewidth]{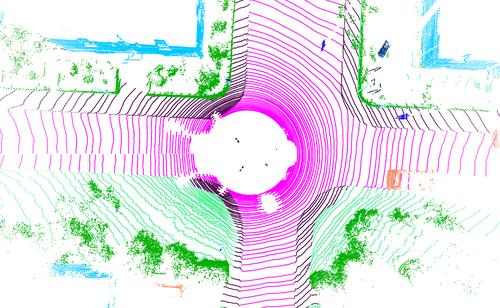}};
    \end{tikzpicture}
    \begin{tikzpicture}
        \draw (0, 0) node[inner sep=0] {\includegraphics[width=0.32\linewidth]{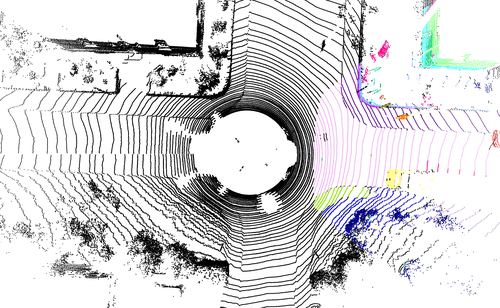}};
    \end{tikzpicture}
    \begin{tikzpicture}
        \draw (0, 0) node[inner sep=0] {\includegraphics[width=0.32\linewidth]{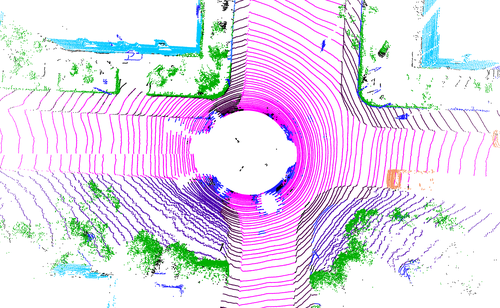}};
    \end{tikzpicture}

    \begin{tikzpicture}
        \draw (0, 0) node[inner sep=0] {\includegraphics[width=0.32\linewidth]{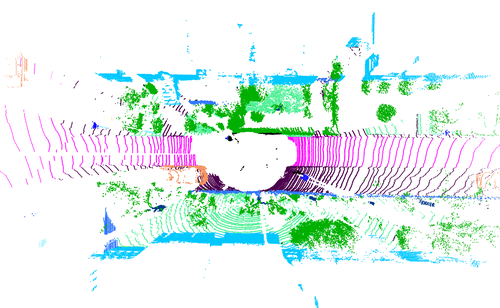}};
    \end{tikzpicture}
    \begin{tikzpicture}
        \draw (0, 0) node[inner sep=0] {\includegraphics[width=0.32\linewidth]{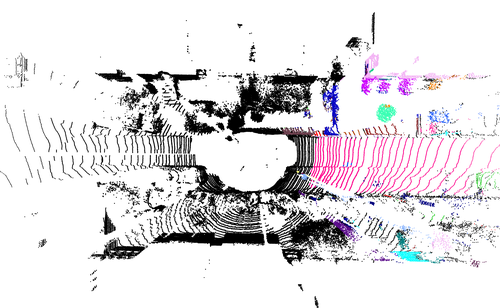}};
    \end{tikzpicture}
    \begin{tikzpicture}
        \draw (0, 0) node[inner sep=0] {\includegraphics[width=0.32\linewidth]{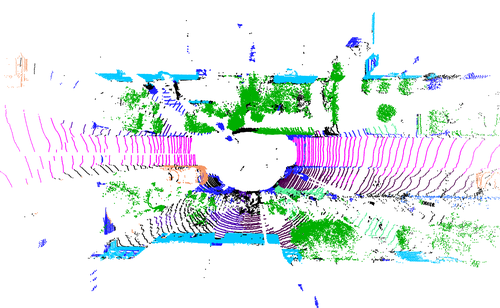}};
    \end{tikzpicture}

    \begin{tikzpicture}
        \draw (0, 0) node[inner sep=0] {\includegraphics[width=0.32\linewidth]{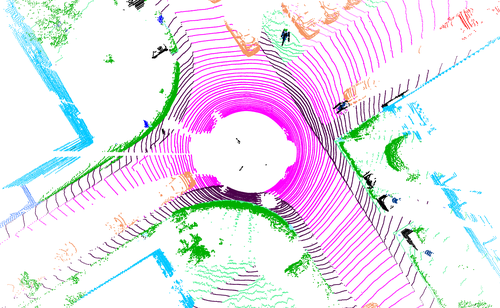}};
    \end{tikzpicture}
    \begin{tikzpicture}
        \draw (0, 0) node[inner sep=0] {\includegraphics[width=0.32\linewidth]{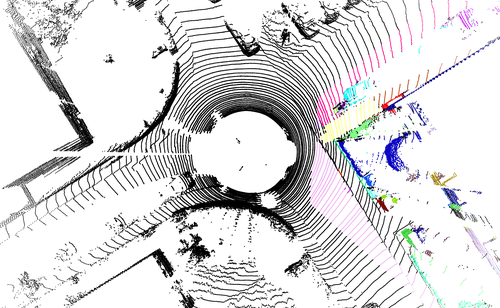}};
    \end{tikzpicture}
    \begin{tikzpicture}
        \draw (0, 0) node[inner sep=0] {\includegraphics[width=0.32\linewidth]{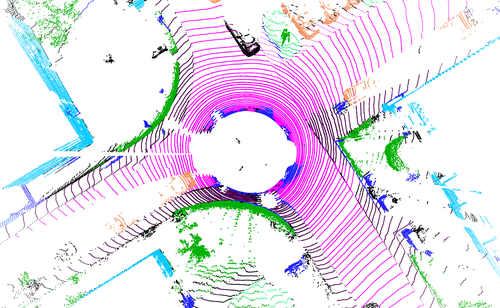}};
    \end{tikzpicture}
    
    \caption{
    \textbf{Qualitative results for SemanticKITTI.}
    We visualize ground truth (GT), pseudo labels, and our model output for several scans of validation sequence \emph{08} of SemanticKITTI~\cite{behley2019iccv}.
    While GT and our output display semantics, the class-agnostic pseudo labels show instances. 
    }
    \label{fig:qual_gt_pseudo_outputs-kitti}
\end{figure*}
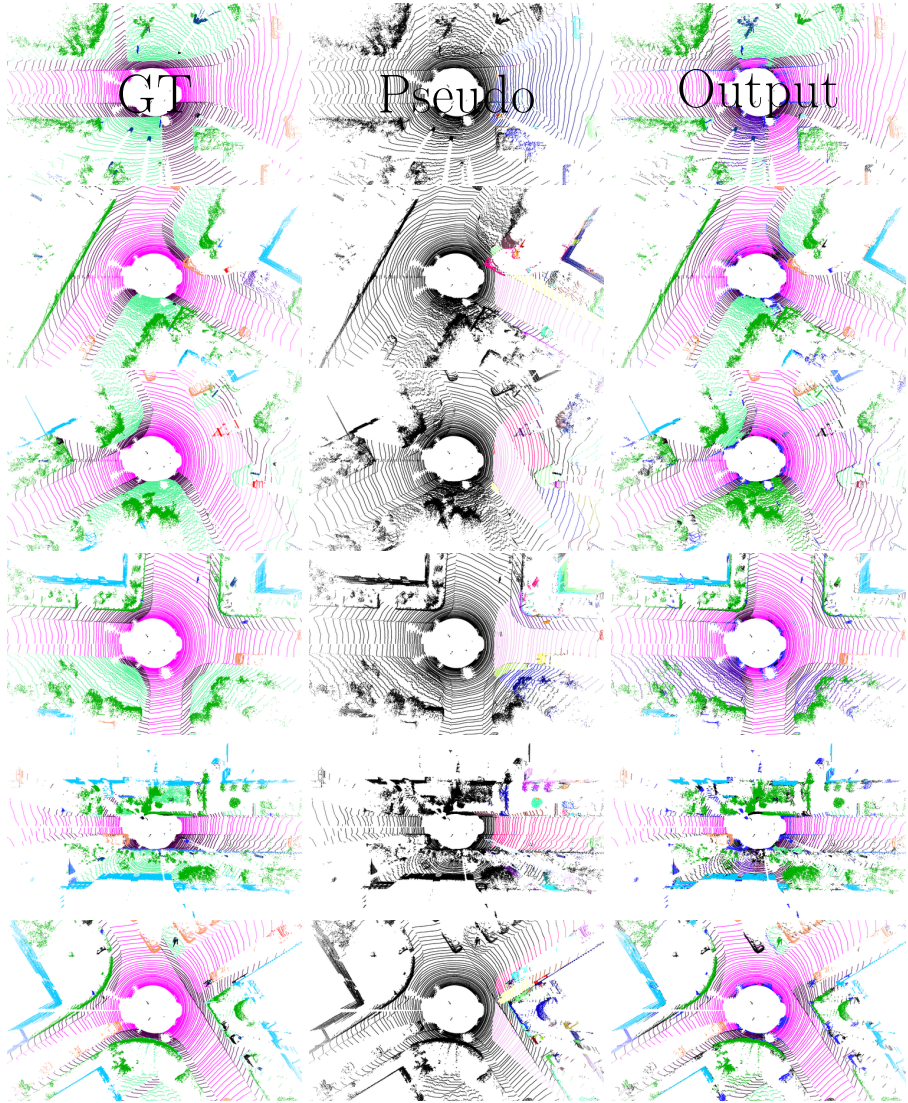

\begin{figure*}[t]
    \centering
    \begin{tikzpicture}
        \draw (0, 0) node[inner sep=0] {\includegraphics[width=0.32\linewidth]{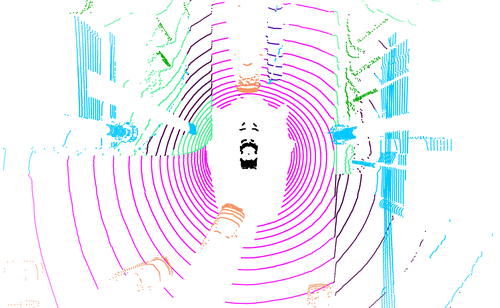}};
        \draw (0, 0) node {\huge \color{black} GT};
    \end{tikzpicture}
    \begin{tikzpicture}
        \draw (0, 0) node[inner sep=0] {\includegraphics[width=0.32\linewidth]{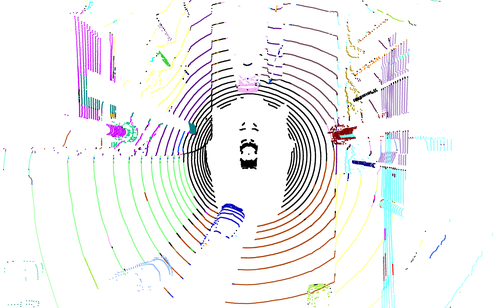}};
        \draw (0, 0) node {\huge \color{black} Pseudo};
    \end{tikzpicture}
    \begin{tikzpicture}
        \draw (0, 0) node[inner sep=0] {\includegraphics[width=0.32\linewidth]{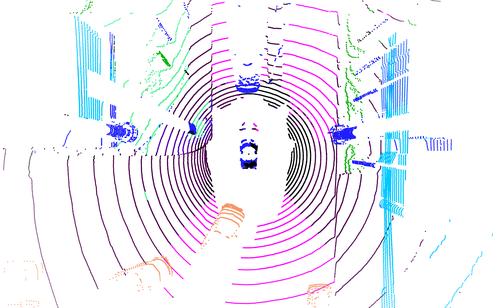}};
        \draw (0, 0) node {\huge \color{black} Output};
    \end{tikzpicture}

    \begin{tikzpicture}
        \draw (0, 0) node[inner sep=0] {\includegraphics[width=0.32\linewidth]{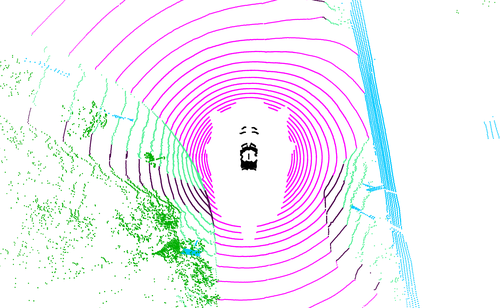}};
    \end{tikzpicture}
    \begin{tikzpicture}
        \draw (0, 0) node[inner sep=0] {\includegraphics[width=0.32\linewidth]{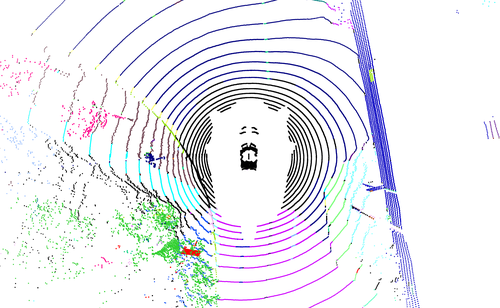}};
    \end{tikzpicture}
    \begin{tikzpicture}
        \draw (0, 0) node[inner sep=0] {\includegraphics[width=0.32\linewidth]{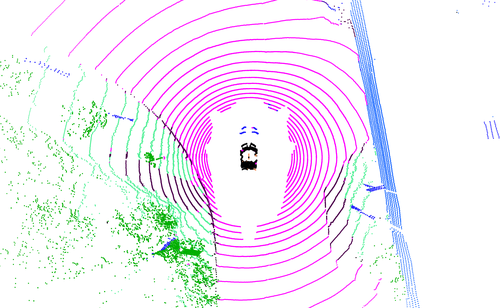}};
    \end{tikzpicture}

    \begin{tikzpicture}
        \draw (0, 0) node[inner sep=0] {\includegraphics[width=0.32\linewidth]{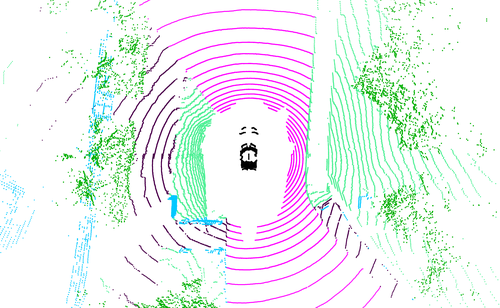}};
    \end{tikzpicture}
    \begin{tikzpicture}
        \draw (0, 0) node[inner sep=0] {\includegraphics[width=0.32\linewidth]{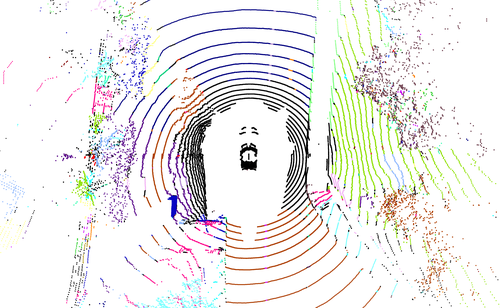}};
    \end{tikzpicture}
    \begin{tikzpicture}
        \draw (0, 0) node[inner sep=0] {\includegraphics[width=0.32\linewidth]{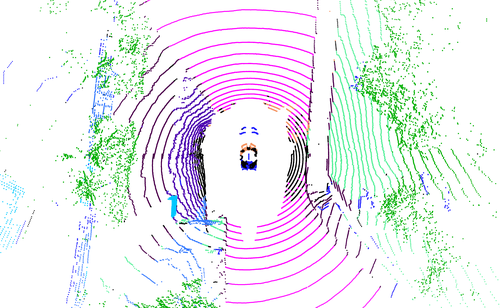}};
    \end{tikzpicture}

    \begin{tikzpicture}
        \draw (0, 0) node[inner sep=0] {\includegraphics[width=0.32\linewidth]{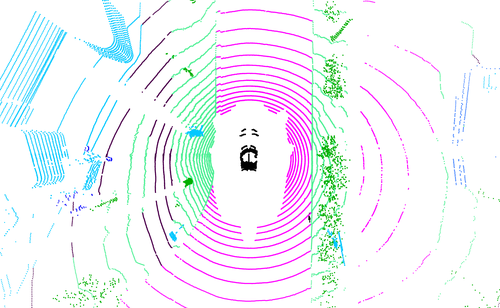}};
    \end{tikzpicture}
    \begin{tikzpicture}
        \draw (0, 0) node[inner sep=0] {\includegraphics[width=0.32\linewidth]{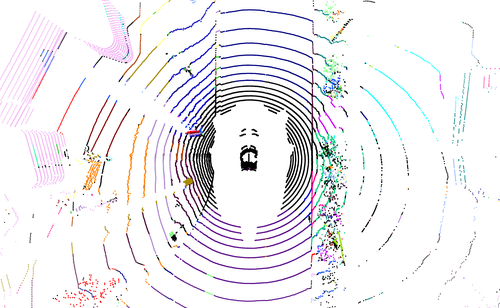}};
    \end{tikzpicture}
    \begin{tikzpicture}
        \draw (0, 0) node[inner sep=0] {\includegraphics[width=0.32\linewidth]{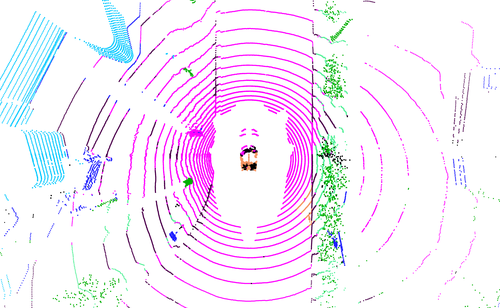}};
    \end{tikzpicture}

    \begin{tikzpicture}
        \draw (0, 0) node[inner sep=0] {\includegraphics[width=0.32\linewidth]{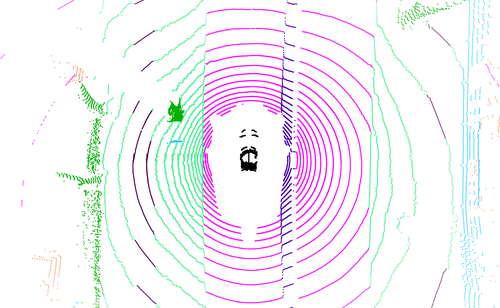}};
    \end{tikzpicture}
    \begin{tikzpicture}
        \draw (0, 0) node[inner sep=0] {\includegraphics[width=0.32\linewidth]{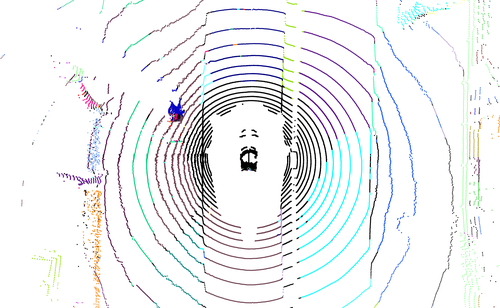}};
    \end{tikzpicture}
    \begin{tikzpicture}
        \draw (0, 0) node[inner sep=0] {\includegraphics[width=0.32\linewidth]{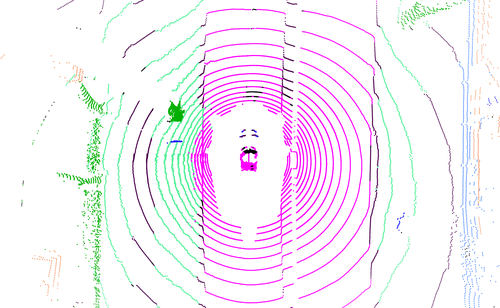}};
    \end{tikzpicture}

    \begin{tikzpicture}
        \draw (0, 0) node[inner sep=0] {\includegraphics[width=0.32\linewidth]{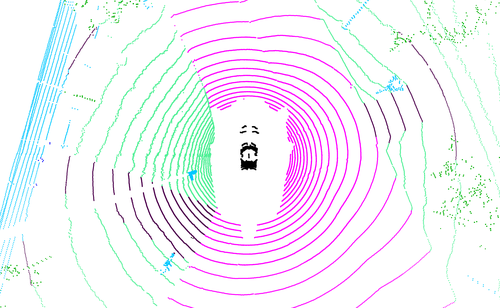}};
    \end{tikzpicture}
    \begin{tikzpicture}
        \draw (0, 0) node[inner sep=0] {\includegraphics[width=0.32\linewidth]{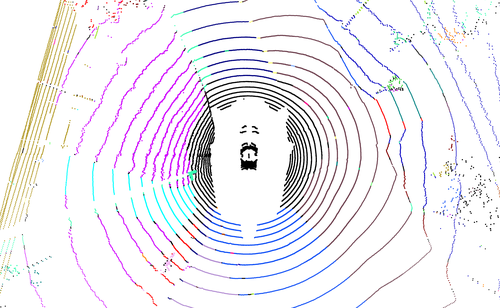}};
    \end{tikzpicture}
    \begin{tikzpicture}
        \draw (0, 0) node[inner sep=0] {\includegraphics[width=0.32\linewidth]{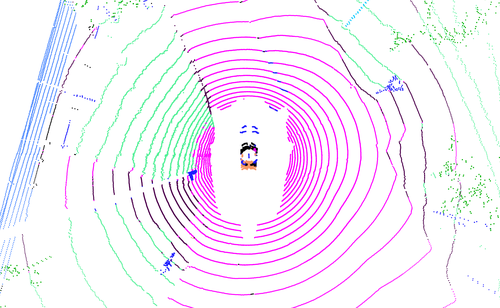}};
    \end{tikzpicture}
    \caption{
    \textbf{Qualitative results for nuScenes.}
    We visualize ground truth (GT), pseudo labels, and our model output for the first scan of validation sequences \emph{0003}, \emph{0013}, \emph{0015}, \emph{0017}, \emph{0035}, and \emph{0038} of nuScenes~\cite{fong21ral}.
    While GT and our output display semantics, the class-agnostic pseudo labels show instances. 
    }
    \label{fig:qual_gt_pseudo_outputs-nuscenes}
\end{figure*}
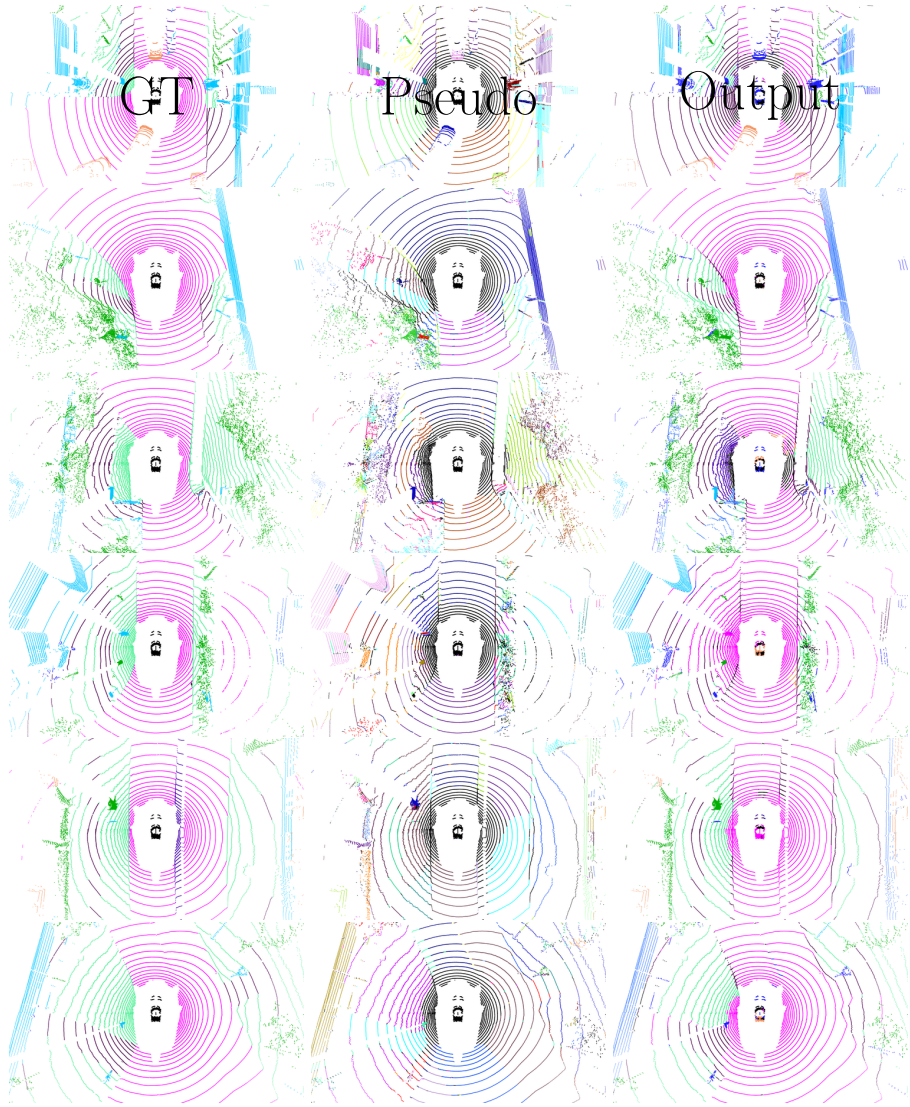

\PAR{Zero-shot semantic segmentation.} To provide further insight, we illustrate semantic ground truth labels, \sal pseudo-labels, that we use to train the \sal model, and \sal model outputs for SemanticKITTI~\cite{behley2019iccv} (\cref{fig:qual_gt_pseudo_outputs-kitti}), nuScenes~\cite{fong21ral} (\cref{fig:qual_gt_pseudo_outputs-nuscenes}) and Waymo Open~\cite{sun20CVPR} datasets.

For ground truth and model outputs, we visualize semantic classes.
Since pseudo-labels are class-agnostic, colors encode object instance IDs in the middle column. 
As can be seen in~\cite{behley2019iccv}, the single-camera setup in SemanticKITTI provides limited supervision only in the frustum. 
Across all shown output examples, the segmentation of \thing and \stuff objects is close to the expected ground truth.

\begin{table*}[t]
    \centering
    \scriptsize
    \caption{
        \textbf{Dataset vocabulary text prompts.}
        To circumvent ambiguous or uninformative class names, we prompt each class with a set of possible text prompts.
        Furthermore, we follow~\cite{radford2021learning} and wrap each class prompt in full-sentence templates.
    }
    \begin{tabular}{l|l}
    \toprule
    Class & Text prompts \\
    \midrule
    \multicolumn{2}{c}{SemanticKITTI~\cite{behley2019iccv}} \\
    \midrule

    \texttt{car} & \texttt{car}, \texttt{jeep}, \texttt{SUV}, \texttt{van} \\
    \texttt{bicycle} & \texttt{bicycle}, \texttt{bike} \\
    \texttt{motorcycle} & \texttt{motorcycle}, \texttt{moped} \\
    \texttt{truck} &\texttt{truck}, \texttt{pickup truck} \\
    \texttt{other-vehicle} & \texttt{other-vehicle}, \texttt{caravan}, \texttt{trailer}, \texttt{bus}, \texttt{tram}, \texttt{train} \\
    \texttt{person} & \texttt{person}, \texttt{pedestrian} \\
    \texttt{bicyclist} & \texttt{bicyclist}, \texttt{bicycle rider} \\
    \texttt{motorcyclist} & \texttt{motorcyclist}, \texttt{motorcycle rider} \\
    \texttt{road} & \texttt{road}, \texttt{lane} \\
    \texttt{parking} & \texttt{parking}, \texttt{parking lot} \\
    \texttt{sidewalk} & \texttt{sidewalk}, \texttt{curb}, \texttt{driveway} \\
    \texttt{other-ground} & \texttt{other-ground}, \texttt{traffic island} \\
    \texttt{building} & \texttt{building}, \texttt{garage}, \texttt{wall}, \texttt{window}, \texttt{stair} \\
    \texttt{fence} & \texttt{fence}, \texttt{separator}, \texttt{small wall}, \texttt{crash barrier} \\
    \texttt{vegetation} & \texttt{vegetation}, \texttt{bush}, \texttt{shrub}, \texttt{foliage}, \texttt{treetop} \\
    \texttt{trunk} & \texttt{trunk}, \texttt{tree trunk} \\
    \texttt{terrain} & \texttt{terrain}, \texttt{gras}, \texttt{soil} \\
    \texttt{pole} & \texttt{pole}, \texttt{lamp post}, \texttt{traffic-sign pole} \\
    \texttt{traffic-sign} & \texttt{traffic-sign}, \texttt{traffic-sign mounting} \\
    \midrule
    \multicolumn{2}{c}{nuScenes~\cite{fong21ral}} \\
    \midrule
    
    \texttt{bicycle} & \texttt{bicycle}, \texttt{bike} \\
    \texttt{bus} & \texttt{bus} \\
    \texttt{car} & \texttt{car}, \texttt{jeep}, \texttt{SUV}, \texttt{van} \\
    \texttt{construction vehicle} & \texttt{construction vehicle}, \texttt{crane}, \texttt{excavator} \\
    \texttt{motorcycle} & \texttt{motorcycle}, \texttt{moped} \\
    \texttt{pedestrian} & \texttt{pedestrian}, \texttt{person} \\
    \texttt{trailer} & \texttt{trailer} \\
    \texttt{truck} & \texttt{truck}, \texttt{pickup truck} \\
    \texttt{barrier} & \texttt{barrier}, \texttt{fence}, \texttt{separator}, \texttt{small wall}, \texttt{crash barrier} \\
    \texttt{traffic cone} & \texttt{traffic cone} \\
    \texttt{driveable surface} & \texttt{driveable surface}, \texttt{road}, \texttt{service lanes}, \texttt{bike lanes} \\
    \texttt{flat surface} & \texttt{flat surface}, \texttt{ground} \\
    \texttt{sidewalk} & \texttt{sidewalk}, \texttt{curbs}, \texttt{driveways} \\
    \texttt{terrain} & \texttt{terrain}, \texttt{gras}, \texttt{soil} \\
    \texttt{manmade} & \texttt{manmade}, \texttt{building}, \texttt{garage}, \texttt{walls}, \texttt{windows}, \texttt{stairs}, \texttt{bench} \\
    \texttt{vegetation} & \texttt{vegetation}, \texttt{bush}, \texttt{shrub}, \texttt{foliage}, \texttt{treetop} \\

    \midrule
    \multicolumn{2}{c}{Super classes} \\
    \midrule
    
    \multirow{3}{*}{\texttt{vehicle}} & \multirow{3}{*}{ \shortstack[l]{\texttt{vehicle}, \texttt{car}, \texttt{truck}, \texttt{bicycle}, \texttt{motorcycle}, \texttt{other-vehicle}, \texttt{jeep}, \texttt{SUV}, \texttt{van},\\ \texttt{bike}, \texttt{moped}, \texttt{pickup truck}, \texttt{caravan}, \texttt{trailer}, \texttt{bus}, \texttt{tram}, \texttt{train}, \\\texttt{construction vehicle}, \texttt{crane}, \texttt{excavator}}} \\
    \\
    \\
    \\
    
    \multirow{2}{*}{\texttt{human}} & \multirow{2}{*}{ \shortstack[l]{\texttt{human}, \texttt{person}, \texttt{bicyclist}, \texttt{motorcyclist}, \texttt{pedestrian}, \texttt{bicycle rider}, \\ \texttt{motorcycle rider}}} \\
    \\
    \\
    
    \multirow{2}{*}{\texttt{ground}} & \multirow{2}{*}{ \shortstack[l]{\texttt{ground}, \texttt{road}, \texttt{sidewalk}, \texttt{parking}, \texttt{other-ground}, \texttt{driveable area}, \\ \texttt{service lane}, \texttt{bike lane}, \texttt{parking lot}, \texttt{curb}, \texttt{driveway}, \texttt{traffic island}}} \\
    \\
    \\
    
    \multirow{2}{*}{\texttt{structure}} & \multirow{2}{*}{ \shortstack[l]{\texttt{structure}, \texttt{building}, \texttt{garage}, \texttt{wall}, \texttt{window}, \texttt{stair}}} \\
    \\
    \\
    
    \multirow{2}{*}{\texttt{nature}} & \multirow{2}{*}{ \shortstack[l]{\texttt{nature}, \texttt{vegetation}, \texttt{trunk}, \texttt{terrain}, \texttt{bush}, \texttt{shrub}, \texttt{foliage}, \texttt{treetop}, \\ \texttt{tree trunk}, \texttt{gras}, \texttt{soil}}} \\
    \\
    \\
    
    \multirow{3}{*}{\texttt{object}} & \multirow{3}{*}{ \shortstack[l]{\texttt{object}, \texttt{fence}, \texttt{pole}, \texttt{traffic-sign}, \texttt{lamp post}, \texttt{traffic-sign pole}, \\ \texttt{traffic-sign mounting}, \texttt{separator}, \texttt{small wall}, \texttt{crash barrier}, \\ \texttt{traffic cone}, \texttt{bench}}} \\
    \\
    \\
    \bottomrule
    \end{tabular}
    \label{tab:text_prompts}
\end{table*}

\begin{figure*}[t]
    \centering
    \begin{tikzpicture}
        \draw (0, 0) node[inner sep=0] {\includegraphics[width=0.42\linewidth]{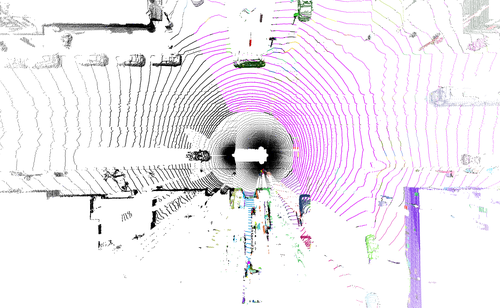}};
        \draw (0, 0) node {\huge \color{black} Pseudo};
    \end{tikzpicture}
    \begin{tikzpicture}
        \draw (0, 0) node[inner sep=0] {\includegraphics[width=0.42\linewidth]{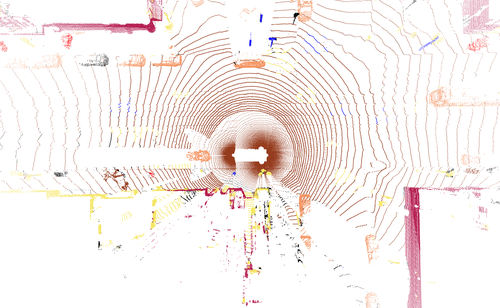}};
        \draw (0, 0) node {\huge \color{black} Output};
    \end{tikzpicture}

    \begin{tikzpicture}
        \draw (0, 0) node[inner sep=0] {\includegraphics[width=0.42\linewidth]{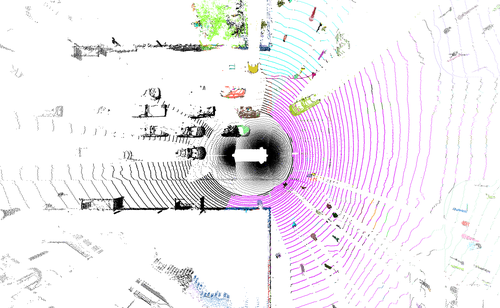}};
    \end{tikzpicture}
    \begin{tikzpicture}
        \draw (0, 0) node[inner sep=0] {\includegraphics[width=0.42\linewidth]{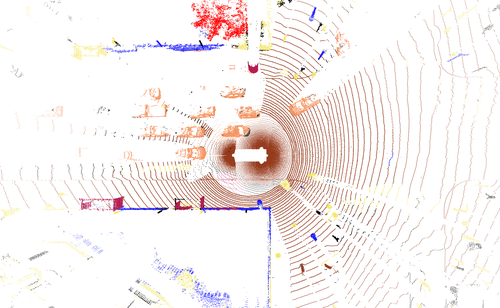}};
    \end{tikzpicture}

    \begin{tikzpicture}
        \draw (0, 0) node[inner sep=0] {\includegraphics[width=0.42\linewidth]{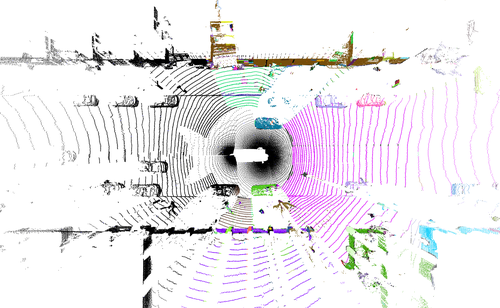}};
    \end{tikzpicture}
    \begin{tikzpicture}
        \draw (0, 0) node[inner sep=0] {\includegraphics[width=0.42\linewidth]{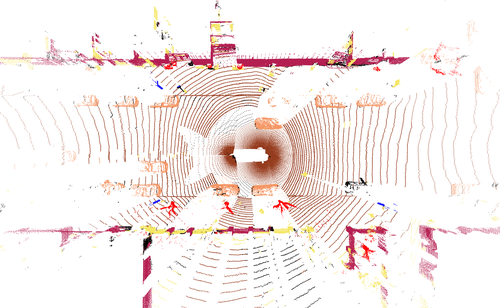}};
    \end{tikzpicture}

    \caption{
    \textbf{Qualitative results for Waymo.}
    We visualize pseudo labels and our model output for the first scan of test sequences \emph{008}, \emph{024}, and \emph{032} of Waymo~\cite{sun20CVPR}.
    Waymo does not provide panoptic ground truth labels.
    While our output displays semantics, the class-agnostic pseudo labels show instances. 
    }
    \label{fig:qual_gt_pseudo_outputs-waymo}
\end{figure*}

\end{document}